\documentclass{article}

\usepackage[utf8]{inputenc} 
\usepackage[T1]{fontenc}    
\usepackage{booktabs}       
\usepackage{amsfonts}       
\usepackage{nicefrac}       
\usepackage{color}
\usepackage{xcolor}
\usepackage{graphicx}
\usepackage{floatrow}
\usepackage[percent]{overpic}
\usepackage{amsmath}

\usepackage{algorithm}
\usepackage{algpseudocode}

\usepackage{microtype}
\usepackage{amsmath}
\usepackage{mathtools}
\usepackage{amsfonts}
\usepackage{graphicx}
\usepackage{color}
\usepackage{booktabs} 
\usepackage{wrapfig}
\usepackage{fullpage}

\usepackage{hyperref}       
\usepackage{url}            

\usepackage{breakurl}
\hypersetup{
    colorlinks=true,
    linkcolor=blue,
    filecolor=magenta,      
    urlcolor=blue,
    breaklinks=true,
}

\newcommand{\omt}[1]{}

\title{A Survey of Deep Learning for Scientific Discovery}
\date{}

\author{
  Maithra Raghu$^{1, 2}$\thanks{Correspondence to  \texttt{maithrar@gmail.com}}
  \qquad
  Eric Schmidt$^{1, 3}$ \\
  $^1$ Google \\
  $^2$ Cornell University \\
  $^3$ Schmidt Futures \\
}

\begin{document}

\maketitle

\begin{abstract}
Over the past few years, we have seen fundamental breakthroughs in core problems in machine learning, largely driven by advances in deep neural networks. At the same time, the amount of data collected in a wide array of scientific domains is dramatically increasing in both size and complexity. Taken together, this suggests many exciting opportunities for deep learning applications in scientific settings. But a significant challenge to this is simply knowing where to start. The sheer breadth and diversity of different deep learning techniques makes it difficult to determine what scientific problems might be most amenable to these methods, or which specific combination of methods might offer the most promising first approach. In this survey, we focus on addressing this central issue, providing an overview of many widely used deep learning models, spanning visual, sequential and graph structured data, associated tasks and different training methods, along with techniques to use deep learning with less data and better interpret these complex models --- two central considerations for many scientific use cases. We also include overviews of the full design process, implementation tips, and links to a plethora of tutorials, research summaries and open-sourced deep learning pipelines and pretrained models, developed by the community. We hope that this survey will help accelerate the use of deep learning across different scientific domains.
\end{abstract}

\section{Introduction}
The past few years have witnessed extraordinary advances in machine learning using deep neural networks. Driven by the rapid increase in available data and computational resources, these neural network models and algorithms have seen remarkable developments, and are a staple technique in tackling fundamental tasks ranging from speech recognition \cite{graves2013speech, oord2016wavenet}, to complex tasks in computer vision such as image classification, (instance) segmentation, action recognition \cite{krizhevsky2012imagenet, he2017mask, wang2018non}, and central problems in natural language, including question answering, machine translation and summarization \cite{rajpurkar2016squad, paulus2017deep, vaswani2017attention, rush2015neural}. Many of these fundamental tasks (with appropriate reformulation) are relevant to a much broader array of domains, and in particular have tremendous potential in aiding the investigation of central scientific questions.

However, a significant obstacle in beginning to use deep learning is simply knowing where to start. The vast research literature, coupled with the enormous number of underlying models, tasks and training methods makes it very difficult to identify which techniques might be most appropriate to try, or the best way to start implementing them. 

The goal of this survey is to help address this central challenge. In particular, it has the following attributes:
\begin{itemize}
    \item The survey overviews a highly diverse set of deep learning concepts, from deep neural network models for varied data modalities (CNNs for visual data, graph neural networks, RNNs and Transformers for sequential data) to the many different key tasks (image segmentation, super-resolution, sequence to sequence mappings and many others) to the multiple ways of training deep learning systems.
    \item But the explanation of these techniques is relatively high level and concise, to ensure the core ideas are accessible to a broad audience, and so that the entire survey can be read end to end easily.
    \item From the perspective of aiding scientific applications, the survey describes in detail (i) methods to use deep learning with less data (self-supervision, semi-supervised learning, and others) and (ii) techniques for interpretability and representation analysis (for going beyond predictive tasks). These are two exciting and rapidly developing research areas, and are also of particular significance to possible scientific use cases.
    \item The survey also focuses on helping quickly ramp up implementation, and in addition to overviews of the entire deep learning design process and a section on implementation tips (Section \ref{sec-implementation-tips}), the survey has a plethora of open-sourced code, research summaries and tutorial references developed by the community throughout the text, including a full section (Section \ref{sec-frameworks-resources}) dedicated to this.
\end{itemize}

\textbf{Who is this survey for?}
We hope this survey will be especially helpful for those with a basic understanding of machine learning, interested in (i) getting a comprehensive but accessible overview of many fundamental deep learning concepts and (ii) references and guidance in helping ramp up implementation. Beyond the core areas of deep learning, the survey focuses on methods to develop deep learning systems with less data, and techniques for interpreting these models, which we hope will be of particular use for those interested in applying these techniques in scientific problems. However, these topics and many others presented, along with the many code/tutorial/paper references may be helpful to anyone looking to learn about and implement deep learning.

\subsection{Outline of Survey}
The survey is structured as follows: 
\begin{itemize}
    \item Section \ref{sec-high-level-considerations} starts with some high level considerations for using deep learning. Specifically, we first discuss some template ways in which deep learning might be applied in scientific domains, followed by a general overview of the entire deep learning design process, and conclude with a brief discussion of other central machine learning techniques that may be better suited to some problems. The first part may be of particular interest to those considering scientific applications, while the latter two parts may be of general interest.
    \item Section \ref{sec-frameworks-resources} provides references to tutorials, open-sourced code model/algorithm implementations, and websites with research paper summaries, all developed by the deep learning community. This section should be very helpful for many readers and we encourage skimming through the links provided.
    \item Section \ref{sec-standardnns-tasks} then overviews many of the standard tasks and models in deep learning, covering convolutional networks and their many uses, graph neural networks, sequence models (RNNs, Transformers) and the many associated sequence tasks. 
    \item Section \ref{sec-supervised-methods} looks at some key variants of the supervised learning training process, such as transfer learning, domain adaptation and multitask learning. These are central to many successful applications of deep learning.
    \item Section \ref{sec-less-data} considers ways to improve the data efficiency for developing deep neural network models, which has been a rapidly evolving area of research, and a core consideration for many applications, including scientific domains. It covers the many variants of self-supervision and semi-supervised learning, as well as data augmentation and data denoising.
    \item Section \ref{sec-interp-inspect-analysis} overviews advances in interpretability and representational analysis, a set of techniques focused on gaining insights into the internals of the end-to-end system: identifying important features in the data, understanding its effect on model outputs and discovering properties of model hidden representations. These are very important for many scientific problems which emphasise understanding over predictive accuracy, and may be of broader interest for e.g. aiding model debugging and preemptively identifying failure modes.
    \item Section \ref{sec-advanced-dl} provides a brief overview of more advanced deep learning methods, specifically generative modelling and reinforcement learning. 
    \item Section \ref{sec-implementation-tips} concludes with some key implementation tips when putting together an end-to-end deep learning system, which we encourage a quick read through!
\end{itemize}

\section{High Level Considerations for Deep Learning}
\label{sec-high-level-considerations}
In this section we first discuss some high level considerations for deep learning techniques. We start with overviews of template ways in which deep learning might be applied in scientific settings, followed by a discussion of the end-to-end design process and some brief highlights of alternate machine learning methods which may be more suited to some problems.

\subsection{Templates for Deep Learning in Scientific Settings}
What are the general ways in which we might apply deep learning techniques in scientific settings? At a very high level, we can offer a few \textit{templates} of ways in which deep learning might be used in such problems:
\begin{itemize}
\item[(1)] \textit{Prediction Problems} Arguably the most straightforward way to apply deep learning is to use it to tackle important \textit{prediction problems}: mapping inputs to predicted outputs.  This predictive use case of deep learning is typically how it is also used in core problems in computing and machine learning. For example, the input might be a biopsy image, and the model must output a prediction of whether the imaged tissue shows signs of cancer. We can also think of this predictive use case as getting the model to \textit{learn a target function}, in our example, mapping from input visual features to the cancer/no cancer output. Using deep learning in this way also encapsulates settings where the target function is very complex, with no mathematical closed form or logical set of rules that describe how to go from input to output. For instance, we might use a deep learning model to (black-box) \textit{simulate} a complex process (e.g. climate modelling), that is very challenging to explicitly model \cite{kasim2020up}.  

\item[(2)] \textit{From Predictions to Understanding} One fundamental difference between scientific questions and core machine learning problems is the emphasis in the former on \textit{understanding} the underlying mechanisms. Oftentimes, outputting an accurate prediction alone is not enough. Instead, we want to gain interpretable insights into what properties of the data or the data generative process led to the observed prediction or outcome. To gain these kinds of insights, we can turn to interpretability and representation analysis methods in deep learning, which focus on determining how the neural network model makes a specific prediction. There has been significant work on both tools to understand what features of the input are most critical to the output prediction, as well as techniques to directly analyze the \textit{hidden representations} of the neural network models, which can reveal important properties of the underlying data.

\item[(3)] \textit{Complex Transformations of Input Data} In many scientific domains, the amount of generated data, particularly visual data (e.g. fluorescence microscopy, spatial sequencing, specimen videos \cite{power2017guide, ji2017adaptive}) has grown dramatically, and there is an urgent need for efficient analysis and automated processing. Deep learning techniques, which are capable of many complex transformations of data, can be highly effective for such settings, for example, using a deep neural network based segmentation model to automatically identify the nuclei in images of cells, or a pose estimation system to rapidly label behaviors seen in videos of mice for neuroscience analysis.
\end{itemize}

\subsection{Deep Learning Workflow}
\begin{figure}
\centering
\includegraphics[width=0.9\columnwidth]{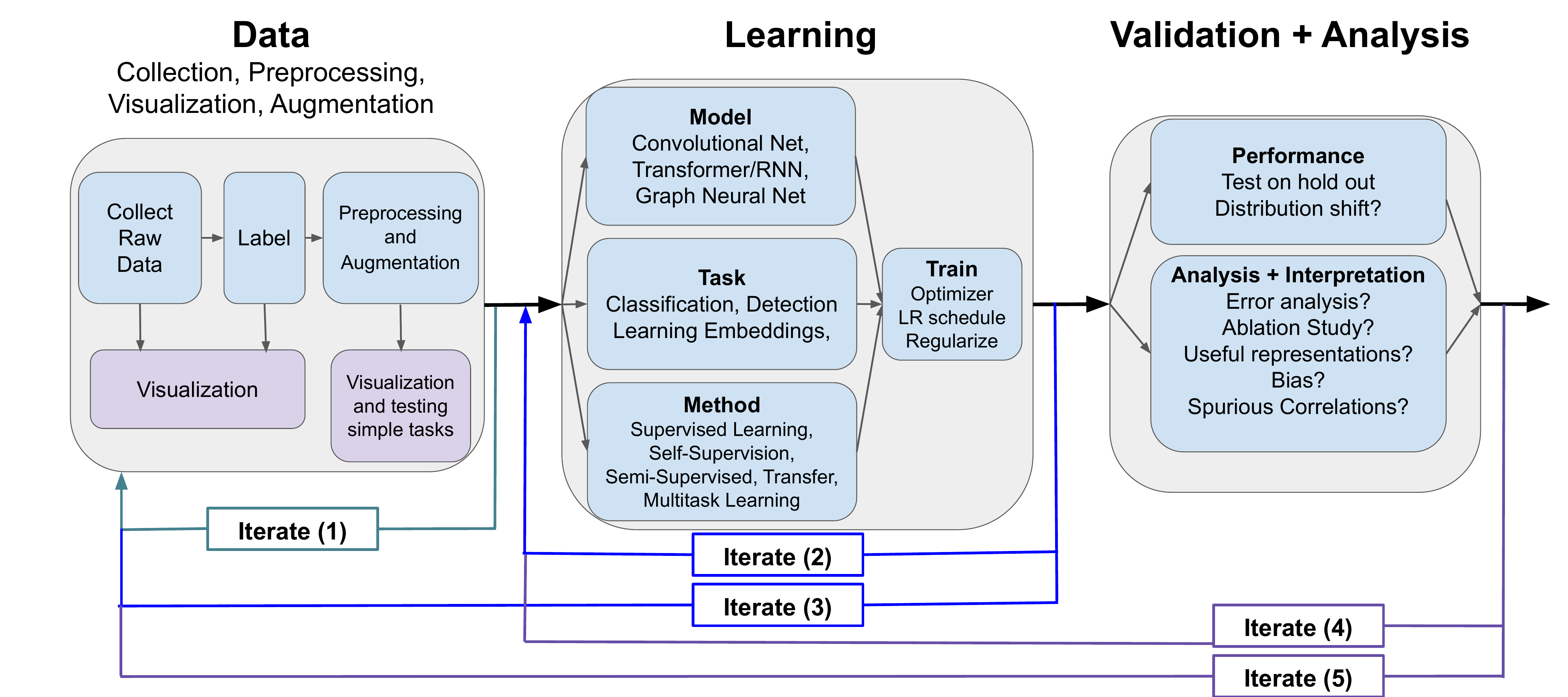} 
\caption{\small \textbf{Schematic of a typical deep learning workflow.} A typical development process for deep learning applications can be viewed as consisting of three sequential stages (i) data related steps (ii) the learning component (iii) validation and analysis. Each one of these stages has several substeps and techniques associated with it, also depicted in the figure. In the survey we will overview most techniques in the learning component, as well as some techniques in the data and validation stages. Note that while a natural sequence is to first complete steps in the data stage, followed by learning and then validation, standard development will likely result in multiple different iterations where the techniques used or choices made in one stage are revisited based off of results of a later stage.}
\label{fig:deeplearning-workflow}
\end{figure}

With these examples of templates for deep learning applications in science, we next look at the end to end workflow for designing a deep learning system. Figure \ref{fig:deeplearning-workflow} illustrates what a typical workflow might look like. 

Having selected the overarching (predictive) problem of interest, we can broadly think of having three stages for designing and using the deep learning system: (i) \textit{data} related steps, such as collection, labelling, preprocessing, visualization, etc (ii) \textit{learning} focused steps, such as choice of deep neural network model, the task and method used to train the model (iii) \textit{validation and analysis} steps, where performance evaluations are conducted on held out data, as well as analysis and interpretation of hidden representations and ablation studies of the overall methods. 

These three stages are naturally sequential. However, almost all of the time, the first attempt at building an end-to-end deep learning system will result in some kind of failure mode. To address these, it is important to keep in mind the \textit{iterative} nature of the design process, with results from the different stages informing the redesign and rerunning of other stages. 

Figure \ref{fig:deeplearning-workflow} provides some examples of common iterations with the backward connecting arrows: (i) the \textit{Iterate (1)} arrow, corresponding to iterations on the data collection process, e.g. having performed some data visualization, the labelling process for the raw instances might require adjusting --- the first labelling mechanism might be too noisy, or not capture the objective of interest (ii) the \textit{Iterate (2)} arrow, corresponding to iterations on the learning setup, due to e.g. deciding that a different task or method might be more appropriate, or decomposing the learning process into multiple steps --- first performing self-supervision followed by supervised learning (iii) the \textit{Iterate (3)} arrow, changing the data related steps based off of the results of the learning step (iv) the \textit{Iterate (4)} arrow, redesigning the learning process informed by the validation results e.g. finding out the model has overfit on the training data at validation and hence reducing training time or using a simpler model (v) the \textit{Iterate (5)} arrow, adapting the data steps based off the validation/analysis results, e.g. finding that the model is relying on spurious attributes of the data, and improving data collection/curation to mitigate this.

\paragraph{Focus of Survey and Nomenclature}
In this survey, we provide a comprehensive overview of many of the techniques in the \textit{learning stage}, along with some techniques (e.g. data augmentation, interpretability and representation analysis, Section \ref{sec-interp-inspect-analysis}) in the data and validation stages. 

For the learning stage, we look at popular \textit{models}, \textit{tasks} and \textit{methods}. By \textit{models} (also sometimes referred to as \textit{architecture}), we mean the actual structure of the deep neural network --- how many layers, of what type, and how many neurons, etc. By \textit{tasks}, we mean the kind of prediction problem, specifically, the type of input and output. For example, in an \textit{image classification} task, the input consists of images and the output a probability distribution over a (discrete) set of different categories (called classes). By \textit{methods}, we refer to the type of learning process used to train the system. For example, \textit{supervised learning} is a very general learning process, consisting of the neural network being given data instances with corresponding \textit{labels}, with the labels providing \textit\textit{supervision}. 

Unlike different models and tasks, methods can be subsets of other methods. For example, \textit{self-supervision}, a method where the neural network is trained on data instances and labels, but the labels automatically created from the data instance, can also be considered a type of \textit{supervised learning}. This can be a little confusing! But it suffices to keep in mind the general notions of \textit{models}, \textit{tasks} and \textit{methods}.

\subsection{Deep Learning or Not?}
As a final note before diving into the different deep learning techniques, when formulating a problem, it is important to consider whether deep learning provides the right set of tools to solve it. The powerful underlying neural network models offer many sophisticated functionalities, such learned complex image transforms. However, in many settings, deep learning may not be the best technique to start with or best suited to the problem. Below we very briefly overview some of the most ubiquitous machine learning methods, particularly in scientific contexts.

\paragraph{Dimensionality Reduction and Clustering} In scientific settings, the ultimate goal of data analysis is often \textit{understanding} --- identifying the underlying mechanisms that give rise to patterns in the data. When this is the goal, \textit{dimensionality reduction}, and/or \textit{clustering} are simple (unsupervised) but  highly effective methods to reveal hidden properties in the data. They are often very useful in the important first step of exploring and visualizing the data (even if more complex methods are applied later.)

\textit{Dimensionality Reduction:} Dimensionality reduction methods are either \textit{linear}, relying on a linear transformation to reduce data dimensionality, or \textit{non-linear}, reducing dimensionality while approximately preserving the non-linear (manifold) structure of the data. Popular linear dimensionality reduction methods include \textit{PCA} and \textit{non-negative matrix factorization}, with some popular non-linear methods including \textit{t-SNE} \cite{maaten2008visualizing} and \textit{UMAP} \cite{mcinnes2018umap}. Most dimensionality reduction methods have high-quality implementations in packages like \texttt{scikit-learn} or on github, e.g. \url{https://github.com/oreillymedia/t-SNE-tutorial} or \url{https://github.com/lmcinnes/umap}.

\textit{Clustering:} Often used in combination with dimensionality reduction, clustering methods provide a powerful, unsupervised way to identify similarities and differences across the data population. Commonly used clustering methods include \textit{k-means} (particularly the \textit{k-means++} variant), \textit{Gaussian Mixture Models} (GMMs), \textit{hierarchical clustering} and \textit{spectral clustering}. Like dimensionality reduction techniques, these clustering methods have robust implementations in packages like \texttt{scikit-learn}. 

In Section \ref{sec-dim-reduction-representations}, we discuss how dimensionality reduction and clustering can be used on the hidden representations of neural networks.

\paragraph{Linear Regression, Logistic Regression (and variants!)} Arguably the most fundamental techniques for \textit{supervised} problems like classification and regression, linear and logistic regression, and their variants (e.g. Lasso, Ridge Regression) may be particularly useful when there is limited data, and a clear set of (possibly preprocessed) features (such as in tabular data.) These methods also often provide a good way to sanity check the overarching problem formulation, and may be a good starting point to test out a very simple version of the full problem. Due to their simplicity, linear and logistic regression are highly interpretable, and provide straightforward ways to perform \textit{feature attribution}.

\paragraph{Decision Trees, Random Forests and Gradient Boosting} Another popular class of methods are decision trees, random forests and gradient boosting. These methods can also work with regression/classification tasks, and are well suited to model non-linear relations between the input features and output predictions. Random forests, which ensemble decision trees, can often be preferred to deep learning methods in settings where the data has a low signal-to-noise ratio. These methods can typically be less interpretable than linear/logistic regression, but recent work \cite{nori2019interpretml} has looked at developing software libraries \url{https://github.com/interpretml/interpret} to address this challenge.

\paragraph{Other Methods and Resources:} Both the aforementioned techniques and many other popular methods such as graphical models, Gaussian processes, Bayesian optimization are overviewed in detail in excellent course notes such as \href{http://www.cs.toronto.edu/~rgrosse/courses/csc411_f18/}{University of Toronto`s Machine Learning Course} or \href{http://cs229.stanford.edu/syllabus.html}{Stanford`s CS229}, detailed articles at \url{https://towardsdatascience.com/} and even interactive textbooks such as \url{https://d2l.ai/index.html} (called Dive into Deep Learning \cite{zhang2019dive}) and \url{https://github.com/rasbt/python-machine-learning-book-2nd-edition}.

\section{Deep Learning Libraries and Resources}
\label{sec-frameworks-resources}
A remarkable aspect of advances in deep learning so far is the enormous number of resources developed and shared by the community. These range from tutorials, to overviews of research papers, to open sourced code. Throughout this survey, we will reference some of these materials in the topic specific sections, but we first list here a few general very useful frameworks and resources.

\paragraph{Software Libraries for Deep Learning:} Arguably the two most popular code libraries for deep learning are \href{https://pytorch.org/}{PyTorch} (with a high level API called \href{https://github.com/PyTorchLightning/pytorch-lightning}{Lightning}) and \href{https://www.tensorflow.org/}{TensorFlow} (which also offers \href{https://keras.io/}{Keras} as a high level API.) Developing and training deep neural network models critically relies on fast, parallelized matrix and tensor operations (sped up through the use of Graphical Processing Units) and performing automatic differentiation for computing gradients and optimization (known as \textit{autodiff}.) Both PyTorch and TensorFlow offer these core utilities, as well as many other functions. Other frameworks include \href{https://chainer.org/}{Chainer}, \href{https://onnx.ai/}{ONNX}, \href{https://mxnet.apache.org/}{MXNET} and \href{https://github.com/google/jax}{JAX}. Choosing the best framework has been the source of significant debate. For ramping up quickly, programming experiences closest to native Python, and being able to use many existing code repositories, PyTorch (or TensorFlow with the Keras API) may be two of the best choices. 

\paragraph{Tutorials:} (i) \url{https://course.fast.ai/} fast.ai provides a free, coding-first course on the most important deep learning techniques as well as an intuitive and easy to use code library, \url{https://github.com/fastai/fastai}, for model design and development. (ii) \url{https://towardsdatascience.com/} contains some fantastic tutorials on almost every deep learning topic imaginable, crowd sourced from many contributors. (iii) Many graduate deep learning courses have excellent videos and lecture notes available online, such as \url{http://www.cs.toronto.edu/~rgrosse/courses/csc421_2019/} for Deep Learning and Neural Networks, or the more topic specific \href{https://www.youtube.com/playlist?list=PLoROMvodv4rOhcuXMZkNm7j3fVwBBY42z}{Stanford`s CS224N NLP with Deep Learning}. A nice collection of some of these topic specific lectures is provided at \url{https://github.com/Machine-Learning-Tokyo/AI_Curriculum}. There are also some basic interactive deep learning courses online, such as \url{https://github.com/leriomaggio/deep-learning-keras-tensorflow}.

\paragraph{Research Overviews, Code, Discussion:} (i) \url{https://paperswithcode.com/} This excellent site keeps track of new research papers and their corresponding opensourced code, trending directions and displays state of the art results (\url{https://paperswithcode.com/sota}) across many standard benchmarks. (ii) Discussion of deep learning research is very active on Twitter. \url{http://www.arxiv-sanity.com/top} keeps track of some of the top most discussed papers and comments. (iii) \url{https://www.reddit.com/r/MachineLearning/} is also a good forum for research and general project discussion. (iv) \url{https://www.paperdigest.org/conference-paper-digest/} contains snippets of all the papers in many different top machine learning conferences. (v) IPAM (Institute for Pure and Applied Mathematics) has a few programs e.g. \url{https://www.ipam.ucla.edu/programs/workshops/new-deep-learning-techniques/?tab=schedule} and \url{https://www.ipam.ucla.edu/programs/workshops/deep-learning-and-medical-applications/?tab=schedule} with videos overviewing deep learning applications in science.

\paragraph{Models, Training Code and Pretrained Models:} As we discuss later in the survey, publicly available models, training code and \textit{pretrained} models are very useful for techniques such as transfer learning. There are many good sources of these, here are a few that are especially comprehensive and/or accessible:
\begin{itemize}
\item[(i)] Pytorch and TensorFlow have a collection of pretrained models, found at \url{https://github.com/tensorflow/models} and \url{https://pytorch.org/docs/stable/torchvision/models.html}.
\item[(ii)] \url{https://github.com/huggingface} Hugging Face (yes, that really is the name), offers a huge collection of both pretrained neural networks and the code used to train them. Particularly impressive is their library of \textit{Transformer} models, a one-stop-shop for sequential or language applications.
\item[(iii)]  \url{https://github.com/rasbt/deeplearning-models} offers many standard neural network architectures, including multilayer perceptrons, convolutional neural networks, GANs and Recurrent Neural Networks.
\item[(iv)] \url{https://github.com/hysts/pytorch_image_classification} does a deep dive into image classification architectures, with training code, highly popular data augmentation techniques such as \textit{cutout}, and careful speed and accuracy benchmarking. See their page for some object detection architectures also.
\item[(v)]  \url{https://github.com/openai/baselines} provides implementations of many popular RL algorithms.
\item[(vi)] \url{https://modelzoo.co/} is a little like paperswithcode, but for models, linking to implementations of neural network architectures for many different standard problems.
\item[(vii)] \url{https://github.com/rusty1s/pytorch_geometric}. Implementations and paper links for many graph neural network architectures.
\end{itemize}

\paragraph{Data Collection, Curation and Labelling Resources:} A crucial step in applying deep learning to a problem is collecting, curating and labelling data. This is a very important, time-intensive and often highly intricate task (e.g. labelling object boundaries in an image for segmentation.) Luckily, there are some resources and libraries to help with this, for example \url{https://github.com/tzutalin/labelImg}, \url{https://github.com/wkentaro/labelme}, \url{https://rectlabel.com/} for images and \url{https://github.com/doccano/doccano} for text/sequential data.

\paragraph{Visualization, Analysis and Compute Resources:} When training deep neural network models, it is critical to visualize important metrics such as loss and accuracy \textit{while} the model is training. Tensorboard \url{https://www.tensorflow.org/tensorboard} (which works with Pytorch and TensorFlow) is a very popular framework for doing this. Related is the \textit{colab} effort \url{https://colab.research.google.com/notebooks/welcome.ipynb}, which, aside from providing a user-friendly, interactive way for model development and analysis (very similar to \href{https://jupyter.org/}{jupyter notebooks}) also provides some (free!) compute resources. 

\section{Standard Neural Network Models and Tasks}
\label{sec-standardnns-tasks}
In this section, we overview the standard \textit{neural network models} and the kinds of \textit{tasks} they can be used for, from convolutional networks for image predictions and transformations to transformer models for sequential data to graph neural networks for chemistry applications.

\subsection{Supervised Learning}
\begin{figure}
\centering
\includegraphics[width=0.8\columnwidth]{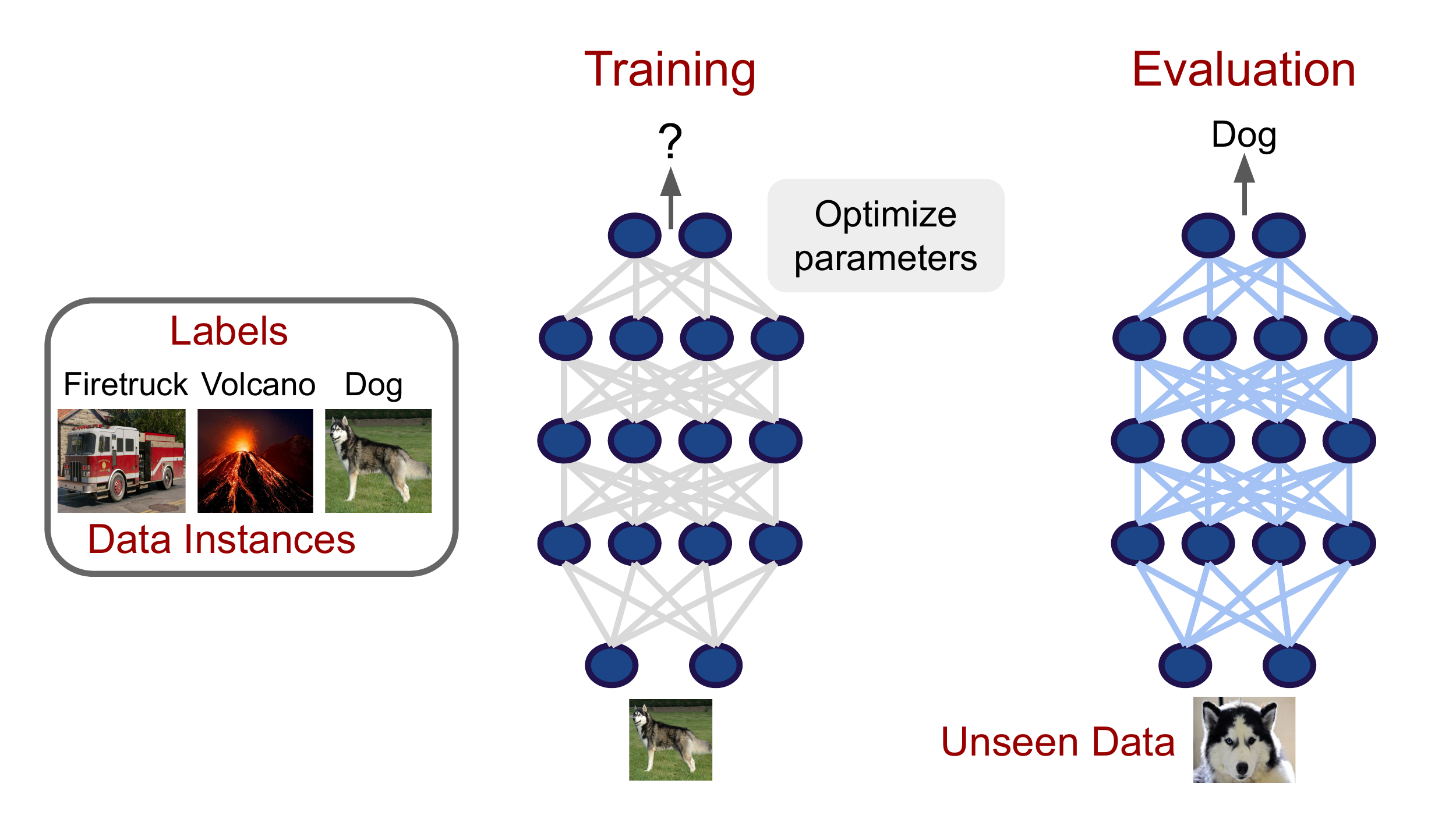} 
\caption{\small \textbf{The Supervised Learning process for training neural networks.} The figure illustrates the supervised learning process for neural networks. Data instances (in this case images) and corresponding labels are collected. During the training step, the parameters of the neural network are optimized so that when input a data instance, the neural network outputs the corresponding label. During evaluation, the neural network is given unseen data instances as input, and if trained successfully, will output a meaningful label (prediction).}
\label{fig:supervised-learning}
\end{figure}
Before diving into the details of the different deep neural network models, it is useful to briefly discuss \textit{supervised learning}, the most standard method to \textit{train} these models. In the supervised learning framework, we are given data instances and an associated label for each data instance, i.e. (data instance, label) pairs. For example, the data instances might comprise of chest x-ray images, and the labels (one for each chest x-ray image) a binary yes/no to whether it shows the symptoms of pneumonia. \textit{Training} the neural network model then consists of finding values for its parameters so that when it is fed in a data instance (chest x-ray) as input, it correctly outputs the corresponding label (yes/no on whether the chest x-ray has pneumonia.) To find these parameter values, we perform iterative \textit{optimization} to guide the neural network parameters to appropriate values, using the given labels to provide supervision. Figure \ref{fig:supervised-learning} shows a schematic of the supervised learning setup for deep learning.

Supervised learning is the most basic yet most critical method for training deep neural networks. As will be seen through the subsequent sections, there can be significant diversity in the kinds of (data, label) pairs used. Even in settings where clear (data, label) pairs are not possible to collect (Sections \ref{sec-self-supervised}, \ref{sec-semi-supervised}), the training problem is often reformulated and recast into a supervised learning framework. 

\subsection{Multilayer Perceptrons}
The first and most basic kind of deep neural network is the \textit{multilayer perceptron.} These models consist of a stack of fully connected layers (matrix multiplications) interleaved with a nonlinear transform. 

Despite their simplicity, they are useful for problems where the data might consist of a set of distinct, (possibly categorical) features, for example, tabular data. These models have more expressive power than logistic/linear regression, though those methods would be a good first step to try. One way to apply these models might be to first preprocess the data to compute the distinct set of features likely to be important, and use this as input. \url{https://github.com/rasbt/deeplearning-models} provides some implementations of some example multilayer perceptron architectures.

\paragraph{Scientific Examples} One recent scientific example is given by the use of simple MLPs for pharamaceutical formulation \cite{yang2019deep}, developing variants of a drug that is stable and safe for patient use.

\subsection{Convolutional Neural Networks}
These are arguably the most well known family of neural networks, and are very useful in working with any kind of image data. They are characterized by having convolutional layers, which allow the neural network to reuse parameters across different spatial locations of an image. This is a highly useful inductive bias for image data, and helping with efficiently learning good features, some, like Gabor filters, which correspond to traditional computer vision techniques. Convolutional neural networks (CNNs) have so many possible uses that we overview some of the most ubiquitous tasks separately below.

\subsubsection{Image Classification}
This is arguably the simplest and most well known application of convolutional neural networks. The model is given an input image, and wants to output a class --- one of a (typically) mutually exclusive set of labels for that image. The earlier example, of mapping a chest x-ray image to a binary disease label, is precisely image classification.

Convolutional neural networks for image classification is an extremely common application of deep learning. There many different types of CNN models for classification: \textit{VGG} --- a simple stack of convolutional layers followed by a fully connected layer \cite{simonyan2014very}, \textit{ResNets} --- which are a family of convolutional networks of different sizes and depths and \textit{skip connections} \cite{he2016deep}, \textit{DenseNets} --- another family of models where unlike standard neural networks, every layer in a "block" is connected to every other layer \cite{huang2017densely}. More recent, complex models include \textit{ResNeXt} \cite{xie2017aggregated} and recently \textit{EfficientNets}, which have separate scaling factors for network depth, width and the spatial resolution of the input image \cite{tan2019efficientnet}. Tutorials, implementations and pretrained versions of many of these models can be found in the references given in Section \ref{sec-frameworks-resources}.

\paragraph{Scientific Examples:} Image classification has found many varied scientific applications, such as in analyzing cryoEM data \cite{tegunov2018real} (with associated code \url{https://github.com/cramerlab/boxnet}). An especially large body of work has looked at \textit{medical imaging} uses of image classification, specifically, using CNNs to predict disease labels. Examples range from ophthalmology \cite{gulshan2016development}, radiology (2D x-rays and 3D CT scans) \cite{yasaka2017deep, anthimopoulos2016lung, rajpurkar2017chexnet}, pathology \cite{liu2017detecting, esteva2017dermatologist}, analyzing brain scans (PET, fMRI) \cite{sarraf2016deepad, ding2018deep}. An excellent survey of the numerous papers in this area is given by \cite{topol2019high}. 

\subsubsection{Object Detection}
\begin{figure}
\centering
\includegraphics[width=0.7\columnwidth]{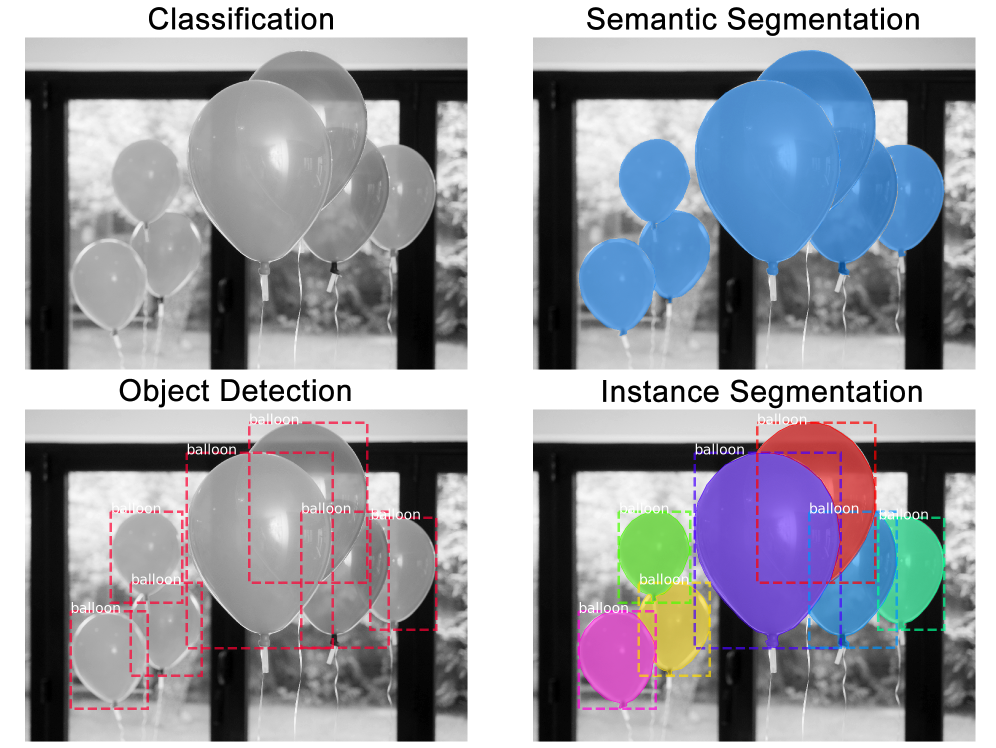} 
\caption{\small \textbf{Differences between Image Classification, Object Detection, Semantic Segmentation and Instance Segmentation tasks. Image source \cite{abdulla2018segmentation}} The figure illustrates the differences between classification, object detection, semantic segmentation and instance segmentation. In classification, the whole image gets a single label (balloons), while in object detection, each balloon is also localized with a bounding box. In semantic segmentation, all the pixels corresponding to balloon are identified, while in instance segmentation, each individual balloon is identified separately.}
\label{fig:cnn-tasks}
\end{figure}
Image classification can be thought of as a global summary of the image. Object detection dives into some of the lower level details of the image, and looks at identifying and \textit{localizing} different objects in the image. For example, given an input image of an outdoor scene having a dog, a person and a tree, object detection would look at both identifying the presence of the dog, person and tree and `circle their location' in the image --- specifically, put a \textit{bounding box} around each of them. The supervised learning task is thus to take an input image and output the coordinates of these bounding boxes, as well as categorizing the kind of object they contain. 

Like image classification, there are many high performing and well established convolutional architectures for object detection. Because of the intricacy of the output task, these models tend to be more complex with a \textit{backbone} component (using an image classification model) and a \textit{region proposal} component for bounding box proposals. But there are still many pretrained models available to download. One of the most successful early models was \textit{Faster R-CNN} \cite{ren2015faster}, which significantly sped up the slow bounding box proposal component. Since then there have been many improved models, including \textit{YOLOv3} \cite{redmon2018yolov3}, and most recently \textit{EfficientDets} \cite{tan2019efficientdet}. Arguably the most popular recent architecture however has been \textit{Mask R-CNN} and its variants \cite{he2017mask, wu2019detectron2}. Mask R-CNN performs some segmentation as well as object detection (see below). Besides some of the resources mentioned in Section \ref{sec-frameworks-resources}, a good source of code and models is \url{https://github.com/rbgirshick}, one of the key authors in a long line of these object detection models. (Note though that there are many other popular implementations, such as \url{https://github.com/matterport/Mask_RCNN}.) This in depth article \href{https://towardsdatascience.com/faster-r-cnn-object-detection-implemented-by-keras-for-custom-data-from-googles-open-images-125f62b9141a}{towardsdatascience object detection Faster R-CNN} offers a detailed tutorial on downloading, setting up and training an object detection model, including helpful pointers to data collection and annotation (the latter using \url{https://rectlabel.com/}.) Most recently the Detectron2 system \url{https://github.com/facebookresearch/detectron2} \cite{wu2019detectron2} builds on Mask R-CNN and offers many varied image task functionalities.

\paragraph{Scientific Examples:} Object detection has also gained significant attention across different scientific applications. It has been used in many medical settings to localize features of interest, for example, tumor cells across different imaging modalities \cite{li2018improved, zhang2016cancer} or fractures in radiology \cite{sa2017intervertebral, thian2019convolutional}. 

\subsubsection{Semantic Segmentation and Instance Segmentation}
Segmentation dives into the lowest possible level of detail --- categorizing every single image \textit{pixel}. In semantic segmentation, we want to categorize pixels according to the high level group they belong to. For example, suppose we are given an image of a street, with a road, different vehicles, pedestrians, etc. We would like to determine if a pixel is part of any pedestrian, part of any vehicle or part of the road --- i.e. label the image pixels as either pedestrian, vehicle or road. Instance segmentation is even more intricate, where not only do we want to categorize each pixel in this way, but do so separately for each instance (and provide instance specific bounding boxes like in object detection). The differences are illustrated in Figure \ref{fig:cnn-tasks} (sourced from \cite{abdulla2018segmentation}.) Returning to the example of the image of the street, suppose the image has three pedestrians. In semantic segmentation, all of the pixels making up these three pedestrians would fall under the same category -- pedestrian. In instance segmentation, these pixels would be further subdivided into those belonging to pedestrian one, pedestrian two or pedestrian three.

Because segmentation models must categorize every pixel, their output is not just a single class label, or a bounding box, but a full image. As a result, the neural network architectures for segmentation have a slightly different structure that helps them better preserve spatial information about the image. A highly popular and successful architecture, particularly for scientific applications, has been the \textit{U-net} \cite{ronneberger2015u}, which also has a 3d volumetric variant \cite{cciccek20163d}. Other architectures include FCNs (Fully Convolutional Networks) \cite{long2015fully}, SegNet \cite{badrinarayanan2017segnet} and the more recent Object Contextual Representations \cite{yuan2019object}. A couple of nice surveys on semantic segmentation methods are given by \href{https://towardsdatascience.com/semantic-segmentation-with-deep-learning-a-guide-and-code-e52fc8958823}{towardsdatascience Semantic Segementation with Deep Learning} and \url{https://sergioskar.github.io/Semantic_Segmentation/}.

For instance segmentation, Mask R-CNN \cite{he2017mask} and its variants \cite{wu2019detectron2} have been extremely popular. This tutorial \href{https://towardsdatascience.com/mask-r-cnn-for-ship-detection-segmentation-a1108b5a083}{Mask R-CNN tutorial with code} provides a step by step example application. The recent Detectron2 package \cite{wu2019detectron2} (\url{https://github.com/facebookresearch/detectron2}) also offers this functionality.

\paragraph{Scientific Examples:} Out of all of the different types of imaging prediction problems, segmentation methods have been especially useful for (bio)medical applications. Examples include segmenting brain MR images \cite{moeskops2016automatic, wachinger2018deepnat}, identifying key regions of cells in different tissues \cite{xu2016deep, song2014deep} and even studying bone structure \cite{liu2018deep}. 

\subsubsection{Super-Resolution}
Super resolution is a technique for transforming low resolution images to high resolution images. This problem has been tackled both using convolutional neural networks and supervised learning, as well as generative models. 

Super resolution formally defined is an underdetermined problem, as there may be many possible high resolution mappings for a low resolution image. Traditional techniques imposed constraints such as sparsity to find a solution. One of the first CNNs for super resolution, \textit{SRCNN} \cite{dong2015image} outlines the correspondences between sparse coding approaches and convolutional neural networks. More recently, \textit{Residual Dense Networks} \cite{zhang2018residual} have been a popular approach for super-resolution on standard benchmarks (with code available \url{https://github.com/yulunzhang/RDN}), as well as Predictive Filter Flow \cite{kong2018image}, (code: \url{https://github.com/aimerykong/predictive-filter-flow}) which has also looked at image denoising and deblurring. In some of the scientific applications below, \textit{U-nets} have also been successful for super resolution.

\paragraph{Scientific Examples:} Super resolution is arguably even more useful for scientific settings than standard natural image benchmarks. Two recent papers look at U-nets for super-resolution of fluorescence microscopy \cite{weigert2018content} (code: \url{https://csbdeep.bioimagecomputing.com/}) and electron microscopy \cite{fang2019deep}. Other examples include super resolution of chest CT scans \cite{umehara2018application} and Brain MRIs \cite{chen2018brain}. 

\subsubsection{Image Registration}
\label{sec-image-registration}
Image registration considers the problem of \textit{aligning} two input images to each other. Particularly relevant to scientific applications, the two input images might be from different imaging modalities (e.g. a 3D scan and a 2D image), or mapping a moving image to a canonical template image (such as in MRIs.) The alignment enables better identification and analysis of features of interest.

The potential of image registration is primarily demonstrated through different scientific applications. At the heart of the technique is a convolutional neural network, often with an encoder-decoder structure (similar to the U-net \cite{ronneberger2015u}) to guide the alignment of two images. Note that while this underlying model is trained through supervised learning, many registration methods \textit{do not require explicit labels}, using similarity functions and smoothness constraints to provide supervision. For example, \cite{balakrishnan2018unsupervised} develop an unsupervised method to perform alignment for Brain MRIs. The code for this and several followup papers \cite{balakrishnan2019voxelmorph, dalca2019learning} provides a helpful example for building off of and applying these methods \url{https://github.com/voxelmorph/voxelmorph}. Other useful resources include \url{https://github.com/ankurhanda/gvnn} (with corresponding paper \cite{handa2016gvnn}) a library for learning common parametric image transformations. 

\subsubsection{Pose Estimation}
\begin{figure}
\centering
\includegraphics[width=0.7\columnwidth]{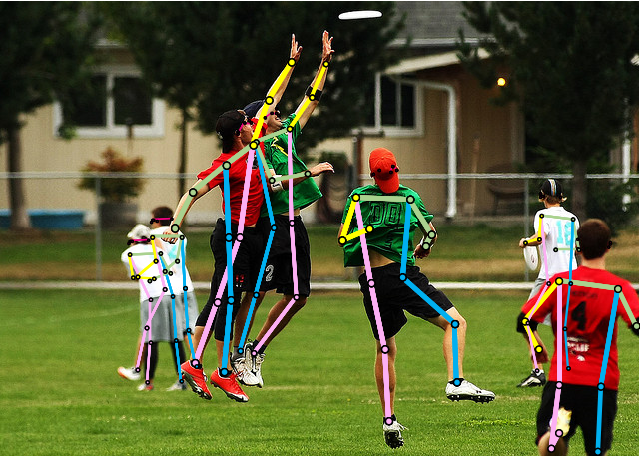} 
\caption{\small \textbf{Pose Estimation. Image source \cite{sun2019deep}} The task of pose estimation, specifically multi-person 2D (human) pose-estimation is depicted in the figure. The neural network model predicts the positions of the main joints (keypoints), which are combined with a body model to get the stick-figure like approximations of pose overlaid on the multiple humans in the image. Variants of these techniques have been used to study animal behaviors in scientific settings.}
\label{fig:pose-estimation}
\end{figure}
Pose estimation, and most popularly human pose estimation, studies the problem of predicting the pose of a human in a given image. In particular, a deep neural network model is trained to identify the location of the main joints, the \textit{keypoints} (e.g. knees, elbows, head) of the person in the image. These predictions are combined with existing \textit{body models} to get the full stick-figure-esque output summarizing the pose. (See Figure \ref{fig:pose-estimation}, sourced from \cite{sun2019deep}, for an illustration.)

(2D) Human pose estimation is a core problem in computer vision with multiple benchmark datasets, and has seen numerous convolutional architectures developed to tackle it. Some of the earlier models include a multi-stage neural network introduced by \cite{wei2016convolutional}, and a stacked hourglass model \cite{newell2016stacked} that alternatingly combines high and low resolutions of the intermediate representations. More recently, HRNet \cite{sun2019deep}, which keeps a high resolution representation throughout the model is a top performing architecture (code at \url{https://github.com/leoxiaobin/deep-high-resolution-net.pytorch}). Also of interest might be \cite{cao2018openpose} provides an end-to-end system for multiperson pose detection in the corresponding code repository \url{https://github.com/CMU-Perceptual-Computing-Lab/openpose}.

\paragraph{Scientific Examples:} Pose estimation has gained significant interest in neuroscience settings, where videos of animals are recorded, and automatically predicting poses in the image can help identify important behaviors. An example is given by \cite{mathis2018deeplabcut, mathis2020deep}, with associated code \url{http://www.mousemotorlab.org/deeplabcut}.

\subsubsection{Other Tasks with Convolutional Neural Networks}
In the preceding sections, we have overviewed some of the most common tasks for which convolutional neural networks are used. However, there are many additional use cases of these models that we have not covered, including \textit{video prediction} \cite{finn2016unsupervised}, \textit{action recognition} \cite{du2015hierarchical} and \textit{style transfer} \cite{gatys2016image}. We hope that the provided references and resources enable future investigation into some of these methods also.

\subsection{Graph Neural Networks}
Many datasets, such as (social) network data and chemical molecules have a \textit{graph} structure to them, consisting of vertices connected by edges. An active area of research, graph neural networks, has looked at developing deep learning methods to work well with this kind of data. The input graph consists of nodes $v$ having some associated feature vector $h_v$, and sometimes edges $e_{uv}$ also having associated features $z_{e_{uv}}$. For example, nodes $v$ might correspond to different atoms, and the edges $e_{uv}$ to the different kinds of chemical bonds between atoms. At a high level, most graph neural networks compute useful information from the data by (i) using the feature vectors of the \textit{neighbors} of each vertex $v$ to compute information on the input graph instance (ii) using this information to update the feature vector of $v$. This process, which respects the connectivity of the graph, is often applied iteratively, with the final output either at the vertex level (Are meaningful vertex feature vectors computed?) or at the level of the full input graph (Is some global property of the entire graph correctly identified?)

\paragraph{Application Characteristics} Problems where the data has an inherent graph structure, and the goal is to learn some function on this graph structure --- either at the per vertex level or a global property of the entire graph. There are also spatio-temporal graph neural networks --- performing predictions on graph structures evolving over time. 

\paragraph{Technical References} Although most graph neural networks follow the high level structure of aggregating information from vertex neighbors and using this information to update feature vectors, there are many many different architectural variants, with connections to other neural network models such as convolutional nets and recurrent models. Recent work has also looked at spatio-temporal graph networks for problems like action recognition in video \cite{li2018spatio}.  A nice unification of many of the first popular methods, such as \cite{duvenaud2015convolutional, battaglia2016interaction, li2015gated}, is given by \cite{gilmer2017neural}. A more recent survey paper \cite{wu2019comprehensive}, provides an \textit{extremely comprehensive} overview of the different kinds of architectures, problems, benchmark datasets and open source resources. Some useful code repositories include \url{https://github.com/rusty1s/pytorch_geometric}, \url{https://github.com/deepmind/graph_nets} and \url{https://github.com/dmlc/dgl}, which together cover most of the popular deep learning frameworks. 

\paragraph{Scientific Examples} Graph neural networks have been very popular for several chemistry tasks, such as predicting molecular properties \cite{duvenaud2015convolutional, hu2019pre, gilmer2017neural, kearnes2016molecular}, determining protein interfaces \cite{fout2017protein, townshend2019end} and even generating candidate molecules \cite{de2018molgan, bresson2019two}. A useful library for many of these chemistry tasks is \url{https://github.com/deepchem}, which also has an associated benchmark task \cite{wu2018moleculenet}. A detailed tutorial of different graph neural networks and their use in molecule generation can be seen at \url{https://www.youtube.com/watch?v=VXNjCAmb6Zw}.

\subsection{Neural Networks for Sequence Data}
A very common attribute for data is to have a \textit{sequential} structure. This might be frames in a video, amino acid sequences for a protein or words in a sentence. Developing neural network models to work with sequence data has been one of the most extensive areas of research in the past few years. A large fraction of this has been driven by progress on tasks in \textit{natural language processing}, which focuses on getting computers to work with the language used by people to communicate. Two popular tasks in this area, which have seen significant advances, have been \textit{machine translation} --- developing deep learning models to translate from one language to another and \textit{question answering} --- taking as input a (short) piece of text and answering a question about it. In the following sections, we first overview some of the main NLP tasks that have driven forward sequence modelling and then the neural network models designed to solve these tasks. 

\subsubsection{Language Modelling (Next Token Prediction)}
\label{sec-language-modelling}
Language modelling is a training method where the deep learning model takes as input the tokens of the sequence up to time/position $t$, and then uses these to predict token $t+1$. This is in fact a \textit{self-supervised} training method (see Section \ref{sec-self-supervised}), where the data provides a natural set of labels without additional labelling needed. In the NLP context, the neural network is fed in a sequence of words, corresponding to a sentence or passage of text, and it tries to predict the next word. For example, given a sentence, "The cat sat on the roof", the network would first be given as input "The" and asked to predict "cat", then be fed in "The cat" and asked to predict "sat", and so on. (There are some additional details in implementation, but this is the high level idea.) Because of the easy availability of data/labels, and the ability to use language modelling at different levels --- for words and even for characters, it has been a popular benchmark in natural language, and also for capturing sequence dependencies in scientific applications, such as protein function prediction \cite{hanson2016improving, heffernan2017capturing}, and using the hidden representations as part of a larger pipeline for protein structure prediction in AlphaFold \cite{senior2020improved} (with opensourced code \url{https://github.com/deepmind/deepmind-research/tree/master/alphafold_casp13}.)

\subsubsection{Sequence to Sequence}
\label{sec-seq-to-seq}
\begin{figure}
\centering
\includegraphics[width=0.8\columnwidth]{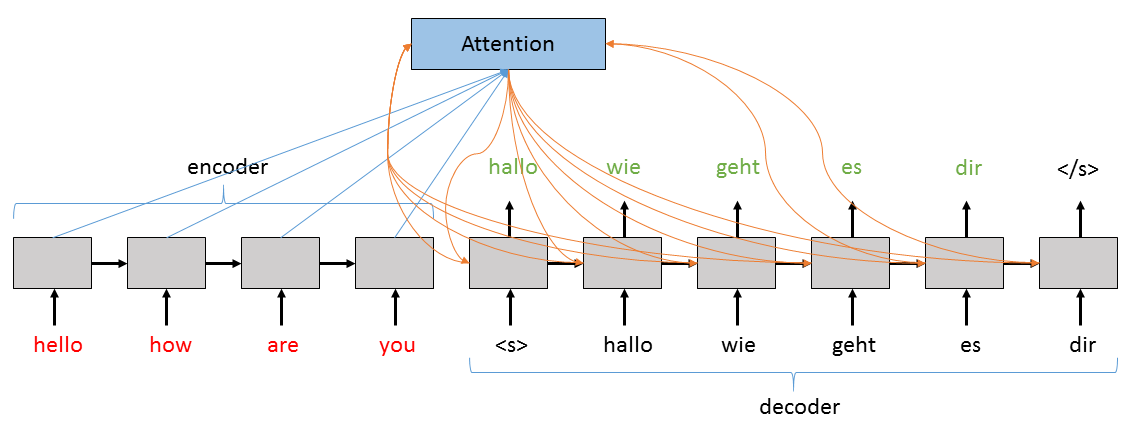} 
\caption{\small \textbf{Illustration of the Sequence to Sequence prediction task. Image source \cite{zhang2019dive}} The figure shows an illustration of a Sequence to Sequence task, translating an input sentence (sequence of tokens) in English to an output sentence in German. Note the encoder-decoder structure of the underlying neural network, with the encoder taking in the input, and the decoder generating the output, informed by the encoder representations and the previously generated output tokens. In this figure, the input tokens are fed in one by one, and the output is also generated one at a time, which is the paradigm when using \textit{Recurrent Neural Networks} as the underlying model. With \textit{Transformer} models, which are now extremely popular for sequence to sequence tasks, the sequence is input all at once, significantly speeding up use.}
\label{fig:seqtoseq}
\end{figure}
Another very popular task for sequence data is \textit{sequence to sequence} --- transforming one sequence to another. This is precisely the setup for machine translation, where the model gets an input sentence (sequence) in say English, and must translate it to German, which forms the output sentence (sequence). Some of the first papers framing this task and tackling it in this way are \cite{bahdanau2014neural, sutskever2014sequence, vinyals2015neural}. Sequence to sequence tasks typically rely on neural network models that have an \textit{encoder-decoder} structure, with the encoder neural network taking in the input sequence and learning to extract the important features, which is then used by the decoder neural network to produce the target output. Figure \ref{fig:seqtoseq}(sourced from \cite{zhang2019dive}) shows an example of this. This paradigm has also found some scientific applications as varied as biology \cite{cao2017prolango} and energy forcasting \cite{marino2016building}. Sequence to sequence models critically rely on a technique called \textit{attention}, which we overview below. For more details on this task, we recommend looking at some of the tutorials and course notes highlighted in Section \ref{sec-frameworks-resources}. 

\subsubsection{Question Answering}
One other popular benchmark for sequence data has been question answering. Here, a neural network model is given a paragraph of text (as context) and a specific question to answer on this context as input. It must then output the part of the paragraph that answers the question. Some of the standard benchmarks for this task are \cite{hermann2015teaching, rajpurkar2016squad}, with \url{http://web.stanford.edu/class/cs224n/slides/cs224n-2019-lecture10-QA.pdf} providing an excellent overview of the tasks and common methodologies. Question answering critically relies on the neural network model understanding the relevance and similarity of different sets of sequences (e.g. how relevant is this part of the context to the question of interest?). This general capability (with appropriate reformulation) has the potential to be broadly useful, both for determining similarity and relevance on other datasets, and for question answering in specialized domains \cite{galko2018biomedical}.

\subsubsection{Recurrent Neural Networks}
\begin{figure}
\centering
\includegraphics[width=0.7\columnwidth]{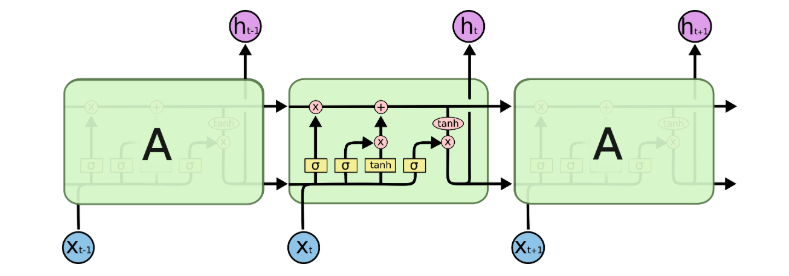} 
\caption{\small \textbf{Diagram of a Recurrent Neural Network model, specifically a LSTM (Long-Short Term Network). Image source \cite{olah2015understandinglstms}} The figure illustrates an LSTM network, a type of Recurrent Neural Network. We see that the input $x_t$ at each timestep also inform the internal network state in the next timestep (hence a recurrent neural network) through a \textit{gating mechanism}. This gating mechanism is called an LSTM, and consists of sigmoid and tanh functions, which transform and recombine the input for an updated internal state, and also emit an output. The mechanics of this gating process are shown in the middle cell of the figure.}
\label{fig:lstm}
\end{figure}
Having seen some of the core tasks in deep learning for sequence data, these next few sections look at some of the key neural network models.

Recurrent neural networks (RNNs) were the first kind of deep learning model successfully used on many of the aforementioned tasks. Their distinguishing feature, compared to CNNs or MLPs (which are \textit{feedforward} neural networks, mapping input straight to output), is that there are \textit{feedback connections}, enabling e.g. the output at each timestep to become the input for the next timestep, and the preservation and modification of an internal state across timesteps. When RNNs are used for sequential data tasks, sequences are input token by token, with each token causing an update of the internal \textit{cell state} of the RNN, and also making the RNN emit a token output. Note that this enables these models to work with \textit{variable length} data --- often a defining characteristic of sequence data. How the input is processed, cell state updated and output emitted are controlled by \textit{gating functions} --- see the technical references!

\paragraph{Application Characteristics:} Problems where the data has a sequential nature (with different sequences of varying length), and prediction problems such as determining the next sequence token, transforming one sequence to another, or determining sequence similarities are important tasks.

\paragraph{Technical References:} Research on sequence models and RNNs has evolved dramatically in just the past couple of years. The most successful and popular kind of RNN is a \textit{bi-LSTM with Attention}, where LSTM (Long-Short Term Memory) \cite{hochreiter1997long} refers to the kind of gating function that controls updates in the network, bi refers to bidirectional (the neural network is run forwards \textit{and} backwards on the sequence) and Attention is a very important technique that we overview separately below. (Some example papers \cite{merityRegOpt, merityAnalysis} and code resources \url{https://github.com/salesforce/awd-lstm-lm}.) This excellent post \url{https://colah.github.io/posts/2015-08-Understanding-LSTMs/} provides a great overview of RNNs and LSTMs in detail. (Figure \ref{fig:lstm} shows a diagram from the post revealing the details of the gating mechanisms in LSTMs.) The post also describes a small variant of LSTMs, Gated Recurrent Units (GRUs) which are also popular in practice \cite{li2015gated}. While RNNs (really bi-LSTMs) have been very successful, they are often tricky to develop and train, due to their recursiveness presenting challenges with optimization (the vanishing/exploding gradients problem \cite{hochreiter1998vanishing, pascanu2012understanding, hanin2018neural}), with performing fast model training (due to generating targets token by token), and challenges learning long term sequential dependencies. A new type of feedforward neural network architecture, the \textit{Transformer} (overviewed below), was proposed to alleviate the first two of these challenges. 

\paragraph{Scientific Examples:} RNNs have found several scientific applications for data with sequential structure, such as in genomics and proteomics \cite{pollastri2002improving, liu2017deep, kojima2019recurrent}. 

\subsubsection{Attention}
A significant problem in using RNNs and working with sequential data is the difficulty in capturing \textit{long range dependencies}. Long range dependencies are when tokens in the sequence that are very far apart from each other must be processed together to inform the correct output. RNNs process sequences in order, token by token, which means they must remember all of the important information from the earlier tokens until much later in the sequence --- very challenging as the memory of these architectures is far from perfect. \textit{Attention} \cite{chorowski2015attention,bahdanau2016end} is a very important technique that introduces \textit{shortcut connections} to earlier tokens, which alleviates the necessity to remember important features for the duration of the entire sequence. Instead it provides a direct way to model long term dependencies --- the neural network has the ability to \textit{look back} and \textit{attend} to what it deems relevant information (through learning) earlier in the input. A very nice overview of attention is provided by \url{https://lilianweng.github.io/lil-log/2018/06/24/attention-attention.html}. A variant of attention, \textit{self-attention}, which can be used to help predictions on a single input sequence, is the core building block of Transformer models. 

\subsubsection{Transformers}
\label{sec-transformers}
\begin{figure}
\centering
\includegraphics[width=0.7\columnwidth]{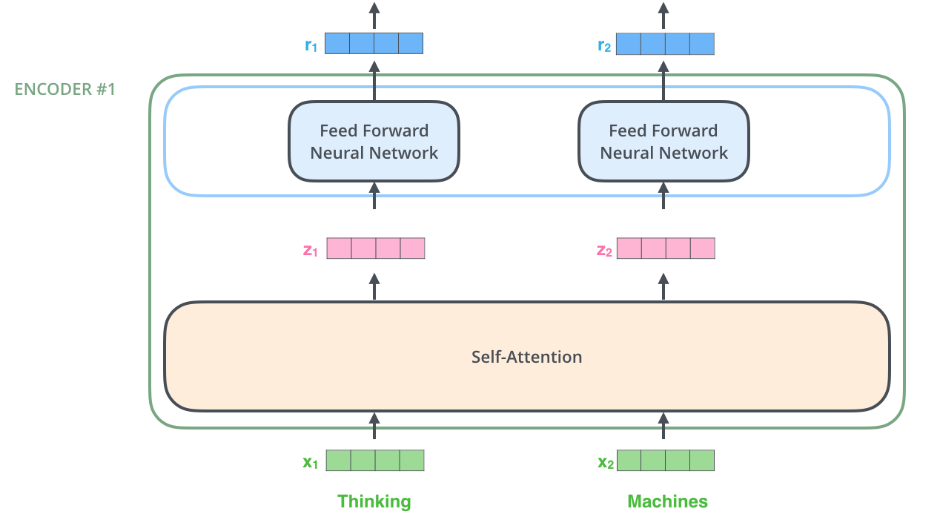} 
\caption{\small \textbf{Image of a couple of layers from a Transformer network. Image source \cite{alammar2018transformer}} The figure depicts the core sequence of layers that are fundamental to Transformer neural networks, a \textit{self-attention} layer (sometimes called a self-attention head) followed by fully connected layers. Note that when working with sequence data, transformers take the entire input sequence all at once, along with positional information (in this case the input sequence being "Thinking Machines".)}
\label{fig:transformer}
\end{figure}
While attention helped with challenges in long range dependencies, RNNs still remained slow to train and tricky to design (due to optimization challenges with vanishing/exploding gradients.) These challenges were inherent to their recurrent, token-by-token nature, prompting the proposal of a new \textit{feedforward} neural network to work with sequential data, the Transformer \cite{vaswani2017attention}, which critically relies on attentional mechanisms (the paper is in fact titled \textit{Attention is All you Need}.) During training transformers take in the \textit{entire} sequence as input all at once, but have \textit{positional embeddings} that respects the sequential nature of the data. Transformers have been exceptionally popular, becoming the dominant approach to many natural language tasks and sequential tasks.

\paragraph{Application Characteristics:} Problems where the data has a sequential nature and long range dependencies that need to be modelled. Given the large number of pretrained transformer models, they can also be very useful in settings where pretrained models on standard benchmarks can be quickly adapted to the target problem.

\paragraph{Technical References:} The original transformer paper \cite{vaswani2017attention} provides a nice overview of the motivations and the neural network architecture. The model was designed with machine translation tasks in mind, and so consists of an \textit{encoder} neural network and a \textit{decoder} neural network. With transformers being adopted for tasks very different to machine translation, the encoder and decoder are often used in stand-alone fashions for different tasks --- for example, the encoder alone is used for question answering, while the decoder is important for text generation. Two very accessible step by step tutorials on the transformer are \href{https://nlp.seas.harvard.edu/2018/04/03/attention.html}{The Annotated Transformer} and \href{http://jalammar.github.io/illustrated-transformer/}{The Illustrated Transformer}. A nice example of some of the language modelling capabilities of this models is given by \cite{radford2019language}.

Since the development of the transformer, there has been considerable research looking at improving the training of these models, adjusting the self-attention mechanism and other variants. A very important result using the transformer has been BERT (Pretraining of deep Bi-directional Transformers for Language understanding) \cite{devlin2018bert}. This paper demonstrates that performing \textit{transfer learning} (see Section \ref{sec-transfer-learning}) using a transformer neural network can be extremely successful for many natural language tasks. (Some of the first papers showing the potential of transfer learning in this area were \cite{howard2018universal, radford2019language}, and since BERT, there have been followups which extend the model capabilities \cite{yang2019xlnet}.) From a practical perspective, the development of transformers, BERT and transfer learning mean that there are many resources available online for getting hold of code and pretrained models. We refer to some of these in Section \ref{sec-frameworks-resources}, but of particular note is \url{https://github.com/huggingface/transformers} which has an excellent library for transformer models. A good overview of BERT and transfer learning in NLP is given in \url{http://jalammar.github.io/illustrated-bert/}.

\paragraph{Scientific Examples:} There have been several interesting examples of transformers used in scientific settings, such as training on protein sequences to find representations encoding meaningful biological properties \cite{rives2019biological}, protein generation via language modelling \cite{madani2020progen}, bioBERT \cite{lee2019biobert} for text mining in biomedical data (with \href{https://github.com/naver/biobert-pretrained}{pretrained model} and \href{https://github.com/dmis-lab/biobert}{training code}), embeddings of scientific text \cite{beltagy2019scibert} (with code \url{https://github.com/allenai/scibert}) and medical question answering \cite{wang2019medical}.

\subsubsection{Other Tasks with Sequence Data}
In the previous sections, we've given an overview of some of the important benchmark tasks for sequential data, and the types of deep learning models available to tackle them. As with convolutional networks, this is not a comprehensive overview, but hopefully thorough enough to help with generating ideas on possible applications and offering pointers to other useful related areas. A few other sequential data tasks that might be of interest are \textit{structured prediction}, where the predicted output has some kind of structure, from tree structures (in e.g. parsing) \cite{chen2014fast, weiss2015structured} to short, executable computer program structure \cite{zhong2017seq2sql} and \textit{summarization}, where passages of text are summarized by a neural network \cite{liu2018generating, zhou2018neural}. We'll also discuss \textit{word embeddings} later in the survey.

\subsection{Section Summary}
In this section, we have overviewed supervised learning, some of the core neural network models and the kinds of important tasks they can be used for. As previously discussed, these topics span an extremely large area of research, so there are some areas, e.g. deep neural networks for set structured data \cite{zaheer2017deep, komiske2019energy}, modelling different invariances --- invariances to specified Lie groups for application to molecular property prediction \cite{finzi2020generalizing}, spherical invariances \cite{cohen2016group, cohen2018spherical} not covered. But we hope the material and references presented help inspire novel contributions to these very exciting and rapidly evolving research directions.

\section{Key (Supervised Learning) Methods}
\label{sec-supervised-methods}
In the previous section we saw different kinds of neural network models, and the many different types of tasks they could be used for. To train the models for these tasks, we typically rely on the supervised learning methodology --- optimize model parameters to correctly output given labels (the supervision) on a set of training data examples. 

In more detail, the standard supervised learning method for deep neural networks consists of (i) collecting data instances (e.g. images) (ii) collecting \textit{labels} for the data instances (e.g. is the image a cat or a dog) (iii) splitting the set of collected (data instance, label) into a training set, validation set and test set (iv) \textit{randomly initializing} neural network parameters (iv) optimizing parameters so the network outputs the correct corresponding label given an input data instance on the training set (v) further tuning and validating on the validation and test sets.

In this section we overview methods that use variants of this process, for example initializing the neural network parameters differently or dealing with shifts between the training data and the test sets. In Section \ref{sec-less-data}, we look at variants that reduce the dependence on collecting labels.

\subsection{Transfer Learning}
\label{sec-transfer-learning}
\begin{figure}
\centering
\includegraphics[width=0.9\columnwidth]{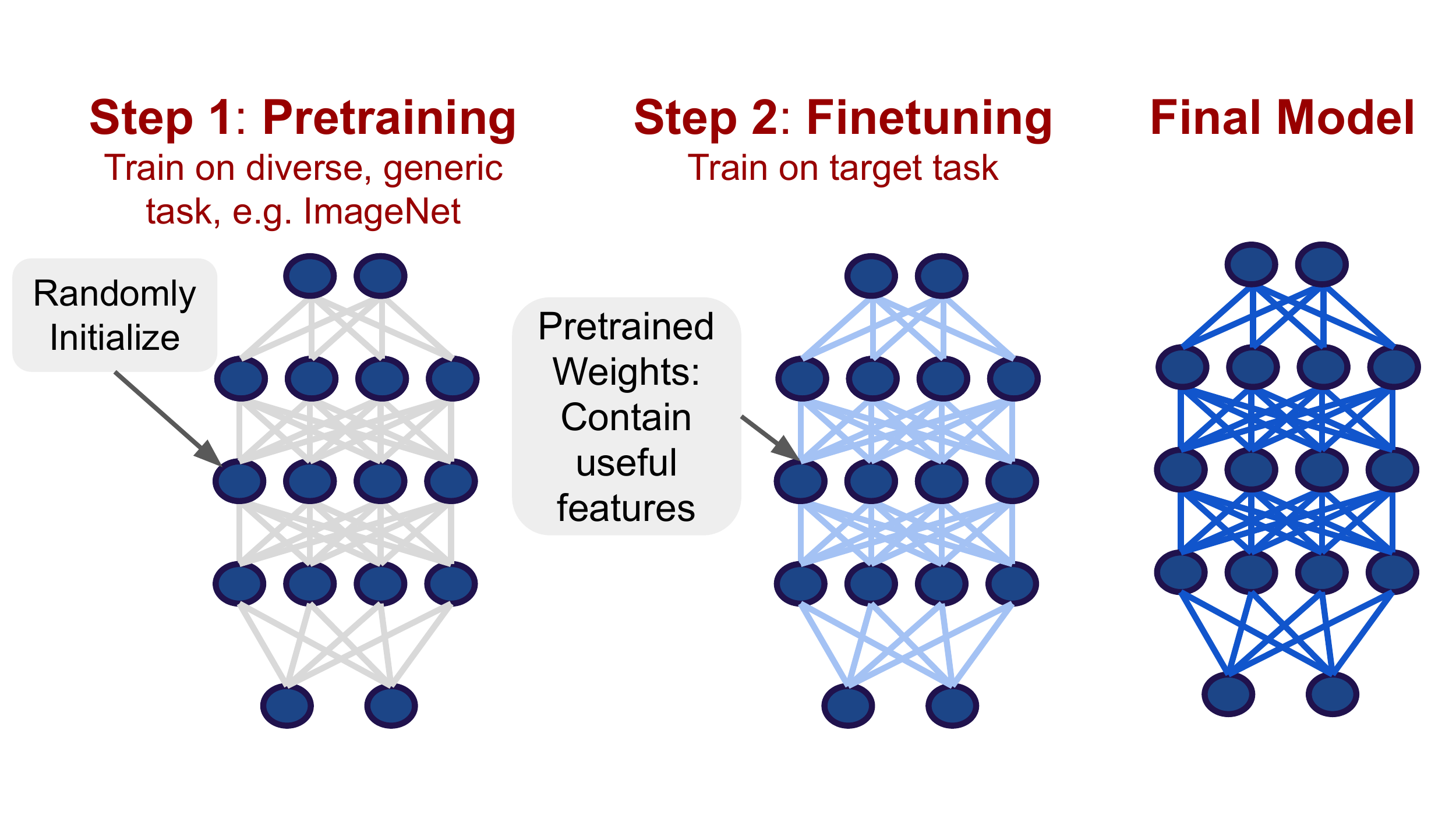} 
\caption{\small \textbf{The Transfer Learning process for deep neural networks.} Transfer learning is a two step process for training a deep neural network. Instead of intializing parameters randomly and directly training on the target task, we first perform a \textit{pretraining} step, on some diverse, generic task. This results in the neural network parameters converging to a set of values, known as the \textit{pretrained weights}. If the pretraining task is diverse enough, these pretrained weights will contain useful features that can be leveraged to learn the target task more efficiently. Starting from the pretrained weights, we then train the network on the target task, known as \textit{finetuning}, giving us the final model.}
\label{fig:transfer-learning}
\end{figure}
Through the preceding sections, we've made references to using \textit{pretrained} models. This is in fact referring to a very important method for training deep neural networks, known as \textit{transfer learning}. Transfer learning is a two step process for training a deep neural network model, a \textit{pretraining} step, followed by a \textit{finetuning} step, where the model in trained on the target task. More specifically, we take a neural network with parameters randomly initialized, and first train it on a standard, generic task --- the pretraining step. For example, in image based tasks, a common pretraining task is ImageNet \cite{deng2009imagenet}, which is an \textit{image classification} task on a large dataset of natural images. With an appropriate pretraining task that is generic and complex enough, the pretraining step allows the neural network to learn useful features, stored in its parameters, which can then be reused for the second step, \textit{finetuning}. In finetuning, the pretrained neural network is further trained (with maybe some minor modifications to its output layer) on the \textit{true target task} of interest. This process is illustrated in Figure \ref{fig:transfer-learning}. But being able to use the features it learned during pretraining often leads to boosts in performance and convergence speed of the target task, as well as needing less labelled data.

Because of these considerable benefits, transfer learning has been extraordinarily useful in many settings, particularly in computer vision \cite{huh2016makes}, which had many early successful applications. As overviewed in Section \ref{sec-transformers}, the recent development of models like ULMFiT \cite{howard2018universal} and especially BERT \cite{devlin2018bert} has also made transfer learning extremely successful in natural language and sequential data settings, with recent work making the transfer learning process even more efficient \cite{houlsby2019parameter, sanh2019distilbert}. Most importantly, the ready availability of standard neural network architectures pretrained on standard benchmarks through many open sourced code repositories on GitHub (examples given in Section \ref{sec-frameworks-resources}) has meant that downloading and finetuning a standard pretrained model has become the \textit{de-facto standard} for most new deep learning applications.

Typically, performing transfer learning is an \textit{excellent} way to start work on a new problem of interest. There is the benefit of using a well-tested, standard neural network architecture, aside from the knowledge reuse, stability and convergence boosts offered by pretrained weights. Note however that the precise effects of transfer learning are not yet fully understood, and an active research area \cite{kornblith2019better, raghu2019transfusion, zhai2019visual, ngiam2018domain, mahajan2018exploring, raffel2019exploring, voita2019bottom} looks at investigating its exact properties. For transfer learning in vision \cite{kornblith2019better, zhai2019visual, kolesnikov2019revisiting} may be of particular interest for their large scale studies and pretraining recommendations. 

\subsection{Domain Adaptation}
Related to transfer learning is the task of \textit{domain adaptation}. In (unsupervised) domain adaptation, we have training data and labels in a \textit{source} domain, but want to develop a deep learning model that will also work on a \textit{target} domain, where the data instances may look different to those in the source domain, but the high level task is the same. For instance, our source domain many consist of images of handwritten digits (zero to nine) which we wish to classify as the correct number. But the target domain many have photographs of house numbers (from zero to nine), that we also wish to classify as the correct number. Domain adaptation techniques help build a model on the source domain that can also work (reasonably) well out-of-the-box on the shifted target domain.

The most dominant approach to domain adaptation in deep learning is to build a model that can (i) perform well on the source domain task, and (ii) learns features that are as invariant to the domain shift as possible. This is achieved through \textit{jointly optimizing} for both of these goals. Returning to our example on handwritten digits and house number photographs, (i) corresponds to the standard supervised learning classification problem of doing well on the (source) task of identifying handwritten digits correctly while (ii) is more subtle, and typically involves explicitly optimizing for the \textit{hidden layer representations} of handwritten digits and house number photographs to look the same as each other --- domain invariance. Some popular ways to implement this include \textit{gradient reversal} \cite{ganin2014unsupervised}, minimizing a distance function on the hidden representations \cite{long2017deep}, and even \textit{adversarial training} \cite{ganin2016domain, shu2018dirt}. More recently, \cite{sun2019unsupervised} look at using \textit{self-supervision} (see Section \ref{sec-self-supervised}) to jointly train on the source and target domains, enabling better adaptation.

Other approaches to domain adaptation include translating data instances from the source to the target domain, and bootstrapping/co-training approaches (see Section \ref{sec-semi-supervised}). Some of these methods are overviewed in tutorials such as \href{https://towardsdatascience.com/deep-domain-adaptation-in-computer-vision-8da398d3167f}{Deep Domain Adaptation in Computer Vision}. 

\subsection{Multitask Learning}
\label{sec-multitask}
In many supervised learning applications, ranging from machine translation \cite{aharoni2019massively} to scientific settings \cite{ramsundar2015massively, poplin2018prediction}, neural networks are trained in a \textit{multitask} way -- predicting several different outputs for a single input. For example, in image classification, given an input medical image, we might train the network not only to predict a disease of interest, but patient age, history of other related disease, etc. This often has beneficial effects even if there is only one prediction of interest, as it provides the neural network with useful additional feedback that can guide it in learning the most important data features. (This can be so useful that sometimes auxiliary prediction targets are defined solely for this purpose.) Additionally, the prediction of multiple targets can mean that more data is available to train the model (only a subset of the data has the target labels of interest, but many more data instances have other auxiliary labels.) The most extreme version of this is to \textit{simultaneously train} on two entirely different datasets. For example, instead of performing a pretraining/finetuing step, the model could be trained on both ImageNet and a medical imaging dataset at the same time.

Multitask learning is usually implemented in practice by giving the neural network multiple \textit{heads}. The head of a neural network refers to its output layer, and a neural network with multiple heads has one head for each predictive task (e.g. one head for predicting age, one for predicting the disease of interest) but \textit{shares} all of the other features and parameters, across these different predictive tasks. This is where the benefit of multitask learning comes from --- the shared features, which comprise of most of the network, get many different sources of feedback. Implementing multitask learning often also requires careful choice of the way to weight the training objectives for these different tasks. A nice survey of some popular methods for multitask learning is given by \url{https://ruder.io/multi-task/index.html#fn4}, and a tutorial on some of the important considerations in \url{http://hazyresearch.stanford.edu/multi-task-learning}. One package for implementing multitask learning is found in \url{https://github.com/SenWu/emmental} and step-by-step example with code excerpts in \href{https://towardsdatascience.com/multitask-learning-teach-your-ai-more-to-make-it-better-dde116c2cd40}{towardsdatascience Multitask Learning: teach your AI more to make it better}.

\subsection{Weak Supervision (Distant Supervision)}
Suppose it is very difficult to collect high quality labels for the target task of interest, and neither is there an existing, standard, related dataset and corresponding pretrained model to perform transfer learning from. How might one provide the deep learning model with enough supervision during the training process? While high quality labels might be hard to obtain, \textit{noisy labels} might be relatively easy to collect. \textit{Weak supervision} refers to the method of training a model on a dataset with these noisy labels (typically for future finetuning), where the noisy labels are often generated in an \textit{automatic process}. 

In computer vision (image based) tasks, some examples are: taking an image level label (for classification) and automatically inferring pixel level labels for segmentation \cite{pathak2015constrained}, clustering hidden representations computed by a pretrained network as pseudo-labels \cite{yan2019clusterfit}, or taking Instagram tags as labels \cite{mahajan2018exploring} for pretraining. In language tasks, examples are given by \cite{mintz2009distant, hoffmann2011knowledge, zeng2015distant}, which provide noisy supervision by assuming all sentences mentioning two entities of interest express a particular relation (also known as \textit{distant supervision}). A nice overview of weak supervision and its connection to other areas is given in \url{https://hazyresearch.github.io/snorkel/blog/ws_blog_post.html}, with a related post looking specifically at medical and scientific applications \url{http://hazyresearch.stanford.edu/ws4science}.

\subsection{Section Summary}
In this section, we have overviewed some of the central supervised learning based methodologies for developing deep learning models. This is just a sampling of the broad collection of existing methods, and again, we hope that the descriptions and references will help facilitate further exploration of other approaches. One method not covered that might be of particular interest is \textit{multimodal learning}, where neural networks are simultaneously trained on data from different modalities, such as images and text \cite{lu2018multimodal, wang2018ajile, kawahara2018seven}. Multimodal learning also provides a good example of the fact that it is often difficult to precisely categorize deep learning techniques as only being useful for a specific task or training regime. For example, we looked at language modelling for sequence tasks in this supervised learning section, but language modelling is also an example of self-supervision (Section \ref{sec-self-supervised}) and generative models (Section \ref{sec-generative-models}). There are many rich combinations of the outlined methods in both this section and subsequent sections, which can prove very useful in the development of an end to end system.

\section{Doing More with Less Data}
\label{sec-less-data}
\label{sec-self-supervised}
\begin{figure}
\centering
\includegraphics[width=0.9\columnwidth]{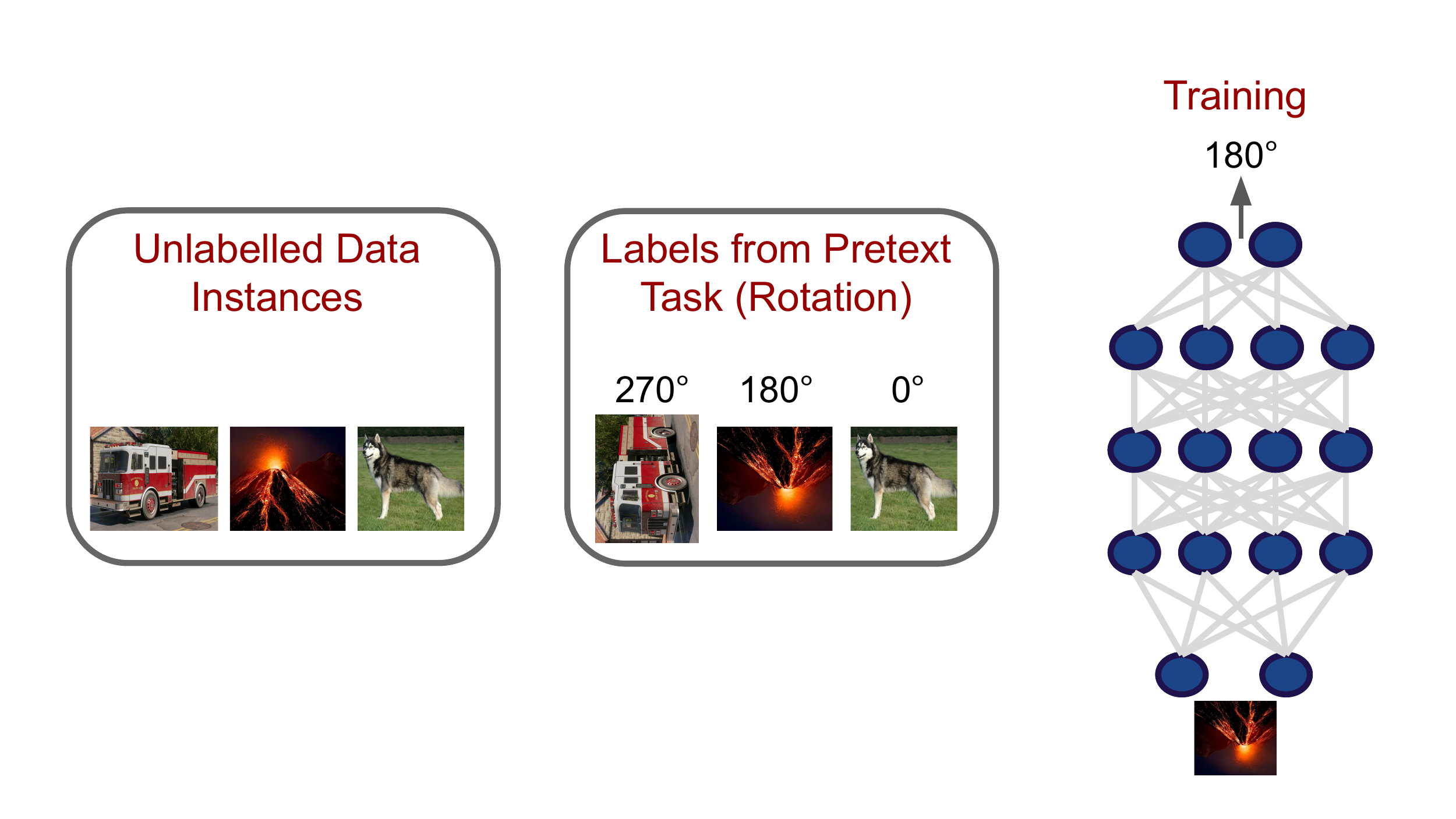} 
\caption{\small \textbf{Training neural networks with Self-Supervision.} The figure illustrates one example of a self-supervision setup. In self-supervision, we typically have a collection of unlabelled data instances, in this case images. We define a \textit{pretext task}, that will automatically generate labels for the data instances. In this case, the pretext task is rotation --- we randomly rotate the images by some amount and label them by the degree of rotation. During training, the neural network is given this rotated image and must predict the degree of rotation. Doing so also requires the neural network learn useful hidden representations of the image data in general, so after training with self-supervision, this neural network can then be successfully and efficiently finetuned on a downstream task.}
\label{fig:self-supervised}
\end{figure}
Supervised learning methods, and specific variants such as transfer learning and multitask learning have been highly successful in training deep neural network models. However, a significant limitation to their use, and thus the use of deep learning, is the dependence on large amounts of \textit{labelled} data. In many specialized domains, such as medicine, collecting a large number of high quality, reliable labels can be prohibitively expensive. 

Luckily, in just the past few years, we've seen remarkable advances in methods that reduce this dependence, particularly \textit{self-supervision} and \textit{semi-supervised learning}. These approaches still follow the paradigm of training a neural network to map raw data instances to a specified label, but critically, these labels are not collected separately, but \textit{automatically defined} via a pretext task. For example, we might take a dataset of images, rotate some of them, and then define the label as the degree of rotation, which is the prediction target for the neural network. This enables the use of \textit{unlabelled data} in training the deep neural network. In this section, we cover both self-supervision and semi-supervised learning as well as other methods such as data augmentation and denoising, all of which enable us to do more with less data.

\subsection{Self-Supervised Learning}
In self-supervision, a \textit{pretext task} is defined such that labels can be \textit{automatically} calculated directly from the raw data instances. For example, on images, we could rotate the image by some amount, label it by how much it was rotated, and train a neural network to predict the degree of rotation \cite{gidaris2018unsupervised} --- this setup is illustrated in Figure \ref{fig:self-supervised}. This pretext task is defined without needing any labelling effort, but can be used to teach the network good representations. These representations can then be used as is or maybe with a little additional data for downstream problems. Arguably the biggest success of self-supervision has been \textit{language modelling} for sequential data and specifically natural language problems, which we overviewed in Section \ref{sec-language-modelling}. Below we outline some of the most popular and successful self-supervision examples for both image and sequential data. (A comprehensive list of self-supervision methods can also be found on this page \url{https://github.com/jason718/awesome-self-supervised-learning}.)  

\subsubsection{Self-Supervised Learning for Images}
A recent, popular and simple self-supervised task for images is to predict image rotations \cite{gidaris2018unsupervised}. Each image instance is transformed with one of four possible rotations and the deep learning model must classify the rotation correctly. Despite its simplicity, multiple studies have shown its success in learning good representations \cite{zhai2019visual, zhai2019s, kolesnikov2019revisiting}. Another popular method examined in those studies is \textit{exemplar} \cite{dosovitskiy2014discriminative}, which proposes a self-supervision task relying on invariance to \textit{image transformations}. For example, we might take a source image of a cat, and perform a sequence of transformations, such as rotation, adjusting contrast, flipping the image horizontally, etc. We get multiple images of the cat by choosing many such sequences, and train the neural network to recognize these all as the same image.

Other methods look at using image patches as \textit{context} to learn about the global image structure and important features. For example, \cite{doersch2015unsupervised} defines a pretext task where the relative locations of pairs of image patches must be determined, while \cite{noroozi2016unsupervised} teaches a neural network to solve jigsaw puzzles. This latter task has been shown to be effective at large scales \cite{goyal2019scaling}, with nice implementations and benchmarking provided by \url{https://github.com/facebookresearch/fair_self_supervision_benchmark}. A recent line of work has looked at using \textit{mutual information} inspired metrics as a way to provide supervision on the relatedness of different image patches \cite{hjelm2018learning, oord2018representation, bachman2019learning, misra2019selfsupervised}, but these may be more intricate to implement. Many of these mutual information based metrics also rely on \textit{contrastive losses} \cite{chen2017sampling}, which, at a high level, provides supervision to the network by making representations of a pair of similar inputs more similar than representations of a pair of different inputs. Very recently, a new self-supervision method, SimCLR \cite{chen2020simple}, uses this to achieve high performance (one implementation at \url{https://github.com/sthalles/SimCLR}.)

Note that some of the \textit{image registration} examples given in Section \ref{sec-image-registration} are also examples of self-supervised learning, where some kind of domain specific similarity function can be automatically computed to assess the quality of the output. Such approaches may be relevant to other domains, and are useful to explore. A great set of open-sourced implementations of many of self-supervision methods is provided by \url{https://github.com/google/revisiting-self-supervised}.

\subsubsection{Self-Supervised Learning for Sequential (Natural Language) Data}
While research on self-supervision techniques for images has been extremely active, the strongest successes of this framework have arguably been with sequential data, particularly text and natural language. The sequential structure immediately gives rise to effective self-supervision pretext tasks. Two dominant classes of pretext tasks operate by either (i) using neighboring tokens of the sequence as input \textit{context} for predicting a target token (ii) taking in all tokens up to a particular position and predicting the next token. The latter of these is \textit{language modelling}, which was overviewed in Section \ref{sec-language-modelling}. The former is the principle behind \textit{word embeddings}.

Word embeddings have been critical to solving many natural language problems. Before the recent successes of full fledged transfer learning in language (Section \ref{sec-transfer-learning}) this simple self-supervised paradigm was where knowledge reuse was concentrated, and formed a highly important component of any deep learning system for natural language (sequential) data. From a scientific perspective, learning word embeddings for sequential data has the potential to identify previously unknown similarities in the data instances. It has already found interesting uses in aiding with the automatic analysis of scientific texts, such as drug name recognition systems \cite{liu2015effects}, biomedical named entity recognition \cite{habibi2017deep}, identifying important concepts in materials science \cite{tshitoyan2019unsupervised} and even detecting chemical-protein interactions \cite{corbett2018improving}.  

The key fundamental ideas of word embeddings are captured in the \textit{word2vec} framework \cite{mikolov2013distributed, mikolov2013efficient}, the original framework relying on either a \textit{Continuous-Bag-of-Words} (CBOW) neural network or a \textit{Skip-Gram} neural network. Actually, both of these models are less neural networks and more two simple matrix multiplications, with the first matrix acting as a \textit{projection}, and giving the desired embedding. In CBOW, the context --- defined as the neighborhood words --- are input, and the model must correctly identify the target output word. In Skip-Gram, this is reversed, with the center word being input, and the context being predicted. For example, given a sentence "There is a cat on the roof", with the target word being cat, CBOW would take in the vector representations of (There, is, a, on, the, roof) and output "cat", while Skip-Gram would roughly swap the inputs and outputs. The simplicity of these methods may make them more suitable for many tasks compared to language modelling. Two nice overviews of the these methods are given by \href{https://towardsdatascience.com/introduction-to-word-embedding-and-word2vec-652d0c2060fa}{Introduction to Word Embeddings and word2vec}, and \url{https://ruder.io/word-embeddings-1/}. Other embedding methods include \cite{pennington2014glove, levy2014neural}. 

\subsubsection{Self-Supervision Summary}
In this section we have outlined many of the interesting developments in self-supervised learning, a very successful way to make use of unlabelled data to learn meaningful representations, either for analysis or other downstream tasks. Self-supervision can be effectively used along with other techniques. For example, in the language modelling application, we saw it used for transfer learning (Section \ref{sec-transfer-learning}), where a deep learning model is first pretrained using the language modelling self supervision objective, and then finetuned on the target task of interest. In the following section, we will other ways of combining self-supervision with labelled data.

\subsection{Semi-Supervised Learning}
\label{sec-semi-supervised}
While collecting \textit{large} labelled datasets can be prohibitively expensive, it is often possible to collect a smaller amount of labelled data. When assembling a brand new dataset, a typical situation is having a small amount of labelled data and a (sometimes significantly) larger number of data instances with no labels. \textit{Semi-supervised learning} looks at precisely this setting, proposing techniques that enable effective learning on labelled and unlabelled data. Below we overview some of the popular methods for semi-supervised learning.

\subsubsection{Self-Supervision with Semi-Supervised Learning}
Following on from the previous section, one natural way to make use of the unlabelled data is to use a self-supervised pretext task. To combine this with the labelled data, we can design a neural network that has \textit{two different outputs heads} (exactly as in multitask learning, see Section \ref{sec-multitask}), with one output head being used for the labelled data, and the other for the self-supervised objective on the unlabelled data. Importantly, this means that the features learned by the neural network are \textit{shared} between the labelled \textit{and} unlabelled data, leading to better representations. This simple approach has been shown to be very effective \cite{zhai2019visual, zhai2019s}.

\subsubsection{Self-Training (Bootstrapping)}
Self-training, sometimes also referred to as \textit{bootstrapping} or \textit{pseudo-labels}, is an iterative method where a deep neural network is first developed in a supervised fashion on the labelled data. This neural network is then used to provide (pseudo) labels to the unlabelled data, which can then be used in conjunction with the labelled data to train a new, more accurate neural network. This approach often works well and can even be repeated to get further improvements. There are a couple of common details in implementation --- often when adding the neural network pseudo-labelled data, we only keep the most \textit{confidently} pseudo-labelled examples. These pseudo-labelled examples may also be used for training with a different objective function compared to the labelled data. One of the early papers proposing this method was \cite{lee2013pseudo}, with a more recent paper \cite{xie2019self} demonstrating significant successes at large scale. Other variants, including \textit{mean teacher} \cite{tarvainen2017mean}, \textit{temporal ensembling} \cite{laine2016temporal} and the recent \textit{MixMatch} \cite{berthelot2019mixmatch} also primarily use the self-training approach, but incorporate elements of \textit{consistency} (see below). There are nice open sourced implementations of these methods, such as \url{https://github.com/CuriousAI/mean-teacher} for mean teacher and \url{https://github.com/google-research/mixmatch} and \url{https://github.com/YU1ut/MixMatch-pytorch} for MixMatch.

\subsubsection{Enforcing Consistency (Smoothness)}
An important theme in many semi-supervised methods has been to provide supervision on the unlabelled data through enforcing \textit{consistency}. If a human was given two images A and B, where B was a slightly perturbed version of A (maybe blurred, maybe some pixels obscured or blacked out), they would give these images the same label --- consistency. We can also apply this principle to provide feedback to our neural network on the unlabelled data, combining it with the labelled data predictions as in multitask learning (Section \ref{sec-multitask}) to form a semi-supervised learning algorithm. A popular method on enforcing consistency is \textit{virtual adversarial training} \cite{miyato2018virtual}, which enforces consistency across carefully chosen image perturbations. Another paper, \textit{unsupervised data augmentation} \cite{xie2019unsupervised}, uses standard data augmentation techniques such as cutout \cite{devries2017improved} for images and back translation for text \cite{sennrich2015improving} to perturb images and enforces consistency across them. \cite{zhai2019s} uses consistency constraints along with other semi-supervised and self-supervised techniques in its full algorithm.

\subsubsection{Co-training}
Another way to provide feedback on unlabelled data is to train two (many) neural network models, each on a different \textit{view} of the raw data. For example, with text data, each model might see a different part of the input sentence. These models can then be given feedback to be maximally consistent with each other, or with a different model which sees all of the data, or even used for self-training, with each different model providing pseudo labels on the instances it is most confident on. This post \url{https://ruder.io/semi-supervised/} gives a nice overview of different co-training schemes, and \cite{clark2018semi, qiao2018deep, han2018co} are some recent papers implementing this in text and images.

\subsubsection{Semi-Supervised Learning Summary}
Semi-supervised learning is a powerful way to reduce the need for labelled data and can significantly boost the efficacy of deep learning models. Semi-supervised learning can be applied in any situation where a meaningful task can be created on the unlabelled data. In this section we have overviewed some natural ways to define such tasks, but there may be many creative alternatives depending on the domain of interest. We hope the references will provide a helpful starting point for implementation and further exploration!

\subsection{Data Augmentation}
As depicted in Figure \ref{fig:deeplearning-workflow}, \textit{data augmentation} is an important part of the deep learning workflow. Data augmentation refers to the process of artificially increasing the size and diversity of the training data by applying a variety of transformations to the raw data instances. For example, if the raw instances were to consist of images, we might artificially \textit{pad} out the image borders and then perform an off center (random) \textit{crop} to give us the final augmented image instance. Aside from increasing the size and diversity of the data, data augmentation offers the additional benefit of encouraging the neural network to be robust to certain kinds of common transformations of data instances. In this section, we overview some of the most popular data augmentation techniques for image and sequential data. These techniques will typically already be part of many open sourced deep learning pipelines, or easy to invoke in any mainstream deep learning software package. There are also some specific libraries written for augmentations, for example \textit{imgaug} \url{https://github.com/aleju/imgaug}, \textit{nlpaug} \url{https://github.com/makcedward/nlpaug} and \textit{albumentations} \url{https://github.com/albumentations-team/albumentations}.

\subsubsection{Data Augmentation for Image Data}
\begin{figure}
\centering
\includegraphics[width=0.6\columnwidth]{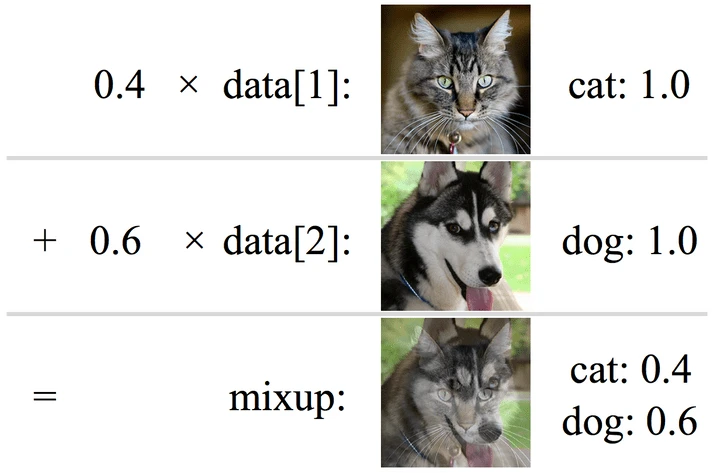} 
\caption{\small \textbf{An illustration of the Mixup data augmentation technique. Image source \cite{dauphin2017mixupimage}} The figure provides an example of the Mixup data augmentation method --- an image of a cat and an image of a dog are linearly combined, with $0.4$ weight on the cat and $0.6$ weight on the dog, to give a new input image shown in the bottom with a smoothed label of $0.4$ weight on cat and $0.6$ weight on dog. Mixup has been a very popular and successful data augmentation method for image tasks.}
\label{fig:mixup}
\end{figure}
Simple augmentations for image data consider transformations such as \textit{horizontal flips} or \textit{random crops} (padding the image borders and taking an off center crop.) Inspired by these simple methods are two very successful image augmentation strategies, \textit{cutout} \cite{devries2017improved}, which removes a patch from the input image, and RICAP \cite{takahashi2019data}, which combines patches from four different input image to create a new image (with new label a combination of the original labels.) This somewhat surprising latter technique of combining images has in fact shown to be very successful in \textit{mixup} \cite{zhang2017mixup}, another data augmentation strategy where linear combinations of images (instead of patches) are used. (This strategy has also been combined with cutout in the recently proposed \textit{cutmix} augmentation strategy \cite{yun2019cutmix}, with code \url{https://github.com/clovaai/CutMix-PyTorch}.)

Other useful augmentation strategies include \textit{TANDA} \cite{ratner2017snorkel} which learns a model to compose data augmentations, the related \textit{randaugment} \cite{cubuk2019randaugment}, choosing a random subset of different possible augmentations, population based augmentation \cite{ho2019population} which randomly searches over different augmentation policies, \cite{howard2013some} applying color distortions to the image and the recently proposed \textit{augmix} \cite{hendrycks2019augmix} (code \url{https://github.com/google-research/augmix}.)

\subsubsection{Data Augmentation for Sequence Data}
Data augmentation for sequential data typically falls into either (i) directly modifying the input sequence, or (ii) in the case of \textit{sequence to sequence} tasks (Section \ref{sec-seq-to-seq}), increasing the number of input-output sequences through noisy translation with the neural network. When directly modifying the input sequence, common perturbations include randomly deleting a sequence token (comparable to the masking approach used in \cite{devlin2018bert}), swapping sets of sequence tokens, and replacing a token with its \textit{synonym}. This latter strategy is usually guided by word embeddings \cite{wang2015s} or contextualized word embeddings \cite{kobayashi2018contextual}. Examples of combining these transformations are given by \cite{wei2019eda, jia2016data}, with code repositories such as \url{https://github.com/makcedward/nlpaug} providing some simple implementations. 

The other dominant approach to data augmentation of sequences is using sequence-to-sequence models to generate new data instances, known as \textit{back-translation} \cite{sennrich2015improving, edunov2018understanding}. Concretely, suppose we have a model to translate from English sequences to German sequences. We can take the output German sequence and use existing tools/noisy heuristics to translate it back to English. This gives us an additional English-German sequence pair.

\subsection{Data (Image) Denoising}
When measuring and collecting high dimensional data, noise can easily be introduced to the raw instances, be they images or single-cell data. As a result there has been significant interest and development of deep learning techniques to denoise the data. Many of these recent methods work even without paired noisy and clean data samples, and many be applicable in a broad range of settings. For instance, \textit{Noise2Noise} \cite{lehtinen2018noise2noise} uses a \textit{U-net} neural network architecture to denoise images given multiple noisy copies. The recent \textit{Noise2Self} \cite{batson2019noise2self} (with code: \url{https://github.com/czbiohub/noise2self}) frames denoising as a self-supervision problem, using different subsets of features (with assumed independent noise properties) to perform denoising, applying it to both images as well as other high dimensional data.

\section{Interpretability, Model Inspection and Representation Analysis}
\label{sec-interp-inspect-analysis}
Many standard applications of deep learning (and machine learning more broadly) focus on \textit{prediction} --- learning to output specific target values given an input. Scientific applications, on the other hand, are often focused on \textit{understanding} --- identifying underlying mechanisms giving rise to observed patterns in the data. When applying deep learning in scientific settings, we can use these observed phenomena as prediction targets, but the ultimate goal remains to understand what attributes give rise to these observations. For example, the core scientific question might be on how certain amino acid sequences (encoding a protein) give rise to particular kinds of protein function. While we might frame this as a prediction problem, training a deep neural network to take as input an amino acid sequence and output the predicted properties of the protein, we would ideally like to understand how that amino acid sequence resulted in the observed protein function.

To answer these kinds of questions, we can turn to \textit{interpretability} techniques. Interpretability methods are sometimes equated to a fully understandable, step-by-step explanation of the model's decision process. Such detailed insights can often be intractable, especially for complex deep neural network models. Instead, research in interpretability focuses on a much broader suite of techniques that can provide insights ranging from (rough) \textit{feature attributions} --- determining what input features matter the most, to \textit{model inspection} --- determining what causes certain neurons in the network to fire. In fact, these two examples also provide a rough split in the type of interpretability method. 

One large set of methods (which we refer to as Feature Attribution and Per Example Interpretability) concentrates on taking a specific input along with a trained deep neural network, and determining what features of the input are most important. The other broad class of techniques looks at taking a trained model, and a \textit{set} of inputs, to determine what different parts of the network have learned (referred to as Model Inspection and Representational Analysis). This latter set of methods can be very useful in revealing important, hidden patterns in the data that the model has implicitly learned through being trained on the predictive task. For example, in \cite{kudugunta2019investigating}, which looks at machine translation, representation analysis techniques are used to illustrate latent linguistic structure learned by the model. We overview both sets of methods below.

\subsection{Feature Attribution and Per Example Interpretability}
We start off by overviewing some of the popular techniques used to provide \textit{feature attribution} at a per example level, answering questions such as which parts of an input image are most important for a particular model prediction. These techniques can be further subcategorized as follows:

\subsubsection{Saliency Maps and Input Masks}
\label{sec-saliency-input-masks}
\begin{figure}
\centering
\includegraphics[width=0.7\columnwidth]{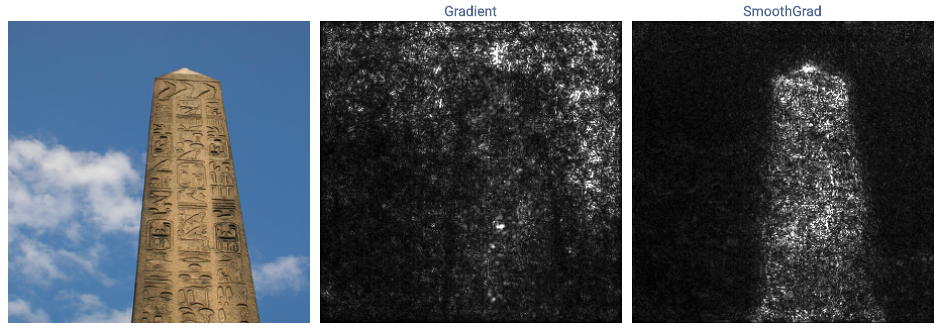} 
\caption{\small \textbf{The output of SmoothGrad, a type of saliency map. Image source \cite{smilkov2017smoothgrad}} The figure shows the original input image (left), raw gradients (middle), which are often too noisy for reliable feature attributions, and SmoothGrad (right), a type of saliency map that averages over perturbations to produce a more coherent feature attribution visualization the input. In particular, we can clearly see that the monument in the picture is important for the model output.}
\label{fig:smoothgrad}
\end{figure}
At a high level, saliency maps take the gradient of the \textit{output prediction} with respect to the \textit{input}. This gives a \textit{mask} over the input, highlighting which regions have large gradients (most important for the prediction) and which have smaller gradients. First introduced by \cite{simonyan2013deep}, there are many variants of saliency maps, such as Grad-CAM \cite{selvaraju2017grad}, SmoothGrad \cite{smilkov2017smoothgrad}, IntGrad \cite{sundararajan2017axiomatic}, which make the resulting feature attributions more robust. These and other methods are implemented in \url{https://github.com/PAIR-code/saliency}. Note that while these methods can be extremely useful, their predictions are not perfect \cite{kindermans2019reliability}, and must be validated further.

Closely related to these saliency methods is \cite{olah2018the}, which provides the ability to inspect the kinds of features causing neurons across different hidden layers to fire. The full, interactive paper can be read at \url{https://distill.pub/2018/building-blocks/} with code and tutorials available at \url{https://github.com/tensorflow/lucid}.

Many other techniques look at computing some kind of input mask, several of them using \textit{deconvolutional} layers, first proposed by \cite{zeiler2014visualizing} and built on by \cite{kindermans2017learning} and \cite{bojarski2016visualbackprop}. Other work looks at directly optimizing to find a sparse mask that will highlight the most important input features \cite{fong2017interpretable} (with associated code \url{https://github.com/jacobgil/pytorch-explain-black-box}) or finding such a mask through an iterative algorithm \cite{carter2018made}.

\subsubsection{Feature Ablations and Perturbations}
Related to some of masking approaches above, but with enough differences to categorize separately are several methods that isolate the crucial features of the input either by performing feature \textit{ablations} or computing perturbations of the input and using these perturbations along with the original input to inform the importance of different features.

Arguably the most well known of the ablation based approaches is the notion of a \textit{Shapely value}, first introduced in \cite{shapley1953value}. This estimates the importance of a particular feature $x_0$ in the input by computing the predictive power of a subset of input features containing $x_0$ and averaging over all possible such subsets. While Shapely values may be expensive to compute naively for deep learning, follow on work \cite{lundberg2017unified} has proposed more efficient (and expressive) variants, with highly popular opensourced implementation: \url{https://github.com/slundberg/shap}. 

The \textit{shap} opensourced implementation above also unifies some related approaches that use \textit{perturbations} to estimate feature values. One such approach is LIME \cite{ribeiro2016should}, which uses multiple local perturbations to enable learning an interpretable local model. Another is DeepLIFT, which uses a reference input to compare activation differences \cite{shrikumar2017learning}, and yet another approach, Layer-wise Relevance Propagation \cite{bach2015pixel} looks at computing relevance scores in a layerwise manner. 

Other work performing ablations to estimate feature importance includes \cite{zintgraf2017visualizing} (with code \url{https://github.com/lmzintgraf/DeepVis-PredDiff}), while \cite{fong2017interpretable}, described in Section \ref{sec-saliency-input-masks} has elements of using input perturbations.  
\subsection{Model Inspection and Representational Analysis}
In this second class of interpretability methods, the focus is on gaining insights not at a \textit{single} input example level, but using a set of examples (sometimes implicitly through the trained network) to understand the salient properties of the data. We overview some different approaches below.

\subsubsection{Probing and Activating Hidden Neurons}
\begin{figure}
\centering
\includegraphics[width=0.7\columnwidth]{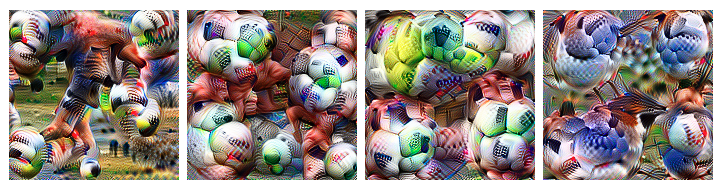} 
\caption{\small \textbf{Visualization of the kinds of features hidden neurons have learned to detect. Image source \cite{olah2017feature}} This figure, from \cite{olah2017feature}, illustrates the result of optimizing inputs to show what features hidden neurons have learned to recognize. In this example, the hidden neuron has learned to detect (especially) soccer balls, tennis balls, baseballs, and even the legs of soccer players.}
\label{fig:featurevisualization}
\end{figure}
A large class of interpretability methods looks at either (i) \textit{probing} hidden neurons in the neural network --- understanding what kinds of inputs it activates for (ii) directly optimizing the \textit{input} to activate a hidden neuron. Both of these techniques can provide useful insights into what the neural network has chosen to pay attention to, which in turn corresponds to important properties of the data. 

Several papers falls into the probing category \cite{yosinski2015understanding, zhou2014object}, with an especially thorough study given by \textit{Network Dissection} \cite{bau2017network}. Here here hidden neurons are categorized by the kinds of features they respond to. The paper website \url{http://netdissect.csail.mit.edu/} contains method details as well as links to the code and data. 

The other broad category of methods take a neural network, fix its parameters, and optimize the \textit{input} to find the kinds of features that makes some hidden neuron activate. There are several papers using this approach, but of particular note is \textit{Feature Visualization} \cite{olah2017feature}, with an interactive article and code at: \url{https://distill.pub/2017/feature-visualization/}. Followup work, \textit{Activation Atlases} \cite{carter2019activation} (with page \url{https://distill.pub/2019/activation-atlas/}), does this across many different concepts, providing a full mapping of the features learned by the neural network. More recently \cite{olah2020zoom} has used this as a building block to further understand how certain computations are performed in a neural network. Also related is \cite{kim2017interpretability}, which looks at finding linear combinations of hidden neurons that correspond to interpretable concepts. 
\begin{figure}
\centering
\includegraphics[width=0.7\columnwidth]{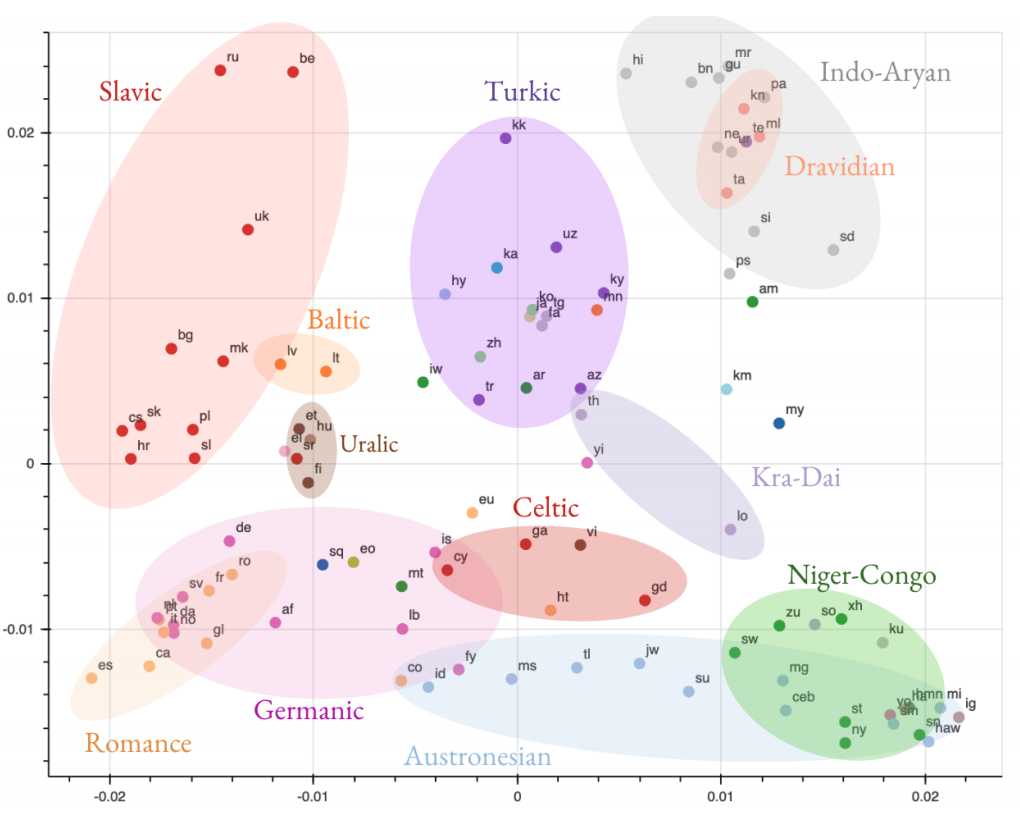} 
\caption{\small \textbf{Clustering neural network hidden representations to reveal linguistic structures. Image source \cite{kudugunta2019investigating}} In work on analyzing multilingual translation systems \cite{kudugunta2019investigating}, representational analysis techniques are used to compute similarity of neural network (Transformer) hidden representations across different languages. Performing clustering on the result reveals grouping of different language representations (each language a point on the plot) according to language families, which affect linguistic structure. Importantly, this analysis uses the neural network to identify key properties of the underlying data, a mode of investigation that might be very useful in scientific domains.}
\label{fig:svcca-clustering}
\end{figure}

\subsubsection{Dimensionality Reduction on Neural Network Hidden Representations}
\label{sec-dim-reduction-representations}
In many standard scientific settings, e.g. analyzing single cell data, dimensionality reduction methods such as PCA, t-SNE \cite{maaten2008visualizing}, UMAP \cite{mcinnes2018umap} are very useful in revealing important factors of variation and critical differences in the data subpopulations e.g. tumor cells vs healthy cells. Such methods can also be used on the \textit{hidden activations} (over some input dataset) of a neural network. Through the process of being trained on some predictive task, the neural network may implicitly learn these important data attributes in its hidden representations, which can then be extracted through dimensionality reduction methods.

\subsubsection{Representational Comparisons and Similarity}
Related to more standard approaches of dimensionality reduction and clustering, a line of work has studied \textit{comparing} hidden representations across different neural network models. Early work applied matching algorithms \cite{li2016convergent} with follow on approaches using \textit{canonical correlation analysis} \cite{raghu2017svcca, morcos2018insights} (with associated code \url{https://github.com/google/svcca}.) This latter approach has been used to identify and understand many representational properties in natural language applications \cite{kudugunta2019investigating, bau2018identifying, voita2019bottom} and even in modelling the mouse visual cortex as an artificial neural network \cite{shi2019comparison}. Another recent technique uses a kernel based approach to perform similarity comparisons \cite{kornblith2019similarity} (with code \url{https://colab.sandbox.google.com/github/google-research/google-research/blob/master/representation_similarity/Demo.ipynb}.)

\subsection{Technical References} The preceding sections contain many useful pointers to techniques and associated open sourced code references. One additional reference of general interest may be \url{https://christophm.github.io/interpretable-ml-book/} a fully open sourced book on interpretable machine learning. This focuses slightly more on more traditional interpretability methods, but has useful overlap with some of the techniques presented above and may suggest promising open directions.

\section{Advanced Deep Learning Methods}
\label{sec-advanced-dl}
The methods and tasks overviewed in the survey so far --- supervised learning, fundamental neural network architectures (and their many different tasks), different paradigms like transfer learning as well as ways to reduce labelled data dependence such as self-supervision and semi-supervised learning --- are an excellent set of first approaches for any problem amenable to deep learning. In most such problems, these approaches will also suffice in finding a good solution. 

Occasionally however, it might be useful to turn to more advanced methods in deep learning, specifically \textit{generative models} and \textit{reinforcement learning}. We term these methods advanced as they are often more intricate to implement, and may require specific properties of the problem to be useful, for example an excellent environment model/simulator for reinforcement learning. We provide a brief overview of these methods below.

\subsection{Generative Models}
\label{sec-generative-models}
At a high level, generative modelling has two fundamental goals. Firstly, it seeks to \textit{model} and enable \textit{sampling} from high dimensional data distributions, such as natural images. Secondly, it looks to learn low(er) dimensional latent encodings of the data that capture key properties of interest. 

To achieve the first goal, generative models take samples of the high dimensional distribution as input, for example, images of human faces, and learn some task directly on these data instances (e.g. encoding and then decoding the instance or learning to generate synthetic instances indistinguishable from the given data samples or generating values per-pixel using neighboring pixels as context). If generative modelling achieved \textit{perfect} success at this first goal, it would make it possible to continuously sample `free' data instances! Such perfect success is extremely challenging, but the past few years has seen enormous progress in the diversity and fidelity of samples from the data distribution.

For the second goal, learning latent encodings of the data with different encoding dimensions correspond to meaningful factors of variation, having an explicit encoder-decoder structure in the model can be helpful in encouraging learning such representations. This is the default structure for certain kinds of generative models such as \textit{variational autoencoders} \cite{kingma2014semi} but has also been adopted into other models, such as BigBiGAN \cite{donahue2019large}, a type of \textit{generative adversarial network}. In the following sections we overview some of these main types of generative models.

\subsubsection{Generative Adversarial Networks}
\begin{figure}
\centering
\includegraphics[width=0.7\columnwidth]{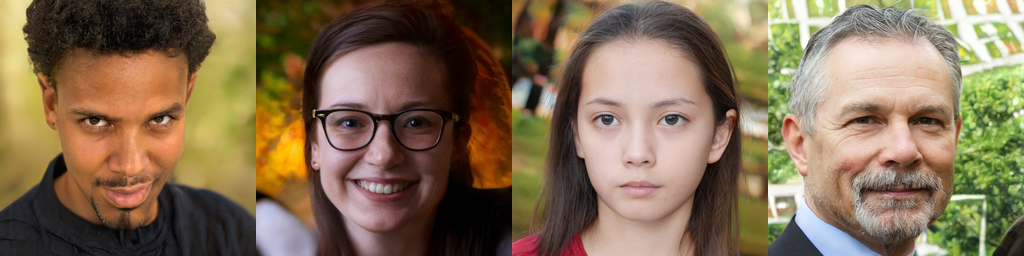} 
\caption{\small \textbf{Human faces generated from scratch by StyleGAN2. Image source \cite{karras2019style}} The figure shows multiple human face samples from StyleGAN2 \cite{karras2019style}. While perfectly modelling and capture full diversity of complex data distributions like human faces remains challenging, the quality and fidelity of samples from recent generative models is very high.}
\label{fig:style-gan}
\end{figure}
Arguably the most well known of all different types of generative models, Generative Adversarial Networks, commonly known as GANs, consist of two neural networks, a \textit{generator} and a \textit{discriminator}, which are pitted in a game against each other. The generator takes as input a \textit{random noise vector} and tries to output samples that look like the data distribution (e.g. synthesize images of peoples faces), while the discriminator tries to distinguish between true samples of the data, and those synthesized by the generator. First proposed in \cite{goodfellow2014generative}, GANs have been an exceptionally popular research area, with the most recent variations, such as BigGAN \cite{brock2018large} (code: \url{https://github.com/ajbrock/BigGAN-PyTorch}), BigBiGAN \cite{donahue2019large} and StyleGAN(2) \cite{karras2019style} (code: \url{https://github.com/NVlabs/stylegan2}) able to generate incredibly realistic images.

\paragraph{Unconditional GANs vs Conditional GANs} The examples given above are all \textit{unconditional} GANs, where the data is generated with only a random noise vector as input. A popular and highly useful variant are \textit{conditional} GANs, where generation is conditioned on additional information, such as a label, or a `source' image, which might be \textit{translated} to a different style. Examples include \textit{pix2pix} \cite{isola2017image} (code: \url{https://phillipi.github.io/pix2pix/}), cycleGAN \cite{zhu2017unpaired}, and applications of these to videos \cite{chan2019everybody}.

GANs have found many scientific applications, from performing data augmentation in medical image settings \cite{ghorbani2019dermgan} to protein generation \cite{repecka2019expanding}. The `adversarial' loss objective of GANs can make them somewhat tricky to train, and useful implementation advice is given in \url{https://www.fast.ai/2019/05/03/decrappify/}, and (for conditional GANs) is included in \url{https://github.com/jantic/DeOldify}.

\subsubsection{Variational Autoencoders}
Another type of generative model is given by the \textit{variational autoencoder}, first proposed by \cite{kingma2014semi}. VAEs have an encoder decoder structure, and thus an explicit latent encoding, which can capture useful properties of the data distribution. They also enable estimation of the \textit{likelihood} of a sampled datapoint --- the probability of its occurrence in the data distribution. VAEs have also been extremely popular, with many variations and extensions proposed \cite{sonderby2016ladder, johnson2016composing, kingma2017improving, grover2018graphite}. Because of the explicit latent encoding and the ability to estimate likelihoods, they have also found use cases in various scientific settings, such as for modelling gene expression in single-cell RNA sequencing \cite{lopez2017deep}.

\subsubsection{Autoregressive Models}
Yet another type of generative model is \textit{autoregressive models}, which take in inputs sequentially and use those to generate an appropriate output. For instance, such models may take in a sequence of pixel values (some of them generated at a previous timestep) and use these to generate a new pixel value for a specific spatial location. Autoregressive models such as PixelRNN \cite{oord2016pixel}, PixelCNN (and variants) \cite{van2016conditional, salimans2017pixelcnn++} and the recently proposed VQ-VAE(2) \cite{razavi2019generating} (code: \url{https://github.com/rosinality/vq-vae-2-pytorch}) offer very high generation quality.

\subsubsection{Flow Models}
A relatively new class of generative models, \textit{flow models}, looks at performing generation using a sequence of invertible transformations, which enables the computation of \textit{exact likelihoods.} First proposed in \cite{dinh2014nice, dinh2016density}, performing an expressive but tractable sequence of invertible transformations is an active area of research \cite{kingma2018glow, ho2019flow++}. A nice introduction to normalizing flows is given in this short video tutorial \url{https://www.youtube.com/watch?v=i7LjDvsLWCg&feature=youtu.be}.

\subsection{Reinforcement Learning}
Reinforcement learning has quite a different framing to the techniques and methods introduced so far, aiming to solve the \textit{sequential decision making} problem. It is typically introduced with the notions of an \textit{environment} and an \textit{agent}. The agent can take a \textit{sequence of actions} in the environment, each of which affect the environment \textit{state} in some way, and also result in possible \textit{rewards} (feedback) --- `positive' for good sequences of actions resulting in a `good' state and `negative' for bad sequences of actions leading to a `bad' state. For example, in a game like chess, the state is the current position of all pieces in play (the game state), an action the moving of a piece, with a good sequence of actions resulting in a win, a bad sequence of actions in a loss and the reward might be one or zero depending on having a win or loss respectively.

With this being the setup, the goal of reinforcement learning is to learn, through interaction with the environment, good sequences of actions (typically referred to as a \textit{policy}). Unlike supervised learning, feedback (the reward) is typically given only after performing the \textit{entire sequence} of actions. Specifically, feedback is \textit{sparse} and \textit{time delayed}. There are a variety of different reinforcement learning use cases depending on the specifics of the problem.

\subsubsection{RL with an Environment Model/Simulator}
Some of the most striking results with RL, such as AlphaGoZero \cite{silver2017mastering}, critically use an environment model/simulator. In such a setting, a variety of learning algorithms \cite{wang2016sample, schulman2017proximal, lillicrap2015continuous} (some code: \url{https://github.com/openai/baselines}) can help the agent learn  a good sequence of actions, often through simultaneously learning a \textit{value function} --- a function that determines whether a particular \textit{environment state} is beneficial or not. Because the benefit of an environment state may depend on the entire sequence of actions (some still in the future), RL is very important in properly assessing the value of the environment state, through implicitly accounting for possible future actions. Combining value functions with traditional search algorithms has been a very powerful way to use RL, and may be broadly applicable to many domains. 

Specifically, if developing a solution to the problem is multistep in nature, with even a noisy validation possible in simulation, using RL to learn a good value function and combining that with search algorithms may lead to discovering new and more effective parts of the search space. Approaches like these have gained traction in considering RL applications to fundamental problems in both computer systems, with \cite{mao2019park} providing a survey and a new benchmark, and machine learning systems \cite{pham2018efficient}, in designing task-specific neural network models. The latter has recently also resulted in scientific use cases --- designing neural networks to emulate complex processes across astronomy, chemistry, physics, climate modelling and others \cite{kasim2020up}.

\subsubsection{RL without Simulators}
In other settings, we don't have access to an environment model/simulator, and may simply have records of sequences of actions (and the ensuing states and rewards). This is the \textit{offline} setting. In this case, we may still try to teach an agent a good policy, using the observed sequences of actions/states/rewards in conjunction with \textit{off-policy} methods \cite{shortreed2011informing, raghu2017continuous, liu2019off}, but thorough validation and evaluation can be challenging. Evaluation in off-policy settings often uses a statistical technique known as \textit{off-policy policy evaluation} (example algorithms include \cite{precup2000eligibility,liu2018representation}). In robotics, reinforcement learning literature has looked at performing transfer learning between policies learned in simulation and policies learned on real data \cite{rusu2016sim}. A thorough overview of deep reinforcement learning is given in \url{http://rail.eecs.berkeley.edu/deeprlcourse/}.

\section{Implementation Tips}
\label{sec-implementation-tips}
In this section, we highlight some useful tips for implementing these models. 

\paragraph{Explore Your Data} Before starting with steps in the learning phase (see Figure \ref{fig:deeplearning-workflow}), make sure to perform a thorough exploration of your data. What are the results of simple dimensionality reduction methods or clustering? Are the labels reliable? Is there \textit{imbalance} amongst different classes? Are different subpopulations appropriately represented?

\paragraph{Try Simple Methods} When starting off with a completely new problem, it is useful to try the simplest version possible. (It might even be worthwhile starting with no learning at all --- how does the naive \textit{majority baseline} perform? For datasets with large imbalances, it may be quite strong!) If the dataset is very large, is there some smaller subsampled/downscaled version that can be used for faster preliminary testing? What is the simplest model that might work well? How does a majority baseline perform? (This ties in settings where the data has class imbalance.) Does the model (as expected) overfit to very small subsets of the data? 

\paragraph{Where possible, start with well tested models/tasks/methods} With the plethora of standard models (many of them pretrained), data augmentation, and optimization methods readily available (Section \ref{sec-frameworks-resources}), most new problems will be amenable to some standard set of these choices. \textit{Start with this!} Debugging the dataset and objective function associated with a new problem at the same time as debugging the neural network model, task choice, optimization algorithm, etc is very challenging. 

Additionally, many of the standard model/task/method choices are very well benchmarked, and exploring performance in these settings is an excellent first step in understanding the inherent challenges of the new problem. Wherever possible, the easiest way to get starting with the learning phase is to clone an appropriate github repository that has the models and training code needed, and make the minimal edits needed to work with the new dataset and objective function.

\paragraph{First Steps in Debugging Poor Performance} Having put together an end-to-end system, you observe that it is not performing well on the validation data. What is the reason? Before getting into more subtle design questions on hyperparameter choice (below), some first things to look at might be (i) Is the model overfitting? If so, more regularization, data augmentation, early stopping, smaller model may help. (ii) Is there a \textit{distribution shift} between the training and validation data? (iii) Is the model underfitting? If so, check the optimization process by seeing if the model \textit{overfits when trained on a smaller subset of the training data}. Test out a simpler task. Check for noise in the labels or data instances and for distribution shift. (iv) Look at the instances on which the model makes errors. Is there some pattern? For imbalanced datasets, loss function reweighting or more augmentation on the rarer classes can help. (v) How stable is the model performance across multiple random reruns? (vi) What are gradient and intermediate representation norms through the training process? 

\paragraph{Which hyperparameters matter most?} A big challenge in improving deep learning performance is the multitude of hyperparameters it is possible to change. In practice, some of the simplest hyperparameters often affect performance the most, such as learning rate and learning rate schedule. Picking an optimizer with subtleties such as weight decay correctly implemented can also be very important, see this excellent article on a very popular optimizer, AdamW \url{https://www.fast.ai/2018/07/02/adam-weight-decay/}. It might also be very useful to visualize the contributions to total loss from the main objective function vs different regularizers such as weight decay. 

Other hyperparameters that can be explored include batch size and data preprocessing, though if standard setups are used for these, varying learning rate related hyperparameters is likely to be the first most useful aspect to explore. To test different hyperparameter settings, it can be very useful to \textit{cross-validate}: hold out a portion of the training data, train different hyperparameter settings on the remaining data, pick whichever hyperparameter setting does best when evaluated on the held out data, and then finally retrain that hyperparameter setting on the full training dataset. 

\paragraph{Validate your model thoroughly!} Deep learning models are notorious for relying on \textit{spurious correlations} in the data to perform their predictions \cite{badgeley2019deep, oakden2019hidden, winkler2019association}. By spurious correlation, we mean features in the data instances that happen to co-occur with a specific label, but will not result in a robust, generalizable model. For example, suppose we have data from different chest x-ray machines (corresponding to different hospitals) that we put together to train a deep learning model. It might be the case that one of these machines, so happens to scan many sick patients. The deep learning model might then implicitly learn about the chest x-ray machine instead of the features of the illness. One of the best tests for ensuring the model is learning in a generalizable way is to evaluate the model on data collected \textit{separately} from the training data, which will introduce some natural \textit{distribution shift} and provide a more robust estimate of its accuracy. Some recent interesting papers exploring these questions include \cite{hendrycks2019benchmarking, recht2019imagenet}.

Relatedly, deep neural networks will also pick up on any biases in the data, for example, learning to pay attention to gender (a sensitive attribute) when made to predict age due to class imbalances leading to spurious correlations. This can pose significant challenges for generalizable conclusions in scientific settings where data may be collected from one population, but the predictions must be accurate across all populations. It is therefore important to perform postprocessing analysis on the model representations to identify the presence of such biases. A line of active research studies how to debias these representations \cite{alvi2018turning, wang2019towards}.

\paragraph{Implementation References} Some of the general design considerations when coming to implementation (along with factors affecting larger scale deployment, not explored in this survey) are discussed in this overview \url{https://github.com/chiphuyen/machine-learning-systems-design/blob/master/build/build1/consolidated.pdf}. 

For specific points on training and debugging deep learning systems, two excellent guides are given by \url{http://josh-tobin.com/assets/pdf/troubleshooting-deep-neural-networks-01-19.pdf} and \url{http://karpathy.github.io/2019/04/25/recipe/}.

\section{Conclusion}
As the amount of data collected across many diverse scientific domains continues to increase in both sheer amount and complexity, deep learning methods offer many exciting possibilities for both fundamental predictive problems as well as revealing subtle properties of the underlying data generation process. In this survey, we overviewed many of the highly successful deep learning models, tasks and methodologies, with references to the remarkably comprehensive open-sourced resources developed by the community. We hope that both the overviews and the references serve to accelerate applications of deep learning to many varied scientific problems!

\clearpage
\twocolumn
\section*{Acknowledgements}
The authors would like to thank Jon Kleinberg, Samy Bengio, Yann LeCun, Chiyuan Zhang, Quoc Le, Arun Chaganty, Simon Kornblith, Aniruddh Raghu, John Platt, Richard Murray, Stu Feldman and Guy Gur-Ari for feedback and comments on earlier versions.
\small
\bibliographystyle{plain}
\bibliography{references}

\begin{thebibliography}{100}

\bibitem{abdulla2018segmentation}
Waleed Abdulla.
\newblock {Splash of Color: Instance Segmentation with Mask R-CNN and
  TensorFlow}, 2018.
\newblock
  \url{https://engineering.matterport.com/splash-of-color-instance-segmentation-with-mask-r-cnn-and-tensorflow-7c761e238b46}.

\bibitem{aharoni2019massively}
Roee Aharoni, Melvin Johnson, and Orhan Firat.
\newblock Massively multilingual neural machine translation.
\newblock {\em arXiv preprint arXiv:1903.00089}, 2019.

\bibitem{alammar2018transformer}
Jay Alammar.
\newblock {The Illustrated Transformer}, 2018.
\newblock \url{http://jalammar.github.io/illustrated-transformer/}.

\bibitem{alvi2018turning}
Mohsan Alvi, Andrew Zisserman, and Christoffer Nell{\aa}ker.
\newblock Turning a blind eye: Explicit removal of biases and variation from
  deep neural network embeddings.
\newblock In {\em Proceedings of the European Conference on Computer Vision
  (ECCV)}, pages 0--0, 2018.

\bibitem{anthimopoulos2016lung}
Marios Anthimopoulos, Stergios Christodoulidis, Lukas Ebner, Andreas Christe,
  and Stavroula Mougiakakou.
\newblock Lung pattern classification for interstitial lung diseases using a
  deep convolutional neural network.
\newblock {\em IEEE transactions on medical imaging}, 35(5):1207--1216, 2016.

\bibitem{bach2015pixel}
Sebastian Bach, Alexander Binder, Gr{\'e}goire Montavon, Frederick Klauschen,
  Klaus-Robert M{\"u}ller, and Wojciech Samek.
\newblock On pixel-wise explanations for non-linear classifier decisions by
  layer-wise relevance propagation.
\newblock {\em PloS one}, 10(7):e0130140, 2015.

\bibitem{bachman2019learning}
Philip Bachman, R~Devon Hjelm, and William Buchwalter.
\newblock Learning representations by maximizing mutual information across
  views.
\newblock {\em arXiv preprint arXiv:1906.00910}, 2019.

\bibitem{badgeley2019deep}
Marcus~A Badgeley, John~R Zech, Luke Oakden-Rayner, Benjamin~S Glicksberg,
  Manway Liu, William Gale, Michael~V McConnell, Bethany Percha, Thomas~M
  Snyder, and Joel~T Dudley.
\newblock Deep learning predicts hip fracture using confounding patient and
  healthcare variables.
\newblock {\em npj Digital Medicine}, 2(1):31, 2019.

\bibitem{badrinarayanan2017segnet}
Vijay Badrinarayanan, Alex Kendall, and Roberto Cipolla.
\newblock Segnet: A deep convolutional encoder-decoder architecture for image
  segmentation.
\newblock {\em IEEE transactions on pattern analysis and machine intelligence},
  39(12):2481--2495, 2017.

\bibitem{bahdanau2014neural}
Dzmitry Bahdanau, Kyunghyun Cho, and Yoshua Bengio.
\newblock Neural machine translation by jointly learning to align and
  translate.
\newblock {\em arXiv preprint arXiv:1409.0473}, 2014.

\bibitem{bahdanau2016end}
Dzmitry Bahdanau, Jan Chorowski, Dmitriy Serdyuk, Philemon Brakel, and Yoshua
  Bengio.
\newblock End-to-end attention-based large vocabulary speech recognition.
\newblock In {\em 2016 IEEE international conference on acoustics, speech and
  signal processing (ICASSP)}, pages 4945--4949. IEEE, 2016.

\bibitem{balakrishnan2018unsupervised}
Guha Balakrishnan, Amy Zhao, Mert~R Sabuncu, John Guttag, and Adrian~V Dalca.
\newblock An unsupervised learning model for deformable medical image
  registration.
\newblock In {\em Proceedings of the IEEE conference on computer vision and
  pattern recognition}, pages 9252--9260, 2018.

\bibitem{balakrishnan2019voxelmorph}
Guha Balakrishnan, Amy Zhao, Mert~R Sabuncu, John Guttag, and Adrian~V Dalca.
\newblock Voxelmorph: a learning framework for deformable medical image
  registration.
\newblock {\em IEEE transactions on medical imaging}, 2019.

\bibitem{batson2019noise2self}
Joshua Batson and Loic Royer.
\newblock Noise2self: Blind denoising by self-supervision.
\newblock {\em arXiv preprint arXiv:1901.11365}, 2019.

\bibitem{battaglia2016interaction}
Peter Battaglia, Razvan Pascanu, Matthew Lai, Danilo~Jimenez Rezende, et~al.
\newblock Interaction networks for learning about objects, relations and
  physics.
\newblock In {\em Advances in neural information processing systems}, pages
  4502--4510, 2016.

\bibitem{bau2018identifying}
Anthony Bau, Yonatan Belinkov, Hassan Sajjad, Nadir Durrani, Fahim Dalvi, and
  James Glass.
\newblock Identifying and controlling important neurons in neural machine
  translation.
\newblock {\em arXiv preprint arXiv:1811.01157}, 2018.

\bibitem{bau2017network}
David Bau, Bolei Zhou, Aditya Khosla, Aude Oliva, and Antonio Torralba.
\newblock Network dissection: Quantifying interpretability of deep visual
  representations.
\newblock In {\em Proceedings of the IEEE Conference on Computer Vision and
  Pattern Recognition}, pages 6541--6549, 2017.

\bibitem{beltagy2019scibert}
Iz~Beltagy, Arman Cohan, and Kyle Lo.
\newblock Scibert: Pretrained contextualized embeddings for scientific text.
\newblock {\em arXiv preprint arXiv:1903.10676}, 2019.

\bibitem{berthelot2019mixmatch}
David Berthelot, Nicholas Carlini, Ian Goodfellow, Nicolas Papernot, Avital
  Oliver, and Colin Raffel.
\newblock Mixmatch: A holistic approach to semi-supervised learning.
\newblock {\em arXiv preprint arXiv:1905.02249}, 2019.

\bibitem{bojarski2016visualbackprop}
Mariusz Bojarski, Anna Choromanska, Krzysztof Choromanski, Bernhard Firner,
  Larry Jackel, Urs Muller, and Karol Zieba.
\newblock Visualbackprop: visualizing cnns for autonomous driving.
\newblock {\em arXiv preprint arXiv:1611.05418}, 2, 2016.

\bibitem{bresson2019two}
Xavier Bresson and Thomas Laurent.
\newblock A two-step graph convolutional decoder for molecule generation.
\newblock {\em arXiv preprint arXiv:1906.03412}, 2019.

\bibitem{brock2018large}
Andrew Brock, Jeff Donahue, and Karen Simonyan.
\newblock Large scale gan training for high fidelity natural image synthesis.
\newblock {\em arXiv preprint arXiv:1809.11096}, 2018.

\bibitem{cao2017prolango}
Renzhi Cao, Colton Freitas, Leong Chan, Miao Sun, Haiqing Jiang, and Zhangxin
  Chen.
\newblock Prolango: protein function prediction using neural machine
  translation based on a recurrent neural network.
\newblock {\em Molecules}, 22(10):1732, 2017.

\bibitem{cao2018openpose}
Zhe Cao, Gines Hidalgo, Tomas Simon, Shih-En Wei, and Yaser Sheikh.
\newblock Open{P}ose: realtime multi-person 2{D} pose estimation using {P}art
  {A}ffinity {F}ields.
\newblock In {\em arXiv preprint arXiv:1812.08008}, 2018.

\bibitem{carter2018made}
Brandon Carter, Jonas Mueller, Siddhartha Jain, and David Gifford.
\newblock What made you do this? understanding black-box decisions with
  sufficient input subsets.
\newblock {\em arXiv preprint arXiv:1810.03805}, 2018.

\bibitem{carter2019activation}
Shan Carter, Zan Armstrong, Ludwig Schubert, Ian Johnson, and Chris Olah.
\newblock Activation atlas.
\newblock {\em Distill}, 2019.
\newblock https://distill.pub/2019/activation-atlas.

\bibitem{chan2019everybody}
Caroline Chan, Shiry Ginosar, Tinghui Zhou, and Alexei~A Efros.
\newblock Everybody dance now.
\newblock In {\em Proceedings of the IEEE International Conference on Computer
  Vision}, pages 5933--5942, 2019.

\bibitem{chen2014fast}
Danqi Chen and Christopher Manning.
\newblock A fast and accurate dependency parser using neural networks.
\newblock In {\em Proceedings of the 2014 conference on empirical methods in
  natural language processing (EMNLP)}, pages 740--750, 2014.

\bibitem{chen2020simple}
Ting Chen, Simon Kornblith, Mohammad Norouzi, and Geoffrey Hinton.
\newblock A simple framework for contrastive learning of visual
  representations.
\newblock {\em arXiv preprint arXiv:2002.05709}, 2020.

\bibitem{chen2017sampling}
Ting Chen, Yizhou Sun, Yue Shi, and Liangjie Hong.
\newblock On sampling strategies for neural network-based collaborative
  filtering.
\newblock In {\em Proceedings of the 23rd ACM SIGKDD International Conference
  on Knowledge Discovery and Data Mining}, pages 767--776, 2017.

\bibitem{chen2018brain}
Yuhua Chen, Yibin Xie, Zhengwei Zhou, Feng Shi, Anthony~G Christodoulou, and
  Debiao Li.
\newblock Brain mri super resolution using 3d deep densely connected neural
  networks.
\newblock In {\em 2018 IEEE 15th International Symposium on Biomedical Imaging
  (ISBI 2018)}, pages 739--742. IEEE, 2018.

\bibitem{chorowski2015attention}
Jan~K Chorowski, Dzmitry Bahdanau, Dmitriy Serdyuk, Kyunghyun Cho, and Yoshua
  Bengio.
\newblock Attention-based models for speech recognition.
\newblock In {\em Advances in neural information processing systems}, pages
  577--585, 2015.

\bibitem{cciccek20163d}
{\"O}zg{\"u}n {\c{C}}i{\c{c}}ek, Ahmed Abdulkadir, Soeren~S Lienkamp, Thomas
  Brox, and Olaf Ronneberger.
\newblock 3d u-net: learning dense volumetric segmentation from sparse
  annotation.
\newblock In {\em International conference on medical image computing and
  computer-assisted intervention}, pages 424--432. Springer, 2016.

\bibitem{clark2018semi}
Kevin Clark, Minh-Thang Luong, Christopher~D Manning, and Quoc~V Le.
\newblock Semi-supervised sequence modeling with cross-view training.
\newblock {\em arXiv preprint arXiv:1809.08370}, 2018.

\bibitem{cohen2016group}
Taco Cohen and Max Welling.
\newblock Group equivariant convolutional networks.
\newblock In {\em International conference on machine learning}, pages
  2990--2999, 2016.

\bibitem{cohen2018spherical}
Taco~S Cohen, Mario Geiger, Jonas K{\"o}hler, and Max Welling.
\newblock Spherical cnns.
\newblock {\em arXiv preprint arXiv:1801.10130}, 2018.

\bibitem{corbett2018improving}
Peter Corbett and John Boyle.
\newblock Improving the learning of chemical-protein interactions from
  literature using transfer learning and specialized word embeddings.
\newblock {\em Database}, 2018, 2018.

\bibitem{cubuk2019randaugment}
Ekin~D Cubuk, Barret Zoph, Jonathon Shlens, and Quoc~V Le.
\newblock Randaugment: Practical data augmentation with no separate search.
\newblock {\em arXiv preprint arXiv:1909.13719}, 2019.

\bibitem{dalca2019learning}
Adrian Dalca, Marianne Rakic, John Guttag, and Mert Sabuncu.
\newblock Learning conditional deformable templates with convolutional
  networks.
\newblock In {\em Advances in neural information processing systems}, pages
  804--816, 2019.

\bibitem{dauphin2017mixupimage}
Yann Dauphin.
\newblock {mixup: Beyond Empirical Risk Minimization Image}, 2017.
\newblock \url{https://www.dauphin.io/}.

\bibitem{de2018molgan}
Nicola De~Cao and Thomas Kipf.
\newblock Molgan: An implicit generative model for small molecular graphs.
\newblock {\em arXiv preprint arXiv:1805.11973}, 2018.

\bibitem{deng2009imagenet}
Jia Deng, Wei Dong, Richard Socher, Li-Jia Li, Kai Li, and Li~Fei-Fei.
\newblock Imagenet: A large-scale hierarchical image database.
\newblock In {\em 2009 IEEE conference on computer vision and pattern
  recognition}, pages 248--255. Ieee, 2009.

\bibitem{devlin2018bert}
Jacob Devlin, Ming-Wei Chang, Kenton Lee, and Kristina Toutanova.
\newblock Bert: Pre-training of deep bidirectional transformers for language
  understanding.
\newblock {\em arXiv preprint arXiv:1810.04805}, 2018.

\bibitem{devries2017improved}
Terrance DeVries and Graham~W Taylor.
\newblock Improved regularization of convolutional neural networks with cutout.
\newblock {\em arXiv preprint arXiv:1708.04552}, 2017.

\bibitem{ding2018deep}
Yiming Ding, Jae~Ho Sohn, Michael~G Kawczynski, Hari Trivedi, Roy Harnish,
  Nathaniel~W Jenkins, Dmytro Lituiev, Timothy~P Copeland, Mariam~S Aboian,
  Carina Mari~Aparici, et~al.
\newblock A deep learning model to predict a diagnosis of alzheimer disease by
  using 18f-fdg pet of the brain.
\newblock {\em Radiology}, 290(2):456--464, 2018.

\bibitem{dinh2014nice}
Laurent Dinh, David Krueger, and Yoshua Bengio.
\newblock Nice: Non-linear independent components estimation.
\newblock {\em arXiv preprint arXiv:1410.8516}, 2014.

\bibitem{dinh2016density}
Laurent Dinh, Jascha Sohl-Dickstein, and Samy Bengio.
\newblock Density estimation using real nvp.
\newblock {\em arXiv preprint arXiv:1605.08803}, 2016.

\bibitem{doersch2015unsupervised}
Carl Doersch, Abhinav Gupta, and Alexei~A Efros.
\newblock Unsupervised visual representation learning by context prediction.
\newblock In {\em Proceedings of the IEEE International Conference on Computer
  Vision}, pages 1422--1430, 2015.

\bibitem{donahue2019large}
Jeff Donahue and Karen Simonyan.
\newblock Large scale adversarial representation learning.
\newblock In {\em Advances in Neural Information Processing Systems}, pages
  10541--10551, 2019.

\bibitem{dong2015image}
Chao Dong, Chen~Change Loy, Kaiming He, and Xiaoou Tang.
\newblock Image super-resolution using deep convolutional networks.
\newblock {\em IEEE transactions on pattern analysis and machine intelligence},
  38(2):295--307, 2015.

\bibitem{dosovitskiy2014discriminative}
Alexey Dosovitskiy, Jost~Tobias Springenberg, Martin Riedmiller, and Thomas
  Brox.
\newblock Discriminative unsupervised feature learning with convolutional
  neural networks.
\newblock In {\em Advances in neural information processing systems}, pages
  766--774, 2014.

\bibitem{du2015hierarchical}
Yong Du, Wei Wang, and Liang Wang.
\newblock Hierarchical recurrent neural network for skeleton based action
  recognition.
\newblock In {\em Proceedings of the IEEE conference on computer vision and
  pattern recognition}, pages 1110--1118, 2015.

\bibitem{duvenaud2015convolutional}
David~K Duvenaud, Dougal Maclaurin, Jorge Iparraguirre, Rafael Bombarell,
  Timothy Hirzel, Al{\'a}n Aspuru-Guzik, and Ryan~P Adams.
\newblock Convolutional networks on graphs for learning molecular fingerprints.
\newblock In {\em Advances in neural information processing systems}, pages
  2224--2232, 2015.

\bibitem{edunov2018understanding}
Sergey Edunov, Myle Ott, Michael Auli, and David Grangier.
\newblock Understanding back-translation at scale.
\newblock {\em arXiv preprint arXiv:1808.09381}, 2018.

\bibitem{esteva2017dermatologist}
Andre Esteva, Brett Kuprel, Roberto~A Novoa, Justin Ko, Susan~M Swetter,
  Helen~M Blau, and Sebastian Thrun.
\newblock Dermatologist-level classification of skin cancer with deep neural
  networks.
\newblock {\em Nature}, 542(7639):115, 2017.

\bibitem{fang2019deep}
Linjing Fang, Fred Monroe, Sammy~Weiser Novak, Lyndsey Kirk, Cara~R Schiavon,
  B~Yu Seungyoon, Tong Zhang, Melissa Wu, Kyle Kastner, Yoshiyuki Kubota,
  et~al.
\newblock Deep learning-based point-scanning super-resolution imaging.
\newblock {\em bioRxiv}, page 740548, 2019.

\bibitem{finn2016unsupervised}
Chelsea Finn, Ian Goodfellow, and Sergey Levine.
\newblock Unsupervised learning for physical interaction through video
  prediction.
\newblock In {\em Advances in neural information processing systems}, pages
  64--72, 2016.

\bibitem{finzi2020generalizing}
Marc Finzi, Samuel Stanton, Pavel Izmailov, and Andrew~Gordon Wilson.
\newblock Generalizing convolutional neural networks for equivariance to lie
  groups on arbitrary continuous data.
\newblock {\em arXiv preprint arXiv:2002.12880}, 2020.

\bibitem{fong2017interpretable}
Ruth~C Fong and Andrea Vedaldi.
\newblock Interpretable explanations of black boxes by meaningful perturbation.
\newblock In {\em Proceedings of the IEEE International Conference on Computer
  Vision}, pages 3429--3437, 2017.

\bibitem{fout2017protein}
Alex Fout, Jonathon Byrd, Basir Shariat, and Asa Ben-Hur.
\newblock Protein interface prediction using graph convolutional networks.
\newblock In {\em Advances in Neural Information Processing Systems}, pages
  6530--6539, 2017.

\bibitem{galko2018biomedical}
Ferenc Galk{\'o} and Carsten Eickhoff.
\newblock Biomedical question answering via weighted neural network passage
  retrieval.
\newblock In {\em European Conference on Information Retrieval}, pages
  523--528. Springer, 2018.

\bibitem{ganin2014unsupervised}
Yaroslav Ganin and Victor Lempitsky.
\newblock Unsupervised domain adaptation by backpropagation.
\newblock {\em arXiv preprint arXiv:1409.7495}, 2014.

\bibitem{ganin2016domain}
Yaroslav Ganin, Evgeniya Ustinova, Hana Ajakan, Pascal Germain, Hugo
  Larochelle, Fran{\c{c}}ois Laviolette, Mario Marchand, and Victor Lempitsky.
\newblock Domain-adversarial training of neural networks.
\newblock {\em The Journal of Machine Learning Research}, 17(1):2096--2030,
  2016.

\bibitem{gatys2016image}
Leon~A Gatys, Alexander~S Ecker, and Matthias Bethge.
\newblock Image style transfer using convolutional neural networks.
\newblock In {\em Proceedings of the IEEE conference on computer vision and
  pattern recognition}, pages 2414--2423, 2016.

\bibitem{ghorbani2019dermgan}
Amirata Ghorbani, Vivek Natarajan, David Coz, and Yuan Liu.
\newblock Dermgan: Synthetic generation of clinical skin images with pathology.
\newblock {\em arXiv preprint arXiv:1911.08716}, 2019.

\bibitem{gidaris2018unsupervised}
Spyros Gidaris, Praveer Singh, and Nikos Komodakis.
\newblock Unsupervised representation learning by predicting image rotations.
\newblock {\em arXiv preprint arXiv:1803.07728}, 2018.

\bibitem{gilmer2017neural}
Justin Gilmer, Samuel~S Schoenholz, Patrick~F Riley, Oriol Vinyals, and
  George~E Dahl.
\newblock Neural message passing for quantum chemistry.
\newblock In {\em Proceedings of the 34th International Conference on Machine
  Learning-Volume 70}, pages 1263--1272. JMLR. org, 2017.

\bibitem{goodfellow2014generative}
Ian Goodfellow, Jean Pouget-Abadie, Mehdi Mirza, Bing Xu, David Warde-Farley,
  Sherjil Ozair, Aaron Courville, and Yoshua Bengio.
\newblock Generative adversarial nets.
\newblock In {\em Advances in neural information processing systems}, pages
  2672--2680, 2014.

\bibitem{goyal2019scaling}
Priya Goyal, Dhruv Mahajan, Abhinav Gupta, and Ishan Misra.
\newblock Scaling and benchmarking self-supervised visual representation
  learning.
\newblock {\em arXiv preprint arXiv:1905.01235}, 2019.

\bibitem{graves2013speech}
Alex Graves, Abdel-rahman Mohamed, and Geoffrey Hinton.
\newblock Speech recognition with deep recurrent neural networks.
\newblock In {\em 2013 IEEE international conference on acoustics, speech and
  signal processing}, pages 6645--6649. IEEE, 2013.

\bibitem{grover2018graphite}
Aditya Grover, Aaron Zweig, and Stefano Ermon.
\newblock Graphite: Iterative generative modeling of graphs.
\newblock {\em arXiv preprint arXiv:1803.10459}, 2018.

\bibitem{gulshan2016development}
Varun Gulshan, Lily Peng, Marc Coram, Martin~C Stumpe, Derek Wu, Arunachalam
  Narayanaswamy, Subhashini Venugopalan, Kasumi Widner, Tom Madams, Jorge
  Cuadros, et~al.
\newblock Development and validation of a deep learning algorithm for detection
  of diabetic retinopathy in retinal fundus photographs.
\newblock {\em Jama}, 316(22):2402--2410, 2016.

\bibitem{habibi2017deep}
Maryam Habibi, Leon Weber, Mariana Neves, David~Luis Wiegandt, and Ulf Leser.
\newblock Deep learning with word embeddings improves biomedical named entity
  recognition.
\newblock {\em Bioinformatics}, 33(14):i37--i48, 2017.

\bibitem{han2018co}
Bo~Han, Quanming Yao, Xingrui Yu, Gang Niu, Miao Xu, Weihua Hu, Ivor Tsang, and
  Masashi Sugiyama.
\newblock Co-teaching: Robust training of deep neural networks with extremely
  noisy labels.
\newblock In {\em Advances in neural information processing systems}, pages
  8527--8537, 2018.

\bibitem{handa2016gvnn}
Ankur Handa, Michael Bloesch, Viorica P{\u{a}}tr{\u{a}}ucean, Simon Stent, John
  McCormac, and Andrew Davison.
\newblock gvnn: Neural network library for geometric computer vision.
\newblock In {\em European Conference on Computer Vision}, pages 67--82.
  Springer, 2016.

\bibitem{hanin2018neural}
Boris Hanin.
\newblock Which neural net architectures give rise to exploding and vanishing
  gradients?
\newblock In {\em Advances in Neural Information Processing Systems}, pages
  582--591, 2018.

\bibitem{hanson2016improving}
Jack Hanson, Yuedong Yang, Kuldip Paliwal, and Yaoqi Zhou.
\newblock Improving protein disorder prediction by deep bidirectional long
  short-term memory recurrent neural networks.
\newblock {\em Bioinformatics}, 33(5):685--692, 2016.

\bibitem{he2017mask}
Kaiming He, Georgia Gkioxari, Piotr Doll{\'a}r, and Ross Girshick.
\newblock Mask r-cnn.
\newblock In {\em Proceedings of the IEEE international conference on computer
  vision}, pages 2961--2969, 2017.

\bibitem{he2016deep}
Kaiming He, Xiangyu Zhang, Shaoqing Ren, and Jian Sun.
\newblock Deep residual learning for image recognition.
\newblock In {\em Proceedings of the IEEE conference on computer vision and
  pattern recognition}, pages 770--778, 2016.

\bibitem{heffernan2017capturing}
Rhys Heffernan, Yuedong Yang, Kuldip Paliwal, and Yaoqi Zhou.
\newblock Capturing non-local interactions by long short-term memory
  bidirectional recurrent neural networks for improving prediction of protein
  secondary structure, backbone angles, contact numbers and solvent
  accessibility.
\newblock {\em Bioinformatics}, 33(18):2842--2849, 2017.

\bibitem{hendrycks2019benchmarking}
Dan Hendrycks and Thomas Dietterich.
\newblock Benchmarking neural network robustness to common corruptions and
  perturbations.
\newblock {\em arXiv preprint arXiv:1903.12261}, 2019.

\bibitem{hendrycks2019augmix}
Dan Hendrycks, Norman Mu, Ekin~D Cubuk, Barret Zoph, Justin Gilmer, and Balaji
  Lakshminarayanan.
\newblock Augmix: A simple data processing method to improve robustness and
  uncertainty.
\newblock {\em arXiv preprint arXiv:1912.02781}, 2019.

\bibitem{hermann2015teaching}
Karl~Moritz Hermann, Tomas Kocisky, Edward Grefenstette, Lasse Espeholt, Will
  Kay, Mustafa Suleyman, and Phil Blunsom.
\newblock Teaching machines to read and comprehend.
\newblock In {\em Advances in neural information processing systems}, pages
  1693--1701, 2015.

\bibitem{hjelm2018learning}
R~Devon Hjelm, Alex Fedorov, Samuel Lavoie-Marchildon, Karan Grewal, Phil
  Bachman, Adam Trischler, and Yoshua Bengio.
\newblock Learning deep representations by mutual information estimation and
  maximization.
\newblock {\em arXiv preprint arXiv:1808.06670}, 2018.

\bibitem{ho2019population}
Daniel Ho, Eric Liang, Ion Stoica, Pieter Abbeel, and Xi~Chen.
\newblock Population based augmentation: Efficient learning of augmentation
  policy schedules.
\newblock {\em arXiv preprint arXiv:1905.05393}, 2019.

\bibitem{ho2019flow++}
Jonathan Ho, Xi~Chen, Aravind Srinivas, Yan Duan, and Pieter Abbeel.
\newblock Flow++: Improving flow-based generative models with variational
  dequantization and architecture design.
\newblock {\em arXiv preprint arXiv:1902.00275}, 2019.

\bibitem{hochreiter1998vanishing}
Sepp Hochreiter.
\newblock The vanishing gradient problem during learning recurrent neural nets
  and problem solutions.
\newblock {\em International Journal of Uncertainty, Fuzziness and
  Knowledge-Based Systems}, 6(02):107--116, 1998.

\bibitem{hochreiter1997long}
Sepp Hochreiter and J{\"u}rgen Schmidhuber.
\newblock Long short-term memory.
\newblock {\em Neural computation}, 9(8):1735--1780, 1997.

\bibitem{hoffmann2011knowledge}
Raphael Hoffmann, Congle Zhang, Xiao Ling, Luke Zettlemoyer, and Daniel~S Weld.
\newblock Knowledge-based weak supervision for information extraction of
  overlapping relations.
\newblock In {\em Proceedings of the 49th Annual Meeting of the Association for
  Computational Linguistics: Human Language Technologies-Volume 1}, pages
  541--550. Association for Computational Linguistics, 2011.

\bibitem{houlsby2019parameter}
Neil Houlsby, Andrei Giurgiu, Stanislaw Jastrzebski, Bruna Morrone, Quentin
  De~Laroussilhe, Andrea Gesmundo, Mona Attariyan, and Sylvain Gelly.
\newblock Parameter-efficient transfer learning for nlp.
\newblock {\em arXiv preprint arXiv:1902.00751}, 2019.

\bibitem{howard2013some}
Andrew~G Howard.
\newblock Some improvements on deep convolutional neural network based image
  classification.
\newblock {\em arXiv preprint arXiv:1312.5402}, 2013.

\bibitem{howard2018universal}
Jeremy Howard and Sebastian Ruder.
\newblock Universal language model fine-tuning for text classification.
\newblock {\em arXiv preprint arXiv:1801.06146}, 2018.

\bibitem{hu2019pre}
Weihua Hu, Bowen Liu, Joseph Gomes, Marinka Zitnik, Percy Liang, Vijay Pande,
  and Jure Leskovec.
\newblock Pre-training graph neural networks.
\newblock {\em arXiv preprint arXiv:1905.12265}, 2019.

\bibitem{huang2017densely}
Gao Huang, Zhuang Liu, Laurens Van Der~Maaten, and Kilian~Q Weinberger.
\newblock Densely connected convolutional networks.
\newblock In {\em Proceedings of the IEEE conference on computer vision and
  pattern recognition}, pages 4700--4708, 2017.

\bibitem{huh2016makes}
Minyoung Huh, Pulkit Agrawal, and Alexei~A Efros.
\newblock What makes imagenet good for transfer learning?
\newblock {\em arXiv preprint arXiv:1608.08614}, 2016.

\bibitem{isola2017image}
Phillip Isola, Jun-Yan Zhu, Tinghui Zhou, and Alexei~A Efros.
\newblock Image-to-image translation with conditional adversarial networks.
\newblock In {\em Proceedings of the IEEE conference on computer vision and
  pattern recognition}, pages 1125--1134, 2017.

\bibitem{ji2017adaptive}
Na~Ji.
\newblock Adaptive optical fluorescence microscopy.
\newblock {\em Nature methods}, 14(4):374, 2017.

\bibitem{jia2016data}
Robin Jia and Percy Liang.
\newblock Data recombination for neural semantic parsing.
\newblock {\em arXiv preprint arXiv:1606.03622}, 2016.

\bibitem{johnson2016composing}
Matthew Johnson, David~K Duvenaud, Alex Wiltschko, Ryan~P Adams, and Sandeep~R
  Datta.
\newblock Composing graphical models with neural networks for structured
  representations and fast inference.
\newblock In {\em Advances in neural information processing systems}, pages
  2946--2954, 2016.

\bibitem{karras2019style}
Tero Karras, Samuli Laine, and Timo Aila.
\newblock A style-based generator architecture for generative adversarial
  networks.
\newblock In {\em Proceedings of the IEEE Conference on Computer Vision and
  Pattern Recognition}, pages 4401--4410, 2019.

\bibitem{kasim2020up}
MF~Kasim, D~Watson-Parris, L~Deaconu, S~Oliver, P~Hatfield, DH~Froula,
  G~Gregori, M~Jarvis, S~Khatiwala, J~Korenaga, et~al.
\newblock Up to two billion times acceleration of scientific simulations with
  deep neural architecture search.
\newblock {\em arXiv preprint arXiv:2001.08055}, 2020.

\bibitem{kawahara2018seven}
Jeremy Kawahara, Sara Daneshvar, Giuseppe Argenziano, and Ghassan Hamarneh.
\newblock Seven-point checklist and skin lesion classification using multitask
  multimodal neural nets.
\newblock {\em IEEE journal of biomedical and health informatics},
  23(2):538--546, 2018.

\bibitem{kearnes2016molecular}
Steven Kearnes, Kevin McCloskey, Marc Berndl, Vijay Pande, and Patrick Riley.
\newblock Molecular graph convolutions: moving beyond fingerprints.
\newblock {\em Journal of computer-aided molecular design}, 30(8):595--608,
  2016.

\bibitem{kim2017interpretability}
Been Kim, Martin Wattenberg, Justin Gilmer, Carrie Cai, James Wexler, Fernanda
  Viegas, and Rory Sayres.
\newblock Interpretability beyond feature attribution: Quantitative testing
  with concept activation vectors (tcav).
\newblock {\em arXiv preprint arXiv:1711.11279}, 2017.

\bibitem{kindermans2019reliability}
Pieter-Jan Kindermans, Sara Hooker, Julius Adebayo, Maximilian Alber, Kristof~T
  Sch{\"u}tt, Sven D{\"a}hne, Dumitru Erhan, and Been Kim.
\newblock The (un) reliability of saliency methods.
\newblock In {\em Explainable AI: Interpreting, Explaining and Visualizing Deep
  Learning}, pages 267--280. Springer, 2019.

\bibitem{kindermans2017learning}
Pieter-Jan Kindermans, Kristof~T Sch{\"u}tt, Maximilian Alber, Klaus-Robert
  M{\"u}ller, Dumitru Erhan, Been Kim, and Sven D{\"a}hne.
\newblock Learning how to explain neural networks: Patternnet and
  patternattribution.
\newblock {\em arXiv preprint arXiv:1705.05598}, 2017.

\bibitem{kingma2017improving}
D~Kingma, Tim Salimans, R~Josefowicz, Xi~Chen, Ilya Sutskever, Max Welling,
  et~al.
\newblock Improving variational autoencoders with inverse autoregressive flow.
\newblock 2017.

\bibitem{kingma2018glow}
Durk~P Kingma and Prafulla Dhariwal.
\newblock Glow: Generative flow with invertible 1x1 convolutions.
\newblock In {\em Advances in Neural Information Processing Systems}, pages
  10215--10224, 2018.

\bibitem{kingma2014semi}
Durk~P Kingma, Shakir Mohamed, Danilo~Jimenez Rezende, and Max Welling.
\newblock Semi-supervised learning with deep generative models.
\newblock In {\em Advances in neural information processing systems}, pages
  3581--3589, 2014.

\bibitem{kobayashi2018contextual}
Sosuke Kobayashi.
\newblock Contextual augmentation: Data augmentation by words with paradigmatic
  relations.
\newblock {\em arXiv preprint arXiv:1805.06201}, 2018.

\bibitem{kojima2019recurrent}
Kaname Kojima, Shu Tadaka, Fumiki Katsuoka, Gen Tamiya, Masayuki Yamamoto, and
  Kengo Kinoshita.
\newblock A recurrent neural network based method for genotype imputation on
  phased genotype data.
\newblock {\em bioRxiv}, page 821504, 2019.

\bibitem{kolesnikov2019revisiting}
Alexander Kolesnikov, Xiaohua Zhai, and Lucas Beyer.
\newblock Revisiting self-supervised visual representation learning.
\newblock {\em arXiv preprint arXiv:1901.09005}, 2019.

\bibitem{komiske2019energy}
Patrick~T Komiske, Eric~M Metodiev, and Jesse Thaler.
\newblock Energy flow networks: deep sets for particle jets.
\newblock {\em Journal of High Energy Physics}, 2019(1):121, 2019.

\bibitem{kong2018image}
Shu Kong and Charless Fowlkes.
\newblock Image reconstruction with predictive filter flow.
\newblock {\em arXiv preprint arXiv:1811.11482}, 2018.

\bibitem{kornblith2019similarity}
Simon Kornblith, Mohammad Norouzi, Honglak Lee, and Geoffrey Hinton.
\newblock Similarity of neural network representations revisited.
\newblock {\em arXiv preprint arXiv:1905.00414}, 2019.

\bibitem{kornblith2019better}
Simon Kornblith, Jonathon Shlens, and Quoc~V Le.
\newblock Do better imagenet models transfer better?
\newblock In {\em Proceedings of the IEEE Conference on Computer Vision and
  Pattern Recognition}, pages 2661--2671, 2019.

\bibitem{krizhevsky2012imagenet}
Alex Krizhevsky, Ilya Sutskever, and Geoffrey~E Hinton.
\newblock Imagenet classification with deep convolutional neural networks.
\newblock In {\em Advances in neural information processing systems}, pages
  1097--1105, 2012.

\bibitem{kudugunta2019investigating}
Sneha~Reddy Kudugunta, Ankur Bapna, Isaac Caswell, Naveen Arivazhagan, and
  Orhan Firat.
\newblock Investigating multilingual nmt representations at scale.
\newblock {\em arXiv preprint arXiv:1909.02197}, 2019.

\bibitem{laine2016temporal}
Samuli Laine and Timo Aila.
\newblock Temporal ensembling for semi-supervised learning.
\newblock {\em arXiv preprint arXiv:1610.02242}, 2016.

\bibitem{lee2013pseudo}
Dong-Hyun Lee.
\newblock Pseudo-label: The simple and efficient semi-supervised learning
  method for deep neural networks.
\newblock In {\em Workshop on Challenges in Representation Learning, ICML},
  volume~3, page~2, 2013.

\bibitem{lee2019biobert}
Jinhyuk Lee, Wonjin Yoon, Sungdong Kim, Donghyeon Kim, Sunkyu Kim, Chan~Ho So,
  and Jaewoo Kang.
\newblock Biobert: pre-trained biomedical language representation model for
  biomedical text mining.
\newblock {\em arXiv preprint arXiv:1901.08746}, 2019.

\bibitem{lehtinen2018noise2noise}
Jaakko Lehtinen, Jacob Munkberg, Jon Hasselgren, Samuli Laine, Tero Karras,
  Miika Aittala, and Timo Aila.
\newblock Noise2noise: Learning image restoration without clean data.
\newblock {\em arXiv preprint arXiv:1803.04189}, 2018.

\bibitem{levy2014neural}
Omer Levy and Yoav Goldberg.
\newblock Neural word embedding as implicit matrix factorization.
\newblock In {\em Advances in neural information processing systems}, pages
  2177--2185, 2014.

\bibitem{li2018spatio}
Chaolong Li, Zhen Cui, Wenming Zheng, Chunyan Xu, and Jian Yang.
\newblock Spatio-temporal graph convolution for skeleton based action
  recognition.
\newblock In {\em Thirty-Second AAAI Conference on Artificial Intelligence},
  2018.

\bibitem{li2018improved}
Hailiang Li, Jian Weng, Yujian Shi, Wanrong Gu, Yijun Mao, Yonghua Wang, Weiwei
  Liu, and Jiajie Zhang.
\newblock An improved deep learning approach for detection of thyroid papillary
  cancer in ultrasound images.
\newblock {\em Scientific reports}, 8(1):6600, 2018.

\bibitem{li2016convergent}
Yixuan Li, Jason Yosinski, Jeff Clune, Hod Lipson, and John~E Hopcroft.
\newblock Convergent learning: Do different neural networks learn the same
  representations?
\newblock In {\em Iclr}, 2016.

\bibitem{li2015gated}
Yujia Li, Daniel Tarlow, Marc Brockschmidt, and Richard Zemel.
\newblock Gated graph sequence neural networks.
\newblock {\em arXiv preprint arXiv:1511.05493}, 2015.

\bibitem{lillicrap2015continuous}
Timothy~P Lillicrap, Jonathan~J Hunt, Alexander Pritzel, Nicolas Heess, Tom
  Erez, Yuval Tassa, David Silver, and Daan Wierstra.
\newblock Continuous control with deep reinforcement learning.
\newblock {\em arXiv preprint arXiv:1509.02971}, 2015.

\bibitem{liu2018deep}
Fang Liu, Zhaoye Zhou, Hyungseok Jang, Alexey Samsonov, Gengyan Zhao, and
  Richard Kijowski.
\newblock Deep convolutional neural network and 3d deformable approach for
  tissue segmentation in musculoskeletal magnetic resonance imaging.
\newblock {\em Magnetic resonance in medicine}, 79(4):2379--2391, 2018.

\bibitem{liu2018generating}
Peter~J Liu, Mohammad Saleh, Etienne Pot, Ben Goodrich, Ryan Sepassi, Lukasz
  Kaiser, and Noam Shazeer.
\newblock Generating wikipedia by summarizing long sequences.
\newblock {\em arXiv preprint arXiv:1801.10198}, 2018.

\bibitem{liu2015effects}
Shengyu Liu, Buzhou Tang, Qingcai Chen, and Xiaolong Wang.
\newblock Effects of semantic features on machine learning-based drug name
  recognition systems: word embeddings vs. manually constructed dictionaries.
\newblock {\em Information}, 6(4):848--865, 2015.

\bibitem{liu2017deep}
Xueliang Liu.
\newblock Deep recurrent neural network for protein function prediction from
  sequence.
\newblock {\em arXiv preprint arXiv:1701.08318}, 2017.

\bibitem{liu2018representation}
Yao Liu, Omer Gottesman, Aniruddh Raghu, Matthieu Komorowski, Aldo~A Faisal,
  Finale Doshi-Velez, and Emma Brunskill.
\newblock Representation balancing mdps for off-policy policy evaluation.
\newblock In {\em Advances in Neural Information Processing Systems}, pages
  2644--2653, 2018.

\bibitem{liu2019off}
Yao Liu, Adith Swaminathan, Alekh Agarwal, and Emma Brunskill.
\newblock Off-policy policy gradient with state distribution correction.
\newblock {\em arXiv preprint arXiv:1904.08473}, 2019.

\bibitem{liu2017detecting}
Yun Liu, Krishna Gadepalli, Mohammad Norouzi, George~E Dahl, Timo Kohlberger,
  Aleksey Boyko, Subhashini Venugopalan, Aleksei Timofeev, Philip~Q Nelson,
  Greg~S Corrado, et~al.
\newblock Detecting cancer metastases on gigapixel pathology images.
\newblock {\em arXiv preprint arXiv:1703.02442}, 2017.

\bibitem{long2015fully}
Jonathan Long, Evan Shelhamer, and Trevor Darrell.
\newblock Fully convolutional networks for semantic segmentation.
\newblock In {\em Proceedings of the IEEE conference on computer vision and
  pattern recognition}, pages 3431--3440, 2015.

\bibitem{long2017deep}
Mingsheng Long, Han Zhu, Jianmin Wang, and Michael~I Jordan.
\newblock Deep transfer learning with joint adaptation networks.
\newblock In {\em Proceedings of the 34th International Conference on Machine
  Learning-Volume 70}, pages 2208--2217. JMLR. org, 2017.

\bibitem{lopez2017deep}
Romain Lopez, Jeffrey Regier, Michael Cole, Michael Jordan, and Nir Yosef.
\newblock A deep generative model for gene expression profiles from single-cell
  rna sequencing.
\newblock {\em arXiv preprint arXiv:1709.02082}, 2017.

\bibitem{lu2018multimodal}
Donghuan Lu, Karteek Popuri, Gavin~Weiguang Ding, Rakesh Balachandar, and
  Mirza~Faisal Beg.
\newblock Multimodal and multiscale deep neural networks for the early
  diagnosis of alzheimer’s disease using structural mr and fdg-pet images.
\newblock {\em Scientific reports}, 8(1):5697, 2018.

\bibitem{lundberg2017unified}
Scott~M Lundberg and Su-In Lee.
\newblock A unified approach to interpreting model predictions.
\newblock In {\em Advances in Neural Information Processing Systems}, pages
  4765--4774, 2017.

\bibitem{maaten2008visualizing}
Laurens van~der Maaten and Geoffrey Hinton.
\newblock Visualizing data using t-sne.
\newblock {\em Journal of machine learning research}, 9(Nov):2579--2605, 2008.

\bibitem{madani2020progen}
Ali Madani, Bryan McCann, Nikhil Naik, Nitish~Shirish Keskar, Namrata Anand,
  Raphael~R Eguchi, Possu Huang, and Richard Socher.
\newblock Progen: Language modeling for protein generation.
\newblock {\em bioRxiv}, 2020.

\bibitem{mahajan2018exploring}
Dhruv Mahajan, Ross Girshick, Vignesh Ramanathan, Kaiming He, Manohar Paluri,
  Yixuan Li, Ashwin Bharambe, and Laurens van~der Maaten.
\newblock Exploring the limits of weakly supervised pretraining.
\newblock In {\em Proceedings of the European Conference on Computer Vision
  (ECCV)}, pages 181--196, 2018.

\bibitem{mao2019park}
Hongzi Mao, Parimarjan Negi, Akshay Narayan, Hanrui Wang, Jiacheng Yang, Haonan
  Wang, Ryan Marcus, Ravichandra Addanki, Mehrdad Khani, Songtao He, et~al.
\newblock Park: An open platform for learning augmented computer systems.
\newblock 2019.

\bibitem{marino2016building}
Daniel~L Marino, Kasun Amarasinghe, and Milos Manic.
\newblock Building energy load forecasting using deep neural networks.
\newblock In {\em IECON 2016-42nd Annual Conference of the IEEE Industrial
  Electronics Society}, pages 7046--7051. IEEE, 2016.

\bibitem{mathis2018deeplabcut}
Alexander Mathis, Pranav Mamidanna, Kevin~M Cury, Taiga Abe, Venkatesh~N
  Murthy, Mackenzie~Weygandt Mathis, and Matthias Bethge.
\newblock Deeplabcut: markerless pose estimation of user-defined body parts
  with deep learning.
\newblock {\em Nature neuroscience}, 21(9):1281, 2018.

\bibitem{mathis2020deep}
Mackenzie~Weygandt Mathis and Alexander Mathis.
\newblock Deep learning tools for the measurement of animal behavior in
  neuroscience.
\newblock {\em Current Opinion in Neurobiology}, 60:1--11, 2020.

\bibitem{mcinnes2018umap}
Leland McInnes, John Healy, and James Melville.
\newblock Umap: Uniform manifold approximation and projection for dimension
  reduction.
\newblock {\em arXiv preprint arXiv:1802.03426}, 2018.

\bibitem{merityRegOpt}
Stephen Merity, Nitish~Shirish Keskar, and Richard Socher.
\newblock {Regularizing and Optimizing LSTM Language Models}.
\newblock {\em arXiv preprint arXiv:1708.02182}, 2017.

\bibitem{merityAnalysis}
Stephen Merity, Nitish~Shirish Keskar, and Richard Socher.
\newblock {An Analysis of Neural Language Modeling at Multiple Scales}.
\newblock {\em arXiv preprint arXiv:1803.08240}, 2018.

\bibitem{mikolov2013efficient}
Tomas Mikolov, Kai Chen, Greg Corrado, and Jeffrey Dean.
\newblock Efficient estimation of word representations in vector space.
\newblock {\em arXiv preprint arXiv:1301.3781}, 2013.

\bibitem{mikolov2013distributed}
Tomas Mikolov, Ilya Sutskever, Kai Chen, Greg~S Corrado, and Jeff Dean.
\newblock Distributed representations of words and phrases and their
  compositionality.
\newblock In {\em Advances in neural information processing systems}, pages
  3111--3119, 2013.

\bibitem{mintz2009distant}
Mike Mintz, Steven Bills, Rion Snow, and Dan Jurafsky.
\newblock Distant supervision for relation extraction without labeled data.
\newblock In {\em Proceedings of the Joint Conference of the 47th Annual
  Meeting of the ACL and the 4th International Joint Conference on Natural
  Language Processing of the AFNLP: Volume 2-Volume 2}, pages 1003--1011.
  Association for Computational Linguistics, 2009.

\bibitem{misra2019selfsupervised}
Ishan Misra and Laurens van~der Maaten.
\newblock Self-supervised learning of pretext-invariant representations, 2019.

\bibitem{miyato2018virtual}
Takeru Miyato, Shin-ichi Maeda, Masanori Koyama, and Shin Ishii.
\newblock Virtual adversarial training: a regularization method for supervised
  and semi-supervised learning.
\newblock {\em IEEE transactions on pattern analysis and machine intelligence},
  41(8):1979--1993, 2018.

\bibitem{moeskops2016automatic}
Pim Moeskops, Max~A Viergever, Adri{\"e}nne~M Mendrik, Linda~S de~Vries,
  Manon~JNL Benders, and Ivana I{\v{s}}gum.
\newblock Automatic segmentation of mr brain images with a convolutional neural
  network.
\newblock {\em IEEE transactions on medical imaging}, 35(5):1252--1261, 2016.

\bibitem{morcos2018insights}
Ari Morcos, Maithra Raghu, and Samy Bengio.
\newblock Insights on representational similarity in neural networks with
  canonical correlation.
\newblock In {\em Advances in Neural Information Processing Systems}, pages
  5727--5736, 2018.

\bibitem{newell2016stacked}
Alejandro Newell, Kaiyu Yang, and Jia Deng.
\newblock Stacked hourglass networks for human pose estimation.
\newblock In {\em European conference on computer vision}, pages 483--499.
  Springer, 2016.

\bibitem{ngiam2018domain}
Jiquan Ngiam, Daiyi Peng, Vijay Vasudevan, Simon Kornblith, Quoc~V Le, and
  Ruoming Pang.
\newblock Domain adaptive transfer learning with specialist models.
\newblock {\em arXiv preprint arXiv:1811.07056}, 2018.

\bibitem{nori2019interpretml}
Harsha Nori, Samuel Jenkins, Paul Koch, and Rich Caruana.
\newblock Interpretml: A unified framework for machine learning
  interpretability.
\newblock {\em arXiv preprint arXiv:1909.09223}, 2019.

\bibitem{noroozi2016unsupervised}
Mehdi Noroozi and Paolo Favaro.
\newblock Unsupervised learning of visual representations by solving jigsaw
  puzzles.
\newblock In {\em European Conference on Computer Vision}, pages 69--84.
  Springer, 2016.

\bibitem{oakden2019hidden}
Luke Oakden-Rayner, Jared Dunnmon, Gustavo Carneiro, and Christopher R{\'e}.
\newblock Hidden stratification causes clinically meaningful failures in
  machine learning for medical imaging.
\newblock {\em arXiv preprint arXiv:1909.12475}, 2019.

\bibitem{olah2015understandinglstms}
Chris Olah.
\newblock {Understanding LSTM Networks}, 2015.
\newblock \url{https://colah.github.io/posts/2015-08-Understanding-LSTMs/}.

\bibitem{olah2020zoom}
Chris Olah, Nick Cammarata, Ludwig Schubert, Gabriel Goh, Michael Petrov, and
  Shan Carter.
\newblock Zoom in: An introduction to circuits.
\newblock {\em Distill}, 5(3):e00024--001, 2020.

\bibitem{olah2017feature}
Chris Olah, Alexander Mordvintsev, and Ludwig Schubert.
\newblock Feature visualization.
\newblock {\em Distill}, 2017.
\newblock https://distill.pub/2017/feature-visualization.

\bibitem{olah2018the}
Chris Olah, Arvind Satyanarayan, Ian Johnson, Shan Carter, Ludwig Schubert,
  Katherine Ye, and Alexander Mordvintsev.
\newblock The building blocks of interpretability.
\newblock {\em Distill}, 2018.
\newblock https://distill.pub/2018/building-blocks.

\bibitem{oord2016wavenet}
Aaron van~den Oord, Sander Dieleman, Heiga Zen, Karen Simonyan, Oriol Vinyals,
  Alex Graves, Nal Kalchbrenner, Andrew Senior, and Koray Kavukcuoglu.
\newblock Wavenet: A generative model for raw audio.
\newblock {\em arXiv preprint arXiv:1609.03499}, 2016.

\bibitem{oord2016pixel}
Aaron van~den Oord, Nal Kalchbrenner, and Koray Kavukcuoglu.
\newblock Pixel recurrent neural networks.
\newblock {\em arXiv preprint arXiv:1601.06759}, 2016.

\bibitem{oord2018representation}
Aaron van~den Oord, Yazhe Li, and Oriol Vinyals.
\newblock Representation learning with contrastive predictive coding.
\newblock {\em arXiv preprint arXiv:1807.03748}, 2018.

\bibitem{pascanu2012understanding}
Razvan Pascanu, Tomas Mikolov, and Yoshua Bengio.
\newblock Understanding the exploding gradient problem.
\newblock {\em CoRR, abs/1211.5063}, 2, 2012.

\bibitem{pathak2015constrained}
Deepak Pathak, Philipp Krahenbuhl, and Trevor Darrell.
\newblock Constrained convolutional neural networks for weakly supervised
  segmentation.
\newblock In {\em Proceedings of the IEEE international conference on computer
  vision}, pages 1796--1804, 2015.

\bibitem{paulus2017deep}
Romain Paulus, Caiming Xiong, and Richard Socher.
\newblock A deep reinforced model for abstractive summarization.
\newblock {\em arXiv preprint arXiv:1705.04304}, 2017.

\bibitem{pennington2014glove}
Jeffrey Pennington, Richard Socher, and Christopher Manning.
\newblock Glove: Global vectors for word representation.
\newblock In {\em Proceedings of the 2014 conference on empirical methods in
  natural language processing (EMNLP)}, pages 1532--1543, 2014.

\bibitem{pham2018efficient}
Hieu Pham, Melody~Y Guan, Barret Zoph, Quoc~V Le, and Jeff Dean.
\newblock Efficient neural architecture search via parameter sharing.
\newblock {\em arXiv preprint arXiv:1802.03268}, 2018.

\bibitem{pollastri2002improving}
Gianluca Pollastri, Darisz Przybylski, Burkhard Rost, and Pierre Baldi.
\newblock Improving the prediction of protein secondary structure in three and
  eight classes using recurrent neural networks and profiles.
\newblock {\em Proteins: Structure, Function, and Bioinformatics},
  47(2):228--235, 2002.

\bibitem{poplin2018prediction}
Ryan Poplin, Avinash~V Varadarajan, Katy Blumer, Yun Liu, Michael~V McConnell,
  Greg~S Corrado, Lily Peng, and Dale~R Webster.
\newblock Prediction of cardiovascular risk factors from retinal fundus
  photographs via deep learning.
\newblock {\em Nature Biomedical Engineering}, 2(3):158, 2018.

\bibitem{power2017guide}
Rory~M Power and Jan Huisken.
\newblock A guide to light-sheet fluorescence microscopy for multiscale
  imaging.
\newblock {\em Nature methods}, 14(4):360, 2017.

\bibitem{precup2000eligibility}
Doina Precup.
\newblock Eligibility traces for off-policy policy evaluation.
\newblock {\em Computer Science Department Faculty Publication Series},
  page~80, 2000.

\bibitem{qiao2018deep}
Siyuan Qiao, Wei Shen, Zhishuai Zhang, Bo~Wang, and Alan Yuille.
\newblock Deep co-training for semi-supervised image recognition.
\newblock In {\em Proceedings of the European Conference on Computer Vision
  (ECCV)}, pages 135--152, 2018.

\bibitem{radford2019language}
Alec Radford, Jeffrey Wu, Rewon Child, David Luan, Dario Amodei, and Ilya
  Sutskever.
\newblock Language models are unsupervised multitask learners.
\newblock {\em OpenAI Blog}, 1(8), 2019.

\bibitem{raffel2019exploring}
Colin Raffel, Noam Shazeer, Adam Roberts, Katherine Lee, Sharan Narang, Michael
  Matena, Yanqi Zhou, Wei Li, and Peter~J Liu.
\newblock Exploring the limits of transfer learning with a unified text-to-text
  transformer.
\newblock {\em arXiv preprint arXiv:1910.10683}, 2019.

\bibitem{raghu2017continuous}
Aniruddh Raghu, Matthieu Komorowski, Leo~Anthony Celi, Peter Szolovits, and
  Marzyeh Ghassemi.
\newblock Continuous state-space models for optimal sepsis treatment-a deep
  reinforcement learning approach.
\newblock {\em arXiv preprint arXiv:1705.08422}, 2017.

\bibitem{raghu2017svcca}
Maithra Raghu, Justin Gilmer, Jason Yosinski, and Jascha Sohl-Dickstein.
\newblock Svcca: Singular vector canonical correlation analysis for deep
  learning dynamics and interpretability.
\newblock In {\em Advances in Neural Information Processing Systems}, pages
  6076--6085, 2017.

\bibitem{raghu2019transfusion}
Maithra Raghu, Chiyuan Zhang, Jon Kleinberg, and Samy Bengio.
\newblock Transfusion: Understanding transfer learning for medical imaging.
\newblock In {\em Advances in Neural Information Processing Systems}, pages
  3342--3352, 2019.

\bibitem{rajpurkar2017chexnet}
Pranav Rajpurkar, Jeremy Irvin, Kaylie Zhu, Brandon Yang, Hershel Mehta, Tony
  Duan, Daisy Ding, Aarti Bagul, Curtis Langlotz, Katie Shpanskaya, et~al.
\newblock Chexnet: Radiologist-level pneumonia detection on chest x-rays with
  deep learning.
\newblock {\em arXiv preprint arXiv:1711.05225}, 2017.

\bibitem{rajpurkar2016squad}
Pranav Rajpurkar, Jian Zhang, Konstantin Lopyrev, and Percy Liang.
\newblock Squad: 100,000+ questions for machine comprehension of text.
\newblock {\em arXiv preprint arXiv:1606.05250}, 2016.

\bibitem{ramsundar2015massively}
Bharath Ramsundar, Steven Kearnes, Patrick Riley, Dale Webster, David
  Konerding, and Vijay Pande.
\newblock Massively multitask networks for drug discovery.
\newblock {\em arXiv preprint arXiv:1502.02072}, 2015.

\bibitem{ratner2017snorkel}
Alexander Ratner, Stephen~H Bach, Henry Ehrenberg, Jason Fries, Sen Wu, and
  Christopher R{\'e}.
\newblock Snorkel: Rapid training data creation with weak supervision.
\newblock {\em Proceedings of the VLDB Endowment}, 11(3):269--282, 2017.

\bibitem{razavi2019generating}
Ali Razavi, Aaron van~den Oord, and Oriol Vinyals.
\newblock Generating diverse high-fidelity images with vq-vae-2.
\newblock {\em arXiv preprint arXiv:1906.00446}, 2019.

\bibitem{recht2019imagenet}
Benjamin Recht, Rebecca Roelofs, Ludwig Schmidt, and Vaishaal Shankar.
\newblock Do imagenet classifiers generalize to imagenet?
\newblock {\em arXiv preprint arXiv:1902.10811}, 2019.

\bibitem{redmon2018yolov3}
Joseph Redmon and Ali Farhadi.
\newblock Yolov3: An incremental improvement.
\newblock {\em arXiv preprint arXiv:1804.02767}, 2018.

\bibitem{ren2015faster}
Shaoqing Ren, Kaiming He, Ross Girshick, and Jian Sun.
\newblock Faster r-cnn: Towards real-time object detection with region proposal
  networks.
\newblock In {\em Advances in neural information processing systems}, pages
  91--99, 2015.

\bibitem{repecka2019expanding}
Donatas Repecka, Vykintas Jauniskis, Laurynas Karpus, Elzbieta Rembeza, Jan
  Zrimec, Simona Poviloniene, Irmantas Rokaitis, Audrius Laurynenas, Wissam
  Abuajwa, Otto Savolainen, et~al.
\newblock Expanding functional protein sequence space using generative
  adversarial networks.
\newblock {\em bioRxiv}, page 789719, 2019.

\bibitem{ribeiro2016should}
Marco~Tulio Ribeiro, Sameer Singh, and Carlos Guestrin.
\newblock Why should i trust you?: Explaining the predictions of any
  classifier.
\newblock In {\em Proceedings of the 22nd ACM SIGKDD international conference
  on knowledge discovery and data mining}, pages 1135--1144. ACM, 2016.

\bibitem{rives2019biological}
Alexander Rives, Siddharth Goyal, Joshua Meier, Demi Guo, Myle Ott, C~Lawrence
  Zitnick, Jerry Ma, and Rob Fergus.
\newblock Biological structure and function emerge from scaling unsupervised
  learning to 250 million protein sequences.
\newblock {\em bioRxiv}, page 622803, 2019.

\bibitem{ronneberger2015u}
Olaf Ronneberger, Philipp Fischer, and Thomas Brox.
\newblock U-net: Convolutional networks for biomedical image segmentation.
\newblock In {\em International Conference on Medical image computing and
  computer-assisted intervention}, pages 234--241. Springer, 2015.

\bibitem{rush2015neural}
Alexander~M Rush, Sumit Chopra, and Jason Weston.
\newblock A neural attention model for abstractive sentence summarization.
\newblock {\em arXiv preprint arXiv:1509.00685}, 2015.

\bibitem{rusu2016sim}
Andrei~A Rusu, Mel Vecerik, Thomas Roth{\"o}rl, Nicolas Heess, Razvan Pascanu,
  and Raia Hadsell.
\newblock Sim-to-real robot learning from pixels with progressive nets.
\newblock {\em arXiv preprint arXiv:1610.04286}, 2016.

\bibitem{sa2017intervertebral}
Ruhan Sa, William Owens, Raymond Wiegand, Mark Studin, Donald Capoferri,
  Kenneth Barooha, Alexander Greaux, Robert Rattray, Adam Hutton, John
  Cintineo, et~al.
\newblock Intervertebral disc detection in x-ray images using faster r-cnn.
\newblock In {\em 2017 39th Annual International Conference of the IEEE
  Engineering in Medicine and Biology Society (EMBC)}, pages 564--567. IEEE,
  2017.

\bibitem{salimans2017pixelcnn++}
Tim Salimans, Andrej Karpathy, Xi~Chen, and Diederik~P Kingma.
\newblock Pixelcnn++: Improving the pixelcnn with discretized logistic mixture
  likelihood and other modifications.
\newblock {\em arXiv preprint arXiv:1701.05517}, 2017.

\bibitem{sanh2019distilbert}
Victor Sanh, Lysandre Debut, Julien Chaumond, and Thomas Wolf.
\newblock Distilbert, a distilled version of bert: smaller, faster, cheaper and
  lighter.
\newblock {\em arXiv preprint arXiv:1910.01108}, 2019.

\bibitem{sarraf2016deepad}
Saman Sarraf, Ghassem Tofighi, et~al.
\newblock Deepad: Alzheimer disease classification via deep convolutional
  neural networks using mri and fmri.
\newblock {\em BioRxiv}, page 070441, 2016.

\bibitem{schulman2017proximal}
John Schulman, Filip Wolski, Prafulla Dhariwal, Alec Radford, and Oleg Klimov.
\newblock Proximal policy optimization algorithms.
\newblock {\em arXiv preprint arXiv:1707.06347}, 2017.

\bibitem{selvaraju2017grad}
Ramprasaath~R Selvaraju, Michael Cogswell, Abhishek Das, Ramakrishna Vedantam,
  Devi Parikh, and Dhruv Batra.
\newblock Grad-cam: Visual explanations from deep networks via gradient-based
  localization.
\newblock In {\em Proceedings of the IEEE International Conference on Computer
  Vision}, pages 618--626, 2017.

\bibitem{senior2020improved}
Andrew~W Senior, Richard Evans, John Jumper, James Kirkpatrick, Laurent Sifre,
  Tim Green, Chongli Qin, Augustin {\v{Z}}{\'\i}dek, Alexander~WR Nelson, Alex
  Bridgland, et~al.
\newblock Improved protein structure prediction using potentials from deep
  learning.
\newblock {\em Nature}, pages 1--5, 2020.

\bibitem{sennrich2015improving}
Rico Sennrich, Barry Haddow, and Alexandra Birch.
\newblock Improving neural machine translation models with monolingual data.
\newblock {\em arXiv preprint arXiv:1511.06709}, 2015.

\bibitem{shapley1953value}
Lloyd~S Shapley.
\newblock A value for n-person games.
\newblock {\em Contributions to the Theory of Games}, 2(28):307--317, 1953.

\bibitem{shi2019comparison}
Jianghong Shi, Eric Shea-Brown, and Michael Buice.
\newblock Comparison against task driven artificial neural networks reveals
  functional properties in mouse visual cortex.
\newblock In {\em Advances in Neural Information Processing Systems}, pages
  5765--5775, 2019.

\bibitem{shortreed2011informing}
Susan~M Shortreed, Eric Laber, Daniel~J Lizotte, T~Scott Stroup, Joelle Pineau,
  and Susan~A Murphy.
\newblock Informing sequential clinical decision-making through reinforcement
  learning: an empirical study.
\newblock {\em Machine learning}, 84(1-2):109--136, 2011.

\bibitem{shrikumar2017learning}
Avanti Shrikumar, Peyton Greenside, and Anshul Kundaje.
\newblock Learning important features through propagating activation
  differences.
\newblock In {\em Proceedings of the 34th International Conference on Machine
  Learning-Volume 70}, pages 3145--3153. JMLR. org, 2017.

\bibitem{shu2018dirt}
Rui Shu, Hung~H Bui, Hirokazu Narui, and Stefano Ermon.
\newblock A dirt-t approach to unsupervised domain adaptation.
\newblock {\em arXiv preprint arXiv:1802.08735}, 2018.

\bibitem{silver2017mastering}
David Silver, Julian Schrittwieser, Karen Simonyan, Ioannis Antonoglou, Aja
  Huang, Arthur Guez, Thomas Hubert, Lucas Baker, Matthew Lai, Adrian Bolton,
  et~al.
\newblock Mastering the game of go without human knowledge.
\newblock {\em nature}, 550(7676):354--359, 2017.

\bibitem{simonyan2013deep}
Karen Simonyan, Andrea Vedaldi, and Andrew Zisserman.
\newblock Deep inside convolutional networks: Visualising image classification
  models and saliency maps.
\newblock {\em arXiv preprint arXiv:1312.6034}, 2013.

\bibitem{simonyan2014very}
Karen Simonyan and Andrew Zisserman.
\newblock Very deep convolutional networks for large-scale image recognition.
\newblock {\em arXiv preprint arXiv:1409.1556}, 2014.

\bibitem{smilkov2017smoothgrad}
Daniel Smilkov, Nikhil Thorat, Been Kim, Fernanda Vi{\'e}gas, and Martin
  Wattenberg.
\newblock Smoothgrad: removing noise by adding noise.
\newblock {\em arXiv preprint arXiv:1706.03825}, 2017.

\bibitem{sonderby2016ladder}
Casper~Kaae S{\o}nderby, Tapani Raiko, Lars Maal{\o}e, S{\o}ren~Kaae
  S{\o}nderby, and Ole Winther.
\newblock Ladder variational autoencoders.
\newblock In {\em Advances in neural information processing systems}, pages
  3738--3746, 2016.

\bibitem{song2014deep}
Youyi Song, Ling Zhang, Siping Chen, Dong Ni, Baopu Li, Yongjing Zhou, Baiying
  Lei, and Tianfu Wang.
\newblock A deep learning based framework for accurate segmentation of cervical
  cytoplasm and nuclei.
\newblock In {\em 2014 36th Annual International Conference of the IEEE
  Engineering in Medicine and Biology Society}, pages 2903--2906. IEEE, 2014.

\bibitem{sun2019deep}
Ke~Sun, Bin Xiao, Dong Liu, and Jingdong Wang.
\newblock Deep high-resolution representation learning for human pose
  estimation.
\newblock In {\em Proceedings of the IEEE Conference on Computer Vision and
  Pattern Recognition}, pages 5693--5703, 2019.

\bibitem{sun2019unsupervised}
Yu~Sun, Eric Tzeng, Trevor Darrell, and Alexei~A Efros.
\newblock Unsupervised domain adaptation through self-supervision.
\newblock {\em arXiv preprint arXiv:1909.11825}, 2019.

\bibitem{sundararajan2017axiomatic}
Mukund Sundararajan, Ankur Taly, and Qiqi Yan.
\newblock Axiomatic attribution for deep networks.
\newblock In {\em Proceedings of the 34th International Conference on Machine
  Learning-Volume 70}, pages 3319--3328. JMLR. org, 2017.

\bibitem{sutskever2014sequence}
I~Sutskever, O~Vinyals, and QV~Le.
\newblock Sequence to sequence learning with neural networks.
\newblock {\em Advances in NIPS}, 2014.

\bibitem{takahashi2019data}
Ryo Takahashi, Takashi Matsubara, and Kuniaki Uehara.
\newblock Data augmentation using random image cropping and patching for deep
  cnns.
\newblock {\em IEEE Transactions on Circuits and Systems for Video Technology},
  2019.

\bibitem{tan2019efficientnet}
Mingxing Tan and Quoc~V Le.
\newblock Efficientnet: Rethinking model scaling for convolutional neural
  networks.
\newblock {\em arXiv preprint arXiv:1905.11946}, 2019.

\bibitem{tan2019efficientdet}
Mingxing Tan, Ruoming Pang, and Quoc~V Le.
\newblock Efficientdet: Scalable and efficient object detection.
\newblock {\em arXiv preprint arXiv:1911.09070}, 2019.

\bibitem{tarvainen2017mean}
Antti Tarvainen and Harri Valpola.
\newblock Mean teachers are better role models: Weight-averaged consistency
  targets improve semi-supervised deep learning results.
\newblock In {\em Advances in neural information processing systems}, pages
  1195--1204, 2017.

\bibitem{tegunov2018real}
Dimitry Tegunov and Patrick Cramer.
\newblock Real-time cryo-em data pre-processing with warp.
\newblock {\em BioRxiv}, page 338558, 2018.

\bibitem{thian2019convolutional}
Yee~Liang Thian, Yiting Li, Pooja Jagmohan, David Sia, Vincent Ern~Yao Chan,
  and Robby~T Tan.
\newblock Convolutional neural networks for automated fracture detection and
  localization on wrist radiographs.
\newblock {\em Radiology: Artificial Intelligence}, 1(1):e180001, 2019.

\bibitem{topol2019high}
Eric~J Topol.
\newblock High-performance medicine: the convergence of human and artificial
  intelligence.
\newblock {\em Nature medicine}, 25(1):44--56, 2019.

\bibitem{townshend2019end}
Raphael Townshend, Rishi Bedi, Patricia Suriana, and Ron Dror.
\newblock End-to-end learning on 3d protein structure for interface prediction.
\newblock In {\em Advances in Neural Information Processing Systems}, pages
  15616--15625, 2019.

\bibitem{tshitoyan2019unsupervised}
Vahe Tshitoyan, John Dagdelen, Leigh Weston, Alexander Dunn, Ziqin Rong, Olga
  Kononova, Kristin~A Persson, Gerbrand Ceder, and Anubhav Jain.
\newblock Unsupervised word embeddings capture latent knowledge from materials
  science literature.
\newblock {\em Nature}, 571(7763):95--98, 2019.

\bibitem{umehara2018application}
Kensuke Umehara, Junko Ota, and Takayuki Ishida.
\newblock Application of super-resolution convolutional neural network for
  enhancing image resolution in chest ct.
\newblock {\em Journal of digital imaging}, 31(4):441--450, 2018.

\bibitem{van2016conditional}
Aaron Van~den Oord, Nal Kalchbrenner, Lasse Espeholt, Oriol Vinyals, Alex
  Graves, et~al.
\newblock Conditional image generation with pixelcnn decoders.
\newblock In {\em Advances in neural information processing systems}, pages
  4790--4798, 2016.

\bibitem{vaswani2017attention}
Ashish Vaswani, Noam Shazeer, Niki Parmar, Jakob Uszkoreit, Llion Jones,
  Aidan~N Gomez, {\L}ukasz Kaiser, and Illia Polosukhin.
\newblock Attention is all you need.
\newblock In {\em Advances in neural information processing systems}, pages
  5998--6008, 2017.

\bibitem{vinyals2015neural}
Oriol Vinyals and Quoc Le.
\newblock A neural conversational model.
\newblock {\em arXiv preprint arXiv:1506.05869}, 2015.

\bibitem{voita2019bottom}
Elena Voita, Rico Sennrich, and Ivan Titov.
\newblock The bottom-up evolution of representations in the transformer: A
  study with machine translation and language modeling objectives.
\newblock {\em arXiv preprint arXiv:1909.01380}, 2019.

\bibitem{wachinger2018deepnat}
Christian Wachinger, Martin Reuter, and Tassilo Klein.
\newblock Deepnat: Deep convolutional neural network for segmenting
  neuroanatomy.
\newblock {\em NeuroImage}, 170:434--445, 2018.

\bibitem{wang2019medical}
Kun Wang, Bite Yang, Guohai Xu, and Xiaofeng He.
\newblock Medical question retrieval based on siamese neural network and
  transfer learning method.
\newblock In {\em International Conference on Database Systems for Advanced
  Applications}, pages 49--64. Springer, 2019.

\bibitem{wang2018ajile}
Nancy~XR Wang, Ali Farhadi, Rajesh~PN Rao, and Bingni~W Brunton.
\newblock Ajile movement prediction: Multimodal deep learning for natural human
  neural recordings and video.
\newblock In {\em Thirty-Second AAAI Conference on Artificial Intelligence},
  2018.

\bibitem{wang2015s}
William~Yang Wang and Diyi Yang.
\newblock That’s so annoying!!!: A lexical and frame-semantic embedding based
  data augmentation approach to automatic categorization of annoying behaviors
  using\# petpeeve tweets.
\newblock In {\em Proceedings of the 2015 Conference on Empirical Methods in
  Natural Language Processing}, pages 2557--2563, 2015.

\bibitem{wang2018non}
Xiaolong Wang, Ross Girshick, Abhinav Gupta, and Kaiming He.
\newblock Non-local neural networks.
\newblock In {\em Proceedings of the IEEE Conference on Computer Vision and
  Pattern Recognition}, pages 7794--7803, 2018.

\bibitem{wang2019towards}
Zeyu Wang, Klint Qinami, Yannis Karakozis, Kyle Genova, Prem Nair, Kenji Hata,
  and Olga Russakovsky.
\newblock Towards fairness in visual recognition: Effective strategies for bias
  mitigation.
\newblock {\em arXiv preprint arXiv:1911.11834}, 2019.

\bibitem{wang2016sample}
Ziyu Wang, Victor Bapst, Nicolas Heess, Volodymyr Mnih, Remi Munos, Koray
  Kavukcuoglu, and Nando de~Freitas.
\newblock Sample efficient actor-critic with experience replay.
\newblock {\em arXiv preprint arXiv:1611.01224}, 2016.

\bibitem{wei2019eda}
Jason~W Wei and Kai Zou.
\newblock Eda: Easy data augmentation techniques for boosting performance on
  text classification tasks.
\newblock {\em arXiv preprint arXiv:1901.11196}, 2019.

\bibitem{wei2016convolutional}
Shih-En Wei, Varun Ramakrishna, Takeo Kanade, and Yaser Sheikh.
\newblock Convolutional pose machines.
\newblock In {\em Proceedings of the IEEE Conference on Computer Vision and
  Pattern Recognition}, pages 4724--4732, 2016.

\bibitem{weigert2018content}
Martin Weigert, Uwe Schmidt, Tobias Boothe, Andreas M{\"u}ller, Alexandr
  Dibrov, Akanksha Jain, Benjamin Wilhelm, Deborah Schmidt, Coleman Broaddus,
  Si{\^a}n Culley, et~al.
\newblock Content-aware image restoration: pushing the limits of fluorescence
  microscopy.
\newblock {\em Nature methods}, 15(12):1090, 2018.

\bibitem{weiss2015structured}
David Weiss, Chris Alberti, Michael Collins, and Slav Petrov.
\newblock Structured training for neural network transition-based parsing.
\newblock {\em arXiv preprint arXiv:1506.06158}, 2015.

\bibitem{winkler2019association}
Julia~K Winkler, Christine Fink, Ferdinand Toberer, Alexander Enk, Teresa
  Deinlein, Rainer Hofmann-Wellenhof, Luc Thomas, Aimilios Lallas, Andreas
  Blum, Wilhelm Stolz, et~al.
\newblock Association between surgical skin markings in dermoscopic images and
  diagnostic performance of a deep learning convolutional neural network for
  melanoma recognition.
\newblock {\em JAMA dermatology}, 155(10):1135--1141, 2019.

\bibitem{wu2019detectron2}
Yuxin Wu, Alexander Kirillov, Francisco Massa, Wan-Yen Lo, and Ross Girshick.
\newblock Detectron2.
\newblock \url{https://github.com/facebookresearch/detectron2}, 2019.

\bibitem{wu2018moleculenet}
Zhenqin Wu, Bharath Ramsundar, Evan~N Feinberg, Joseph Gomes, Caleb Geniesse,
  Aneesh~S Pappu, Karl Leswing, and Vijay Pande.
\newblock Moleculenet: a benchmark for molecular machine learning.
\newblock {\em Chemical science}, 9(2):513--530, 2018.

\bibitem{wu2019comprehensive}
Zonghan Wu, Shirui Pan, Fengwen Chen, Guodong Long, Chengqi Zhang, and Philip~S
  Yu.
\newblock A comprehensive survey on graph neural networks.
\newblock {\em arXiv preprint arXiv:1901.00596}, 2019.

\bibitem{xie2019unsupervised}
Qizhe Xie, Zihang Dai, Eduard Hovy, Minh-Thang Luong, and Quoc~V Le.
\newblock Unsupervised data augmentation.
\newblock {\em arXiv preprint arXiv:1904.12848}, 2019.

\bibitem{xie2019self}
Qizhe Xie, Eduard Hovy, Minh-Thang Luong, and Quoc~V Le.
\newblock Self-training with noisy student improves imagenet classification.
\newblock {\em arXiv preprint arXiv:1911.04252}, 2019.

\bibitem{xie2017aggregated}
Saining Xie, Ross Girshick, Piotr Doll{\'a}r, Zhuowen Tu, and Kaiming He.
\newblock Aggregated residual transformations for deep neural networks.
\newblock In {\em Proceedings of the IEEE conference on computer vision and
  pattern recognition}, pages 1492--1500, 2017.

\bibitem{xu2016deep}
Jun Xu, Xiaofei Luo, Guanhao Wang, Hannah Gilmore, and Anant Madabhushi.
\newblock A deep convolutional neural network for segmenting and classifying
  epithelial and stromal regions in histopathological images.
\newblock {\em Neurocomputing}, 191:214--223, 2016.

\bibitem{yan2019clusterfit}
Xueting Yan, Ishan Misra, Abhinav Gupta, Deepti Ghadiyaram, and Dhruv Mahajan.
\newblock Clusterfit: Improving generalization of visual representations.
\newblock {\em arXiv preprint arXiv:1912.03330}, 2019.

\bibitem{yang2019deep}
Yilong Yang, Zhuyifan Ye, Yan Su, Qianqian Zhao, Xiaoshan Li, and Defang
  Ouyang.
\newblock Deep learning for in vitro prediction of pharmaceutical formulations.
\newblock {\em Acta pharmaceutica sinica B}, 9(1):177--185, 2019.

\bibitem{yang2019xlnet}
Zhilin Yang, Zihang Dai, Yiming Yang, Jaime Carbonell, Ruslan Salakhutdinov,
  and Quoc~V Le.
\newblock Xlnet: Generalized autoregressive pretraining for language
  understanding.
\newblock {\em arXiv preprint arXiv:1906.08237}, 2019.

\bibitem{yasaka2017deep}
Koichiro Yasaka, Hiroyuki Akai, Osamu Abe, and Shigeru Kiryu.
\newblock Deep learning with convolutional neural network for differentiation
  of liver masses at dynamic contrast-enhanced ct: a preliminary study.
\newblock {\em Radiology}, 286(3):887--896, 2017.

\bibitem{yosinski2015understanding}
Jason Yosinski, Jeff Clune, Anh Nguyen, Thomas Fuchs, and Hod Lipson.
\newblock Understanding neural networks through deep visualization.
\newblock {\em arXiv preprint arXiv:1506.06579}, 2015.

\bibitem{yuan2019object}
Yuhui Yuan, Xilin Chen, and Jingdong Wang.
\newblock Object-contextual representations for semantic segmentation.
\newblock {\em arXiv preprint arXiv:1909.11065}, 2019.

\bibitem{yun2019cutmix}
Sangdoo Yun, Dongyoon Han, Seong~Joon Oh, Sanghyuk Chun, Junsuk Choe, and
  Youngjoon Yoo.
\newblock Cutmix: Regularization strategy to train strong classifiers with
  localizable features.
\newblock In {\em Proceedings of the IEEE International Conference on Computer
  Vision}, pages 6023--6032, 2019.

\bibitem{zaheer2017deep}
Manzil Zaheer, Satwik Kottur, Siamak Ravanbakhsh, Barnabas Poczos, Russ~R
  Salakhutdinov, and Alexander~J Smola.
\newblock Deep sets.
\newblock In {\em Advances in neural information processing systems}, pages
  3391--3401, 2017.

\bibitem{zeiler2014visualizing}
Matthew~D Zeiler and Rob Fergus.
\newblock Visualizing and understanding convolutional networks.
\newblock In {\em European conference on computer vision}, pages 818--833.
  Springer, 2014.

\bibitem{zeng2015distant}
Daojian Zeng, Kang Liu, Yubo Chen, and Jun Zhao.
\newblock Distant supervision for relation extraction via piecewise
  convolutional neural networks.
\newblock In {\em Proceedings of the 2015 Conference on Empirical Methods in
  Natural Language Processing}, pages 1753--1762, 2015.

\bibitem{zhai2019s}
Xiaohua Zhai, Avital Oliver, Alexander Kolesnikov, and Lucas Beyer.
\newblock S4l: Self-supervised semi-supervised learning.
\newblock {\em arXiv preprint arXiv:1905.03670}, 2019.

\bibitem{zhai2019visual}
Xiaohua Zhai, Joan Puigcerver, Alexander Kolesnikov, Pierre Ruyssen, Carlos
  Riquelme, Mario Lucic, Josip Djolonga, Andre~Susano Pinto, Maxim Neumann,
  Alexey Dosovitskiy, et~al.
\newblock The visual task adaptation benchmark.
\newblock {\em arXiv preprint arXiv:1910.04867}, 2019.

\bibitem{zhang2019dive}
Aston Zhang, Zachary~C Lipton, Mu~Li, and Alexander~J Smola.
\newblock Dive into deep learning.
\newblock {\em Unpublished draft. Retrieved}, 3:319, 2019.

\bibitem{zhang2017mixup}
Hongyi Zhang, Moustapha Cisse, Yann~N Dauphin, and David Lopez-Paz.
\newblock mixup: Beyond empirical risk minimization.
\newblock {\em arXiv preprint arXiv:1710.09412}, 2017.

\bibitem{zhang2016cancer}
Junkang Zhang, Haigen Hu, Shengyong Chen, Yujiao Huang, and Qiu Guan.
\newblock Cancer cells detection in phase-contrast microscopy images based on
  faster r-cnn.
\newblock In {\em 2016 9th International Symposium on Computational
  Intelligence and Design (ISCID)}, volume~1, pages 363--367. IEEE, 2016.

\bibitem{zhang2018residual}
Yulun Zhang, Yapeng Tian, Yu~Kong, Bineng Zhong, and Yun Fu.
\newblock Residual dense network for image super-resolution.
\newblock In {\em Proceedings of the IEEE Conference on Computer Vision and
  Pattern Recognition}, pages 2472--2481, 2018.

\bibitem{zhong2017seq2sql}
Victor Zhong, Caiming Xiong, and Richard Socher.
\newblock Seq2sql: Generating structured queries from natural language using
  reinforcement learning.
\newblock {\em arXiv preprint arXiv:1709.00103}, 2017.

\bibitem{zhou2014object}
Bolei Zhou, Aditya Khosla, Agata Lapedriza, Aude Oliva, and Antonio Torralba.
\newblock Object detectors emerge in deep scene cnns.
\newblock {\em arXiv preprint arXiv:1412.6856}, 2014.

\bibitem{zhou2018neural}
Qingyu Zhou, Nan Yang, Furu Wei, Shaohan Huang, Ming Zhou, and Tiejun Zhao.
\newblock Neural document summarization by jointly learning to score and select
  sentences.
\newblock {\em arXiv preprint arXiv:1807.02305}, 2018.

\bibitem{zhu2017unpaired}
Jun-Yan Zhu, Taesung Park, Phillip Isola, and Alexei~A Efros.
\newblock Unpaired image-to-image translation using cycle-consistent
  adversarial networks.
\newblock In {\em Proceedings of the IEEE international conference on computer
  vision}, pages 2223--2232, 2017.

\bibitem{zintgraf2017visualizing}
Luisa~M Zintgraf, Taco~S Cohen, Tameem Adel, and Max Welling.
\newblock Visualizing deep neural network decisions: Prediction difference
  analysis.
\newblock {\em arXiv preprint arXiv:1702.04595}, 2017.

\end{thebibliography}

\clearpage

\appendix

\end{document}